\documentclass[sn-mathphys,Numbered]{sn-jnl}% Math and Physical Sciences Reference Style
\usepackage{graphicx}%

\usepackage{amsmath,amssymb,amsfonts}%
\usepackage{amsthm}%
\usepackage{mathrsfs}%
\usepackage[title]{appendix}%
\usepackage[table,xcdraw]{xcolor}
\definecolor{green}{rgb}{0.60,0.98,0.60}
\usepackage{textcomp}%
\usepackage{manyfoot}%
\usepackage{booktabs}%
\usepackage{algorithm}%
\usepackage{algorithmicx}%
\usepackage{algpseudocode}%
\usepackage{listings}%
\usepackage{caption}
\usepackage{subcaption}

\usepackage{multirow}%
\usepackage{multicol}%
\theoremstyle{thmstyleone}%
\theoremstyle{thmstyletwo}%

\theoremstyle{thmstylethree}%
\begin{document}

\title[Article Title]{Context-Aware Temporal Embedding of Objects in Video Data}


\author*[1]{\fnm{Ahnaf} \sur{Farhan}}\email{rownak.utep@gmail.com}
\author*[1]{\fnm{M. Shahriar} \sur{Hossain}}\email{mhossain@utep.edu}

\affil*[1]{\orgdiv{Department of Computer Science}, \orgname{The University of Texas at El Paso}, \orgaddress{\street{500 W University Ave}, \city{El Paso}, \postcode{79968}, \state{Texas}, \country{USA}}}


\abstract{
In video analysis, understanding the temporal context is crucial for recognizing object interactions, event patterns, and contextual changes over time. The proposed model leverages adjacency and semantic similarities between objects from neighboring video frames to construct context-aware temporal object embeddings. Unlike traditional methods that rely solely on visual appearance, our temporal embedding model considers the contextual relationships between objects, creating a meaningful embedding space where temporally connected object's vectors are positioned in proximity. Empirical studies demonstrate that our context-aware temporal embeddings can be used in conjunction with conventional visual embeddings to enhance the effectiveness of downstream applications. Moreover, the embeddings can be used to narrate a video using a Large Language Model (LLM). This paper describes the intricate details of the proposed objective function to generate context-aware temporal object embeddings for video data and showcases the potential applications of the generated embeddings in video analysis and object classification tasks. 

%It contributes to the advancement of temporal object representation methods, offering a promising avenue for understanding complex video data and enhancing the performance of various computer vision applications.

%Through extensive experiments, we demonstrate the effectiveness and utility of our model in capturing temporal context and showcasing its potential applications in video analysis and video object classification tasks. This paper contributes to the advancement of temporal object representation methods, offering a promising avenue for understanding complex video data and enhancing the performance of various computer vision applications.

%This paper explores the concept of temporal embedding for visual objects in video data, presenting a novel approach to generate temporal object representations within a video sequence that can capture the context of objects. 

}




\maketitle

\section{Introduction}\label{sec:introduction}

The rapid advancement of technology and the widespread adoption of web and mobile applications have resulted in an exponential increase in data generation, particularly in the form of images and videos. This data, predominantly unstructured and unlabeled, presents significant challenges for extracting meaningful insights. The surge in unstructured video data necessitates innovative approaches for extracting useful features and trends for video analysis.

In the field of image and video data analysis, significant progress has been made in computer vision, particularly in areas such as classification~\cite{Krizhevsky2012, Kukleva2019UnsupervisedLO, biomedinformatics3040068}, recognition~\cite{He_2016_CVPR,Liang_2015_CVPR, Huang_2017}, object detection~\cite{He_2016_CVPR,Redmon_2018}, object tracking~\cite{Wojke_2017, luo2021multiple}, and segmentation~\cite{He_2017_ICCV,minaee2021image}. These developments primarily depend on visual features and employ various Convolutional Neural Network (CNN) models~\cite{Donahue13, dosovitskiy_2014, xu_2015_ieee, Gan_2018_CVPR}. While visual features are instrumental in many computer vision tasks, they are often insufficient for fully understanding scenes and events. The addition of contextual features can greatly improve the detection of patterns and insights in visual data. Contextual features involve examining the interaction between objects, providing essential insights for a deeper comprehension of the scene and a more accurate interpretation of the actions and events occurring within the scene.

In the domain of feature extraction from unstructured and unlabeled data, embedding models have become increasingly popular recently for their potential to extract and leverage both visual and contextual features effectively. Embeddings are compressed vector-space representations of objects. Contextual embedding brings contextually similar or connected objects near to each other in a vector space. Contextual embedding has been popular with text~\cite{Mikolov13a, Mikolov13b, Pennington_2014} and imagery data~\cite{LUDDECKE19, Frome13, Donahue13}. The concept of contextual object embedding in video data goes beyond traditional object embedding techniques in imagery data by incorporating the sequence information (temporal) of the frames. The frame-sequence information helps in encoding frame-level contexts, using the fact that visual objects that appear in nearby frames are more contextually connected than objects that appear only in nonconsecutive frames.
%The ability to compute contextual similarity of visual objects has the potential to aid in automatic generation of in-video ads, video recommendations, and video summarization. 
%While developing feature vectors for visual objects, 

\begin{figure}[!h]
\begin{center}
  \includegraphics[width=0.7\textwidth]{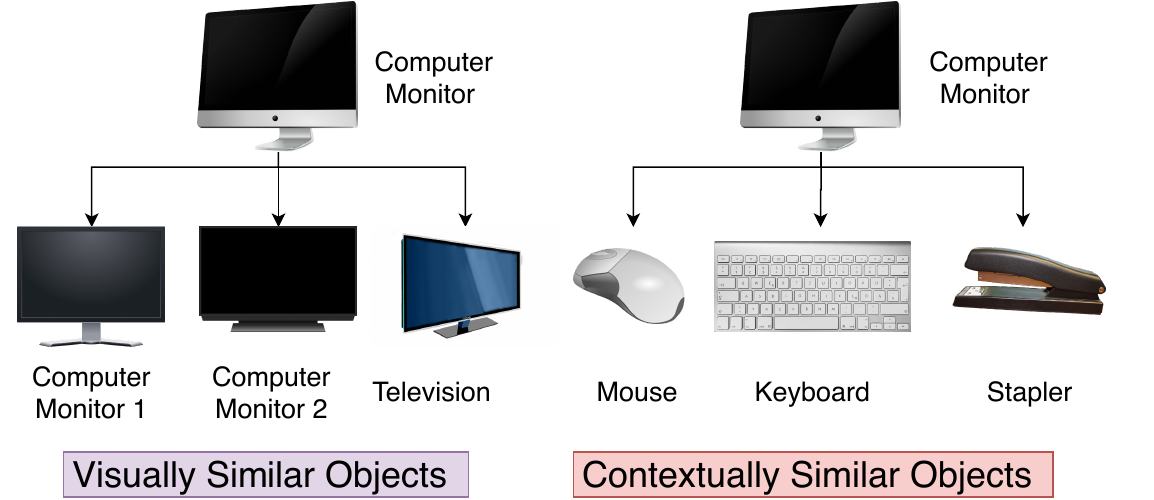}
\end{center}
\caption{Difference between visual similarity and contextual similarity.}
\label{fig:visual_vs_contextual}
\end{figure}
Most computer vision applications~\cite{Gasulla16, Donahue13} rely on visual similarity to generate feature vectors for visual objects. Visual similarity focuses on bringing objects that look similar in the vicinity. In contrast, contextual similarity brings visual objects -- that appear in the same frame or nearby frames of a video -- in proximity. As an example, Fig.~\ref{fig:visual_vs_contextual} shows that objects visually similar to a ``computer monitor'' are other computer monitors or televisions. In contrast, visual objects contextually similar to a ``computer monitor'' are a ``keyboard'', a ``mouse'', and a ``stapler''. Contextual similarity focuses on the spatial placement of visual objects in video frames. Additionally, in our work, the frame sequence plays an important role in constructing the context too. For example, if the visual object ``stapler'' is seen within a few frames after the visual object ``computer monitor'' is detected, then the visual object ``stapler'' will have some degree of contextual similarity with the ``computer monitor''. Despite the significant potential, the learning of contextual features remains largely unexplored in the computer vision area. Some recent studies~\cite{Ramanathan15,Knights2020TemporallyCE} have begun to integrate visual features from neighboring frames to capture the context. However, these approaches primarily focused on visual similarities, which do not correspond to contextual relevance.

\begin{figure}[!h]
\begin{center}
  \includegraphics[width=0.8\textwidth]{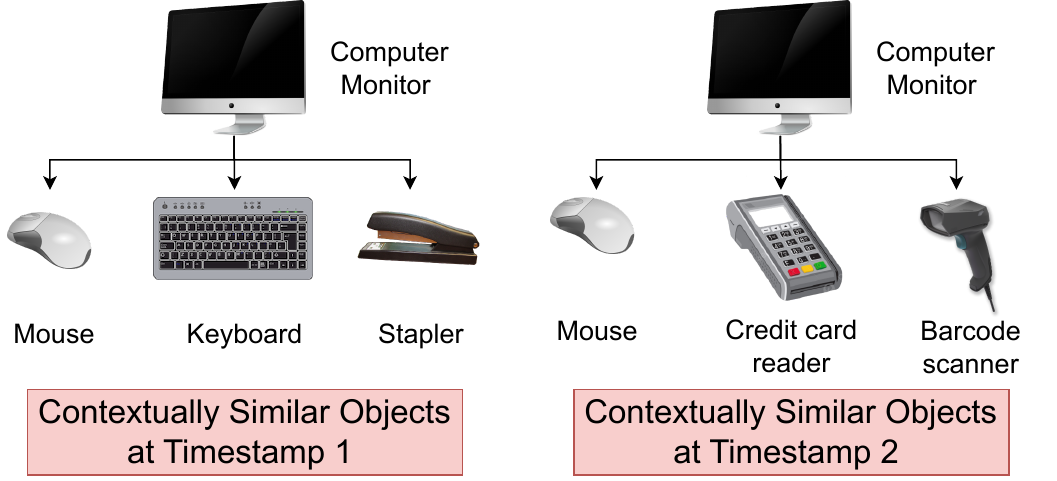}
\end{center}
\caption{Context changes over time.}
\label{fig:visual_vs_contxtual2}
\end{figure}

Furthermore, the context of objects within a video is dynamic and changes over time. Accurately modeling this temporal shift in object context is crucial for understanding how associations between objects evolve. For example, as shown in Figure~\ref{fig:visual_vs_contxtual2}, the transition of items associated with a computer monitor, from staplers to barcode scanners, indicates a shift from an office setting to a retail environment. Visual features can not capture these kinds of contextual transformations alone. Thus, it is important to develop temporal object embeddings that can precisely depict the changing context of objects in a video to capture these evolving associations effectively.

Additionally, a significant research gap exists in the area of temporal embedding for video data. Several recent research studies~\cite{Kukleva2019UnsupervisedLO, VidalMata2020JointVE} in this field concentrate on incorporating temporal context by predicting frame timestamps using frame features for specific applications such as segmentation and action recognition. However, these models do not generate separate embedding for each timestamp. Rather, they utilize temporal context to generate global embeddings of frames or other entities. This leaves a void in developing a generic, dynamic model capable of tracking the evolving context of objects in videos.

Temporal contextual embeddings of visual objects have the potential to aid in several computer vision applications, such as extracting insight from video and extracting narrative of video. Furthermore, the rich contextual information contained in the embeddings can be utilized in object detection models when visual features are obscure or missing. 
For example, pre-trained temporal contextual embeddings can help in recognizing an object by analyzing the context of the object. 
%Moreover, when an object is blocked by other objects which make the object partially visible, the context information from vizObj2Vec could be useful to infer the object.
%In our paper, we aim to demonstrate a similar self-supervised method for visual objects from video data. The supervised task involves labeling and getting the location of the visual objects from the video. We process the frames from the video data to extract the labels and location of visual objects. Then we train the visual object embedding model with all the pairs of objects including their distances from the same frame and also from surrounding frames. Our underlying hypothesis is that all the objects in the same frame and also from surrounding frames are contextually related. Training the embedding model with contextually related object pairs and their distances will result in a representation of visual objects which preserves contextual information. 

In this paper, we introduce a novel temporal contextual object embedding for video data, designed to capture the context of detected objects and their contextual changes over time. The objective is to construct dynamic embeddings that can capture the evolution of objects' context in video. This dynamic embedding comprises a sequence of vectors for every identified object at each timestamp in a video, where each timestamp includes consecutive frames. The model learns the temporal context of each object by leveraging the spatial distance between objects within each frame and adjacent frames, as well as the frequency of objects across all timestamps.
We use conventional visual object detection tools~\cite{Redmon16, Bochkovskiy20} to label visual objects in video frames as a preprocessing step of our model. Then, we train a neural network model to learn embeddings in a hidden layer from the distances between each pair of objects in each frame and several surrounding frames. Based on our investigations, visual objects in the same frame as well as from surrounding frames are contextually related. Our temporal contextual embedding model brings the visual objects that are within a similar context together in the embedding space. Furthermore, this model effectively tracks the evolving context of each object over time, thereby enriching the understanding of a video's narrative.

%VisualObj2vec has the potentials in several computer vision tasks. The rich contextual information contained in visualObj2vec can be utilized in computer vision systems when visual features are obscure or missing. For example, when object detection models get a problem to detect an object due to low confidence, visualObj2vec can help to recognize the objects by analyzing the context. Moreover, when an object is blocked by other objects which make the object partially visible or invisible, the context information from visualObj2vec could be useful to infer the object.

%In this paper, we propose visualObj2vec, a contextual visual object embedding model based on their co-occurrence in video frames. We provide visualization of the object vectors and analyze the captured latent relationship between objects. We highlighted the impressive performance on indoor-outdoor object separation by the projection of the object vectors in low dimensional space. As contextual embedding of the visual object in video data is not practiced in compter vision before, We have conducted empirical analysis and case-based analysis for different parameters and different methods that we implemented.

The contributions of our context-aware temporal embeddings of objects are summarized as follows:
\begin{itemize}
    \item We describe a context selection process for visual objects in video data that incorporates proximity between objects in each frame as well as the surrounding frames.
    \item We provide a mechanism to define the context of a visual object as a decay function of distances with visual objects in the surrounding frames.
    % \item We incorporate a mechanism to include motion in contextual embedding for visual objects in a video.
%\item We provide a mechanism to define frame difference function according to frame distance and motion of the events.  

\item We provide objective functions that leverage the distance and frequency of objects to learn temporal contextual embeddings of objects.
 \item We introduce a neural network model that utilizes the objective functions to create embedding vectors for each visual object for each timestamp in a video.
\item We demonstrate ways to combine our context-aware temporal embeddings with conventional visual embeddings to enhance downstream applications.
 
\end{itemize}

\section{Related Works}\label{sec:related_works}
% Related works from chatgpt
Temporal modeling of visual data, particularly in the context of video analysis, has garnered substantial attention within the computer vision community. Over the years, a diverse array of techniques and methodologies has emerged, aiming to capture the temporal dynamics of objects and scenes. In this section, we provide an overview of the key approaches and advancements of embedding models in text, image, and video datasets that have paved the way for our proposed temporal visual object embedding model.

\subsection{Text Embeddings}

The use of context, which is implicitly derived from the data, has been proven to be a powerful source of information for learning representations in natural language processing (NLP)~\cite{Mikolov13b, Mikolov13a}. Contextual embedding maps each word of a large text corpus to a low dimensional feature vector. The embedding model self-supervises by observing each word and the words surrounding it in the original data to preserve the contextual relationships between words in the generated vector space. A wide variety of applications in NLP, such as semantic word similarity~\cite{Mikolov13a}, automatic summarization~\cite{kageback14}, named entity recognition~\cite{Siencnik15}, sentiment analysis~\cite{Xue14}, and identifying future trends~\cite{Farhan_2023} benefit from contextual embeddings. In this paper, we present a model that creates contextual embeddings for visual objects from video data instead of text.

\subsection{Image Embeddings}
Embedding models are widely used in the imagery domain, encompassing tasks such as retrieval~\cite{barz2019hierarchy, Cao_2017_CVPR}, few-shot learning~\cite{Tian_2020,ye2020few}, self-supervised learning~\cite{Rahman_2022,trinh2019selfie}. The most popular application of embedding in image collections is the classification of images~\cite{akata2015label}. A common approach involves training a CNN with labeled images to generate representations of image classes, such as dog, cat, chair, and table~\cite{Gasulla16, Donahue13}. Frome et al.~\cite{Frome13} developed a method to embed both images and corresponding labels into a joint space. Frome's deep visual-semantic embedding model (DeViSE), was trained to identify visual objects using both labeled image data and the semantic information from unannotated text. All these methods train a CNN with the images, associated labels, and/or unannotated text around the images to generate the image class embedding. While many applications focus on the visual features of the image and semantics of labels, a wide variety of other applications require contextual embedding, which is overlooked in these methods. Although our approach is designed for video data, our proposed model is also suitable for imagery data. In imagery data, our model considers the context of objects within each image. With video data, objects in the surrounding frames are also included in the context. 

The work of Lüddecke et al.~\cite{LUDDECKE19} included image-level context for each visual object, which closely relates to our work. However, Lüddecke's approach is not suitable for an extension to video data where the context of an object may span multiple surrounding frames. In our work, the context of a reference object is a continuous function of the positions of the other objects in the current frame (where the reference object is situated) and the surrounding frames. Later in the experimental results section, we demonstrate the strength of our proposed definition of context in creating embedding for objects in video data and image data. 

\subsection{Video Embeddings}
%Habibian et al.~\cite{Habibian17} designed a model to construct embedding for video clips using textual descriptions associated with the clips.

Various categories of embedding models have been explored for video data, including embeddings for entire video clips ~\cite{Hu16, Habibian17, Wang2018, zhuang_2020}, video frames~\cite{Misra16,Lee17, Ramanathan15}, and visual objects~\cite{mettes_2017}. Despite the differences in the entities for which these embeddings are learned, the underlying goal remains consistent: to develop representations that effectively group similar entities while distinguishing dissimilar ones.
In ~\cite{zhuang_2020} features of video frames are projected into a compact latent space via a deep neural network, whose parameters are then tuned to optimally distribute embedded video instances so that similar videos aggregate while dissimilar videos separate. Inspired by word embedding models, the method proposed by Han et al.~\cite{Han2019} involves segmenting a video clip into several distinct, non-overlapping blocks where the blocks are characterized by a unique latent representation. Subsequently, a neural network is employed to learn contextual representation of video blocks by predicting the latent representations of future video blocks, given the representation of past blocks. Several other research efforts focused on generating embedding for units smaller than clips, such as video frames. Misra et al.~\cite{Misra16} and Lee et al.~\cite{Lee17} trained the sequence of frames of a video in a Convolutional Neural Network (CNN) to generate an embedding-vector for each frame. All these approaches focus on generating vectors/embeddings for video clips or video frames by leveraging visual features, whereas the goal of our work is to obtain embedding for visual objects in a video by leveraging contextual features. 

In the subsequent part, we concentrate on literature pertaining to embedding models that capture contextual and temporal features in video data, aligning closely with the focus of our work.

\subsubsection{Contextual Embeddings}\label{subsec:contx_emb}
Our proposed model is rooted in the concept of contextual embeddings. This approach resonates with contextualized word embeddings in natural language processing, where words are embedded based on their contextual usage in sentences. Similar ideas have been explored in video analysis. Bertasius and Torresani~\cite{Bertasius2020} proposed a model that uses a pretrained language model to learn contextualized object embeddings by associating video frames from instructional videos with text narrations. It employs a detection model to predict object instances and corresponding embeddings and a contrastive loss function to align these embeddings with contextualized word embeddings. This approach is limited only to the video with text multimodal data. 

~\cite{Ramanathan15} employed a skip-gram model on video frames to learn contextual embedding of frames. ~\cite{Knights2020TemporallyCE} propose a method similar to skip-gram, where their loss function promotes high cosine similarity between embeddings of adjacent frames in a video while ensuring low similarity with negative frames from different videos. In addition to the objective of bringing adjacent frames closer, ~\cite{Pan2016LearningDI} also aims to cluster visually similar frames by incorporating a graph structure regularization based on a similarity matrix encompassing all video frames. All these approaches are based on the concept that context frames around the target frame represent the context of the target frame better than other frames. However, they focus on a limited range of adjacent frames, overlooking the potential context from distant frames. 

Our work differs by introducing a context discrepancy score between objects in target and surrounding frames, assigning lower scores to nearby objects and higher to distant ones. This approach enables the capture of long-range contextual relationships. Initially, our research concentrated on the static contextual embedding of objects within video data, with some findings already published~\cite{farhan_2020}. In this paper, we extended our research towards the temporal embedding of objects in video data, exploring how these objects interact and how the context of objects changes over time.

\subsubsection{Temporal Embeddings}\label{subsec:temp_emb}
In recent years, there has been a shift toward embedding-based approaches to capture temporal dynamics. Embeddings offer a compact and expressive representation of objects that can encompass both visual appearance and temporal context. To capture this temporal context, some techniques employ frame features to estimate the frame's relative time.
Kukleva et al.~\cite{Kukleva2019UnsupervisedLO} utilize frame features to predict the time of frames, aiding in learning temporal embeddings. Vidal-Mata et al.~\cite{VidalMata2020JointVE} adopt a temporal self-attention mechanism, combining a visual embedding from a predictive U-Net architecture with a temporal embedding that predicts the timestamp of given frames. Some other transformer-based models focus on image patches or object bounding boxes instead of whole frames. For instance, Wang et al.~\cite{Wang2021BEVTBP} introduce a transformer-based framework featuring dual encoders for image and video streams, employing decoders to generate spatio-temporal representations by predicting tokens for masked image and video patches. Zhang et al.~\cite{Zhang2022IsAO} present a transformer-based method, where an encoder generates visual representations, while a trajectory encoder processes object bounding boxes. An Object Learner module merges these streams using cross-attention Transformers to generate spatio-temporal embedding of the object bounding box. Though these approaches are powerful, they face challenges due to their high computational demands and limited capacity in contextual data aggregation. Furthermore, these models are not focused on tracking contextual changes over time, as they do not create embeddings for the same instance at different timestamps. Our model constructs temporal contextual object embeddings that explicitly capture not only the context of objects but also the change of context over time.

%Several works have explored learning temporal representations for video data using unsupervised, self-supervised, or weakly supervised techniques. These approaches often leverage large-scale video datasets and contrastive learning to capture meaningful temporal relationships. 

%While their focus is on learning generalizable representations, our model is designed with a specific emphasis on enhancing object-level representations within a temporal context. 

% \subsubsection{Temporal Context and Semantic Relationships}\label{subsec:temp_contxt}
% The integration of semantic relationships between objects and their temporal context has led to more robust video analysis techniques. Graph-based methods, for instance, model objects as nodes and their interactions as edges in a temporal graph. These approaches leverage graph convolutional networks to capture dependencies between objects over time, enhancing both object recognition and scene understanding. However, they often depend heavily on accurate object tracking, making them susceptible to tracking errors and object misalignments.

\subsubsection{Temporal Object Embedding}\label{subsec:temp_obj_repr}
In video analysis, it is crucial to understand how object representations evolve over time in a video. This requires establishing a dynamic embedding space that can generate embeddings for objects at individual time slices. Despite its importance, research in this domain is sparse. Some studies have concentrated on trajectory-based methods, utilizing object motion paths to decipher temporal embeddings, specifically for object tracking applications. Yan et al.~\cite{Yan2021LearningST} developed a transformer-based model with an encoder that handles spatio-temporal feature dependencies and a decoder dedicated to predicting spatial positions of target objects. Their approach is enriched by a dynamically updated template from intermediate frames, enhancing the temporal information with changes in target appearance. On a similar note, Wan et al.~\cite{Wan2021MultipleOT} pioneered the use of temporal priors embedding, utilizing long-term dynamics of tracked targets over video clips. This method employs logical reasoning to assess the activation status of a target, maintaining accurate tracking even under occlusion and identifying when targets enter or leave the scene. While these methods significantly advanced in tracking object representation over time, they fall short of capturing the context of objects.

In summary, the landscape of temporal modeling in video analysis has evolved significantly, with a growing recognition of the importance of contextual information and temporal relationships. Our work builds upon these foundations by introducing a novel approach that leverages both adjacency and contextual similarities between objects to create coherent temporal embeddings. By combining the strengths of contextualized embeddings and temporal modeling, we aim to provide a more comprehensive representation of visual objects in video data, contributing to the broader advancement of video analysis techniques.

\section{Problem Description}\label{sec:problem_description}

Let $F=\{f_1, f_2,....f_{|F|}\}$ be a video consisting of $|F|$ frames and $O=\{o_1, o_2,\ldots,o_{|O|}\}$ be the set of $|O|$ visual objects extracted from the video $F$. As an example of visual objects -- $o_i$ can be any item such as a cup, a table, a refrigerator, or any artifact detected by preprocessing tools, such as YOLO9000~\cite{Redmon16} and YOLOv4~\cite{Bochkovskiy20}, or any human-annotated visual objects. 
The set of visual objects in frame $f$ is $O_f \subset O$. In this paper, the phrase \textit{visual object} and the word \textit{object} are used interchangeably. 

The objective of this paper is two-fold, as outlined below.

\begin{enumerate}

    \item \noindent \textbf{Static Embedding Generation for Visual Objects:} Our initial approach is to develop a static embedding model. This model constructs an embedding set $E_{static}=\{e_1, e_2, \ldots, e_{|O|} \}$ for each visual object in $O$ within video $F$. The output, $E_{static}$, is a $|O|\times|e|$ matrix where $|e|$ is a user-defined integer parameter denoting the length of each embedding vector.

    \item \textbf{Temporal Embedding Generation for Visual Objects:} 
Considering our objective of temporal embedding, we focus on video recordings captured over extended periods. Such recordings, like security camera videos marked with date and time, or dashcam clips with timestamps, are ubiquitous. We define $T=\{t_1, t_2,....t_{|T|}\}$ as the set encompassing $|T|$ timestamps. Each timestamp $t_j$ has $n_f = \lceil |F|/|T| \rceil$ frames.
%All the video frames $f_i \in F$ are uniformly distributed across these timestamps and are labeled accordingly. Each timestamp $t_j$ comprises $n_f = \lceil |F|/|T| \rceil$ frames. 
Here, we assume the timestamps are ordered chronologically. We note that the set of visual objects $O$ contains all visual objects. It is possible for certain $o_i \in O$ to be absent in some timestamps $t_j \in T$.

Given such a time-annotated video dataset, our goal is to formulate a temporal embedding model denoted as $E_{temporal}$. This model is represented as $E_{temporal}=\{{[e_1, e_2, \ldots, e_{|O|}}]_1, [{e_1, e_2, \ldots, e_{|O|}}]_2, \ldots , [{e_1, e_2, \ldots, e_i, \ldots, e_{|O|}}]_j , \ldots, [{e_1, e_2, \ldots, e_{|O|}}]_{|T|} \}$ for all visual objects in $O$ and across all timestamps $T$ derived from video $F$. The output temporal embedding, $E_{temporal}$, is a $|O|\times|T|\times|e|$ matrix where $|e|$ is a user-settable integer parameter denoting the length of each embedding vector. 

\end{enumerate}

\begin{figure*}[!h]
\centering
\begin{tabular}{c}
\includegraphics[width=0.9\textwidth]{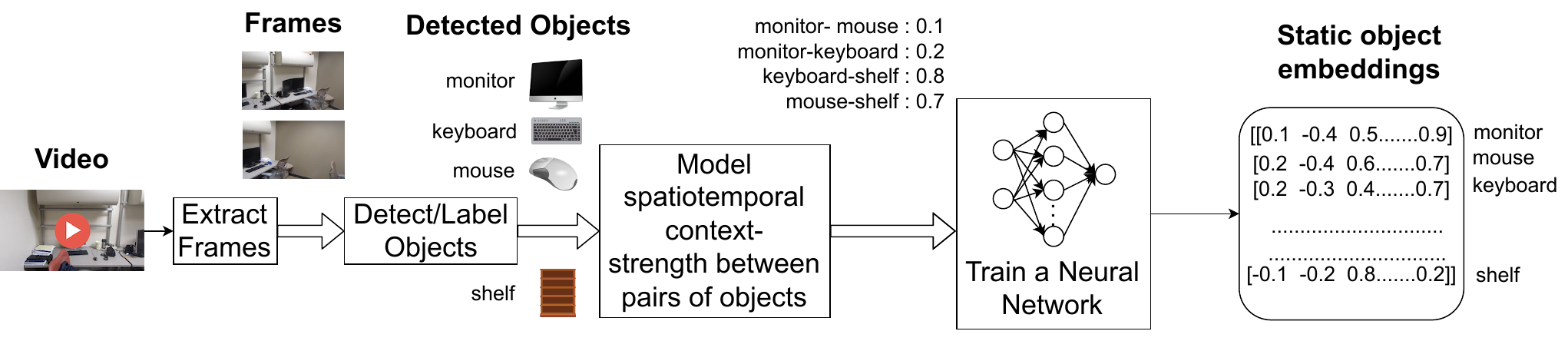}\\
(a) Static Embedding\\
\\
\includegraphics[width=1\textwidth]{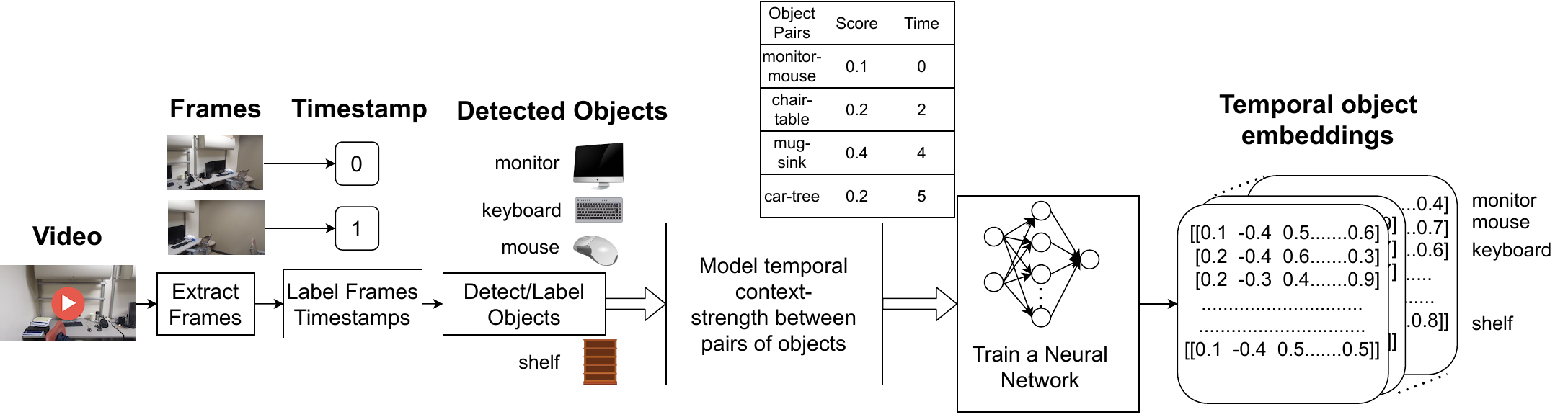}\\
(b) Temporal Embedding\\
\end{tabular}
\caption{Complete pipeline of embedding-generation from a video.}
\label{fig:pipeline}
\end{figure*}

\begin{table}[!h]
\caption{A list of selected symbols used in this paper.}\label{tab:symbols_list}%
\begin{tabular}{|l|p{0.8\linewidth}|}
\toprule
Symbol  & Description\\
\midrule
$O$ & The set of all visual objects detected or labeled in a video.\\
\midrule
$O_f$ & The set of visual objects in frame $f$.\\
\midrule
$T$ & The set of timestamps. $|T|$ denotes the total number of timestamps.\\
\midrule
${t_i}$ & A timestamp with sequence number $i$.\\
\midrule
$F$ & The set of all frames extracted from a video. $|F|$ denotes the total number of frames.\\
\midrule
${f_i}$ & A frame with sequence number $i$.\\
\midrule
${f^j}$ & Set of frames present in the $j^{th}$ timestamp.\\
\midrule
${n_f}$ & Number of frames per timestamp.\\
\midrule
$\vartheta_{k}(E)$ & $k_{th}$ objective function.\\
\midrule
$E_{static}$ & 2-dimensional embedding matrix that contains the static embeddings for each object.\\
\midrule
$E_{temporal}$ & 3-dimensional embedding matrix that contains the embeddings for each object at every timestamp.\\
\midrule
${e_i}^j$ & The embedding vector of a object $i$ at timestamp $j$.\\
\midrule
$e_i$ &  Embedding vectors of the object $i$ in static embedding. The combined embedding vectors of the object $i$ across all timestamps in temporal embedding.\\
\midrule
$N$ & Number of object pairs in training data.\\
\midrule
$\delta_i(o_r, o_c)$ & Normalized distance between $i^{th}$ pair of objects $o_r$ and $o_c$ in training data.\\
\midrule
$\delta_i(o_r, o_c, t_r)$ & Normalized distance between $i^{th}$ pair of objects $o_r$ and $o_c$ at timestamp $t_r$.\\
\midrule
$\gamma(t_r, \sigma)$ & A list of temporal diffusion weights where the reference timestamp is $t_r$.\\
\midrule
$\gamma(t_r,t_c, \sigma)$ &  Temporal diffusion weight of timestamp $t_c$ with $t_r$ being the central timestamp.\\
\midrule
$dist(x,y)$ & Calculate the cosine distance between two vectors $x$ and $y$.\\
\midrule
$dist(x,y,t)$ & Calculate the cosine distance between two vectors $x$ and $y$ at timestamp $t$.\\
\midrule
$N_g(f(o_k,t))$ & Normalized frequency of object $o_k$ at timestamp t.\\
\botrule
\end{tabular}
\end{table}

\section{Methodology}\label{sec:methodology}
Fig.~\ref{fig:pipeline} summarizes the pipeline of the static and temporal visual object embedding frameworks. Table \ref{tab:symbols_list} lists symbols used in this paper. The static embedding framework (Figure~\ref{fig:pipeline}(a)) takes a video as input, extracts the frames, detects visual objects and the location of the objects in frames, computes the contextual similarity between pairs of objects, and then trains a neural network to generate static embeddings of size $|O|\times|e|$. 

To construct a static embedding space, we design an objective function that considers the spatial distance between each pair of objects in a reference frame and surrounding frames, including a frame-level diffusion. The objective function minimizes the difference between similarities of a pair of embeddings and computed contextual similarity between the two objects in the corresponding pair. 
 
The temporal embedding framework (Figure~\ref{fig:pipeline}(b)) takes a video as input, extracts the frames, labels each frame with corresponding timestamps, detects visual objects and the location of the objects in frames, generates the temporal contextual similarity between each pair of objects, and then trains a neural network to generate temporal embeddings of size $|O|\times|T|\times|e|$. 
The objective function for constructing a temporal embedding space brings important considerations into account, such as the weight of interaction between each pair of objects, temporal diffusion, the frequency of objects across timestamps, and the connection between embedding spaces of different timestamps.

For both models, static and temporal, a tailored contextual similarity between each pair of objects aids in creating the training data. We use a shallow neural network to accommodate the embedding generation with our intricate objective functions. The designed objective functions make the generated embeddings reflect more latent contextual relationships. Subsequent subsections provide details of the training data generation process and the objective functions for both static and temporal embedding generation frameworks.

%This section describes the details of our vizObj2Vec model to create contextual embedding for each object detected or annotated in a video frames. 

% Given a large collection of unlabeled video frames, our goal is to learn an effective contextual embedding for all the visual objects in the frames. We wish to learn an embedding such that, the context objects surrounding a specific reference object can determine the representation of that reference object, similar to the intuition from word2vec~\cite{Mikolov13a}. The specific object for which we are generating vector will be called as reference object throughout this paper. Every object in a video is considered reference object one by one. Now the question is how to select the context of a reference object? In this section, we will first describe how we define object context in a video and then how we train a neural network model with the context of visual objects to obtain the contextual representation of visual objects. 

\subsection{Frame Extraction}
Frames are initially extracted from the video at a predetermined frame-per-minute rate. For the static embedding approach, these frames are labeled solely based on their frame numbers, resulting in the sequence $F=\{f_1, f_2,\ldots,f_{|F|}\}$, where $|F|$ is the total number of frames extracted from a video.

In temporal embedding, as mentioned in the problem description (Section \ref{sec:problem_description}), each timestamp refers to a segment of the given video footage. The number of timestamps, $|T|$, is a user-provided parameter. The number of frames per timestamp is $n_f =\lceil |F|/|T| \rceil$. %Let $f_i^j$ be the $i$th frame in timestamp $t_j$. 
We denote all the frames as, $F=\{{f_1}^1, {f_2}^1,\ldots,{f_{n_f}}^1,{f_{(n_f+1)}}^2,\ldots,{f_{(n_f*|T|)}}^{|T|}\}$, where the subscript of a frame indicates the frame's sequence number in the video and the superscript indicates in which timestamp the frame belongs. 

\subsection{Detection and Selection of Context Objects}\label{sec:context_selection}
We use YOLO9000~\cite{Redmon16} and YOLOv4~\cite{Bochkovskiy20} to detect and label all the visual objects in every frame of a video, or we manually label the objects in reasonably-sized video for our experiments. In addition, we keep track of (i) the location coordinate of the center of each detected visual object in each frame and (ii) the timestamp of the corresponding frame where the object is being detected. 

In text embedding (such as word2vec \cite{Mikolov13a}), a context is defined by a window of words surrounding a center word. Each word in every document is considered as a center word when embedding vectors are being generated. Mikolov et al.~\cite{Mikolov13a, Mikolov13b} defined the context window size as an integer number, say $k$, where the window contains $k$ previous words and $k$ next words of the center word. That is, the window size $k$ represents a context window of $2k+1$ words, including the center word. 

\noindent\textbf{The difference between visual object embedding and text embedding:} The selection of visual context objects surrounding a visual reference object (analogous to a center word in text embedding) is not straightforward in our work because we do not consider a frame a sequence of objects like a document is considered a sequence of words in a text corpus. Moreover, objects in surrounding frames may play a role in the context of a visual object in a video, whereas in a document collection, the sequence of documents does not play any role in conventional text embedding. 

\noindent\textbf{Definition of a context window:} 
%In our research, we have experimented with several mechanisms to define \textit{a context window for a visual object} in a video frame. 
Considering each of the detected objects as reference objects, the context objects of a reference object are selected using either one of the following mechanisms or a combination of some of the following mechanisms: 
\begin{itemize}
\item \textbf{Context 1: A frame as a context window:} In this mechanism, a frame is regarded as a context window. Considering each object in a frame as the reference object, all other objects surrounding the reference object are considered the context of the reference object. Let $o_r$ be a reference object of a frame $f_r$. Each visual object (excluding the reference object) $o_c$ in the frame $f_r$ is considered a context object of the reference object $o_r$. Considering $O_{f_r}$ is the set of objects from frame $f_r$, then the context objects are $\{o_c : o_c \in O_{f_r}, o_c \neq o_r\}$. 

% \begin{figure}[!h]
% \begin{center}
%   \includegraphics[width=0.9\textwidth]{Figures/method_1_context_selection.pdf}
% \end{center}
% \caption{Illustration of visual objects, the center frame $f_r$ (the light green frame), and surrounding frames (light red frames) from set $sFrames(f_r,w_f)$.}
% \label{fig:context_selection}
% \end{figure}

\begin{figure}[!h]
\begin{center}
  \includegraphics[width=0.8\textwidth]{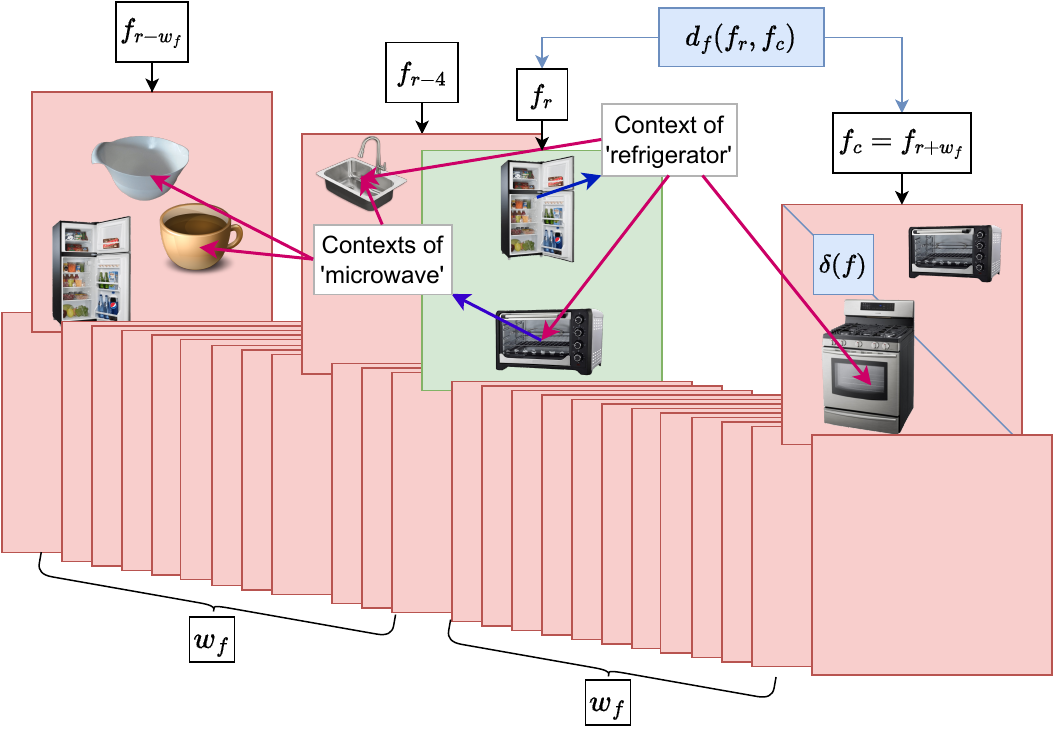}
\end{center}
\caption{Illustration of visual objects, the reference frame $f_r$ (the light green frame), and surrounding frames (light red frames) from set $sFrames(f_r,w_f)$.}
\label{fig:context_selection}
\end{figure}

\item \textbf{Context 2: Surrounding frames as the context window:} In this mechanism, the surrounding frames of a reference frame are regarded as a context window for every object in the reference frame. Considering each object in the reference frame as the reference object, all the objects in the surrounding frames of the reference frame are considered the context of the reference object. Let $\text{sFrames}(f_r, w_f)$ be the surrounding frames of a reference frame $f_r$, such that $\text{sFrames}(f_r, w_f)=\{f_{r-w_f}, f_{r-(w_f-1)}, \ldots, f_{r-1},f_{r+1} \ldots, f_{r+(w_f-1)}, f_{r+w_f}\}$ (Fig.~\ref{fig:context_selection}). Here,  $w_f$ is a user-settable parameter that indicates the number of frames, prior and after $f_r$, selected for context objects. The parameter $w_f$ can also be described as the frame window size. Considering $O_{f_k}$ is the set of objects in frame $f_k$, the context objects of reference $o_r$ are $\{o_c : o_c \in \cup_{k=r-w_f}^{r+w_f}{O_{f_k}}, o_c \not \in O_{f_r}\}$. Here $O_{f_r}$ denotes the set of objects in reference frame $f_r$. Objects from the reference frame are excluded since these objects are chosen as a context in the previous ``Context 1: A frame as context window'' approach.

\begin{figure}[!h]
\begin{center}
  \includegraphics[width=1\textwidth]{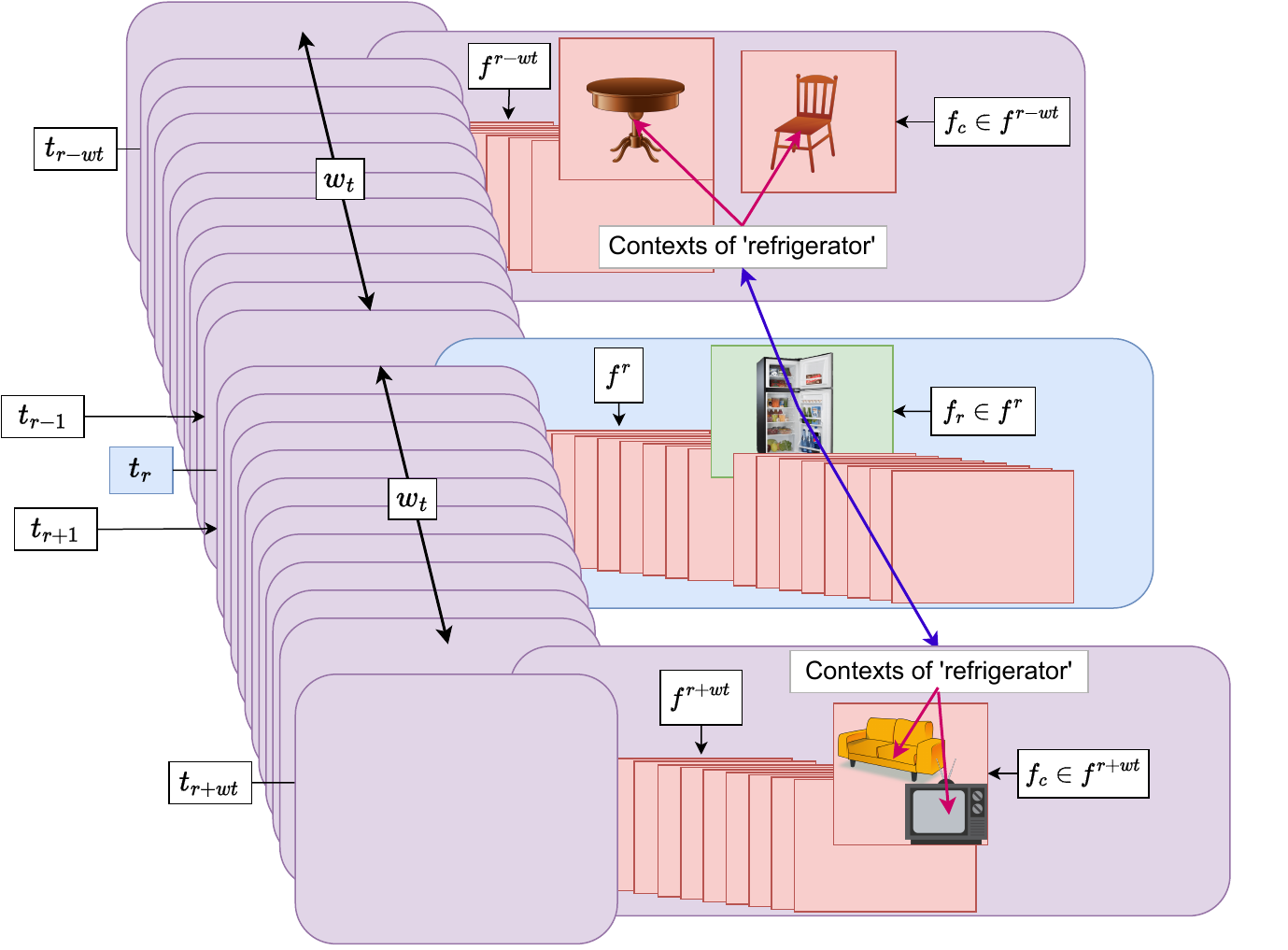}
\end{center}
\caption{Illustration of visual objects, surrounding timestamps, and the reference frame $f_r$ (the light green frame). The purple rectangle denotes the surrounding timestamps from set $sTimestamps(t_r, w_t)$.}
\label{fig:context_selection2}
\end{figure}

\item \textbf{Context 3: Frames from neighboring timestamps as context window:} Providing all the frames are labeled with timestamps, in this mechanism, context objects are selected from frames associated with timestamps adjacent to the central frame's timestamp. Considering each object in the reference frame as the reference object, objects in the frame from neighboring timestamps are considered the context of the reference object. Let $\text{sTimestamps}(t_r, w_t)$ be the neighboring timestamps of reference timestamp $t_r$ such that the reference frame $f_r$ belongs to the timestamp $t_r$. Here $\text{sTimestamps}(t_r, w_t)=\{t_{r-w_t}, t_{r-(w_t-1)}, \ldots, t_{r-1},t_{r+1} \ldots, t_{r+(w_t-1)}, t_{r+w_t}\}$ (Fig.~\ref{fig:context_selection2}). Thus, any selected frame from the neighboring $\text{sTimestamps}(t_r, w_t)$ is, $\{f_c : f_c \in \cup_{p=r-w_t}^{c=r+w_t}{f^p}, f_c \not \in f^r\}$ . Here,  $w_t$ is a user-settable parameter that indicates the number of timestamps, prior and after $t_r$, selected for neighbor timestamps. The parameter $w_t$ can also be described as the time window size. The context objects of reference $o_r$ are $\{o_c : o_c \in \cup_{p=r-w_t}^{r+w_t}{O_{f^p}}, o_c \not \in O_{f^r}\}$. Here, $O_{f^r}$ denotes the set of objects across all frames corresponding to the reference timestamp $t_r$. Objects from the reference timestamps are excluded because many of these objects have been identified as context in our previous two approaches. Our aim in this approach is to recognize and encapsulate new contextual objects from surrounding timestamps that are not found within the reference timestamp.

% \item \textbf{Neighbor of context objects:}
\end{itemize}

In our work, contexts 1 and 2 are used in the static embedding generation, whereas all three contexts are used in the temporal embedding generation process.

\subsection{Training Data Preparation: Contextual Discrepancy Score between a Reference Object and a Context Object}
\label{sec:context_Strength}
%Our vizObj2Vec neural network model fits a score representing the strength of the context between every center-context object pair (Fig.~\ref{fig:vizObj2Vec}). 
%The context strength is a monotonically decreasing nonlinear function of distance between the center-context object pair. The idea is to set a maximum score to the nearest context objects and minimum score for the far-most context objects and predict the cosine similarity between the center-context objects pair. In the next sub-section, we have discussed the context object score selection process.
%We have two different types of context objects, a) Context objects from the current frame b) Context objects from surrounding frames. 
%Calculating distance between the center-context pair from current frame is a simple task. 
%We take the euclidean distance between the pixel location of center point of center-context pair which is expressed as, 

We generate embeddings using a neural network setting, where in the input, we provide a pair of objects -- the reference object and a context object -- and in the output, we provide the contextual discrepancy between the two objects. The contextual discrepancy between two objects reflects the contextual difference between the objects. The higher the contextual discrepancy score, the more different the contexts of the two objects are. We do not need a timestamp for static embeddings as an input to the neural network. However, for the neural network to generate temporal embeddings for objects, we need the timestamp of the reference object as an input in addition to the input objects. The contextual discrepancy scores of all contextual pairs of objects compose the training data for embedding generation. 

We explain contextual discrepancy score computation for static embeddings in Section \ref{sec:diffusion_for_static_embedding} and temporal embeddings in Section \ref{sec:diffusion_for_temporal_embedding}.

\subsubsection{Diffusion-based contextual discrepancy score for static embedding}\label{sec:diffusion_for_static_embedding}

We first assess the spatial distance between objects within frames to compute the contextual discrepancy score. The spatial distance between any reference object $o_r$ and a context object $o_c$ is calculated using the Euclidean distance, $d(o_r,o_c) = \lVert{o_r,o_c}\rVert$. This measurement stems from the central coordinates of each object in their corresponding frames.
\begin{equation}\label{eq:spatial_distance}
\begin{split}
\emph{d}(o_r,o_c) &= \lVert{o_r,o_c}\rVert \nonumber
\end{split}
\end{equation}

We used three approaches to normalize the distances: i) Distance threshold ii) Min-Max Scaling and iii) Gaussian Decay. Using the distance threshold approach, the reference-context discrepancy score is set to 0 if a context object is within $d_\theta$ distance of the reference object, otherwise, it is set to 1. Here, $d_\theta$ is a hyperparameter, which can be fine-tuned during experimental trials.
\begin{equation}
N_t(o^r,o^c) = 
     \begin{cases}
       \text{0} &\quad\text{; if }d(o^r,o^c) < d_\theta\\
       \text{1} &\quad\text{; otherwise.} \\ 
     \end{cases}
\label{eq:threshold}
\end{equation}

As the minimum possible distance between two objects is zero, the min-max scaling of the distances could be expressed using the following equation:
\begin{equation}\label{eq:min_max_norm}
\begin{split}
N_m(o_r,o_c) &= \frac{\emph{d}(o_r,o_c)}{\max(\emph{d})}
\end{split}
\end{equation}
where $\max(\emph{d})$ is the longest possible distance of two objects in a frame which the maximum value for all pairs in $d(o_r, o_c)$.

For gaussian distribution we convert the distance $d(o_r,o_c)$ to a nonlinear score using a Gaussian decay function. 
\begin{equation} \label{eq:gauss_decay}
\begin{split}
\emph{G}(o_r,o_c) &= \dfrac{1}{\sqrt{2\pi{\sigma}^2}}{e}^{\dfrac{-(d(o_r,o_c))^2}{2{\sigma}^2}} \\
\end{split}
\end{equation}
\noindent where $\sigma$ is a user-settable parameter.
%The Gaussian distribution of the euclidean distances helps us to maximize the score of nearest objects and minimize the score of far-most objects in a non-linear way. The Gaussian distribution of distance is described in the following equation.\\

To bring the Gaussian distance into the range of 0 to 1, we first determine its maximum potential value, given as $\emph{G}_\text{max} = \dfrac{1}{\sqrt{2\pi{\sigma}^2}}$, assuming the smallest possible distance is $0$. By dividing $\emph{G}(o_r,o_c)$ with $\emph{G}_\text{max}$, the resulting values range between 1 (indicating the closest objects) and 0 (signifying the most distant objects). The reference-context discrepancy score is then derived by inverting this normalized Gaussian decay value, as
% \begin{equation}
% \begin{split}
% \emph{G}_\text{max} &= \dfrac{1}{\sqrt{2\pi{\sigma}^2}}
% \end{split}
% \end{equation}

\begin{equation}
\begin{split}
N_g(o_r,o_c) &= 1 - \frac{\emph{G}(o_r,o_c)}{\emph{G}_\text{max}}\\
\end{split}
\label{eq:gauss_decay_norm}
\end{equation}

\noindent $N_g(o_r,o_c)$ varies between 0 and 1. A value of 0 indicates the minimum possible discrepancy score between the reference-context object pairs. A value of 1 indicates the maximum discrepancy score.

The preliminary contextual discrepancy score for a reference-context object pair is denoted by $\delta(o_r,o_c)$. This score can be derived from either the min-max scaling $N_m(o_r,o_c)$ as per Eq.~\eqref{eq:min_max_norm} or the normalized Gaussian decay $N_g(o_r,o_c)$ as described in Eq.~\eqref{eq:gauss_decay_norm}. This preliminary contextual discrepancy score is employed in training the static embedding model.

%The initial context score between two objects is determined by measuring the distance between their central coordinates within their respective frames. This distance is then normalized using either min-max normalization or a Gaussian decay function. When comparing reference-context object pairs from different frames or timestamps, we incorporate specific weighting factors which are elaborated upon in the following sections.

\subsubsection{Diffusion-based contextual discrepancy score for temporal embedding}\label{sec:diffusion_for_temporal_embedding}

% In order to construct a training dataset, we collect all the reference context object pairs, the context weights, and timestamps of the corresponding reference objects in a weight list $\delta_{ijt} = \delta(O_i, O_j, t_r)$. Here, the list $\delta_{ijt}$ is a list containing $\lvert\delta_{ijt}\rvert$ number rows representing the number of reference-context object pairs.

We applied the preliminary contextual discrepancy score in the previous section when selecting objects from the current and surrounding frames. For temporal embeddings, the context objects are also picked from neighboring timestamps, and we modify the preliminary score $\delta(o_r, o_c)$ with a weight factor. The weight for each timestamp is derived using a Gaussian filter, as shown in (Eq.~\ref{eq:gauss}). 

% \begin{equation}\label{eq:gauss}
%     \gamma(t, \sigma)=\left<\left(\frac{1}{\sqrt{2\pi\sigma^2}}e^{-\frac{(i-t)^2}{2\sigma^2}}\right)\text{with } i=1,\ldots,|T|\right>
% \end{equation}

\begin{equation}\label{eq:gauss}
    \gamma(t, \sigma)= \cup_{i=1}^{|T|}{\left(\frac{1}{\sqrt{2\pi\sigma^2}}e^{-\frac{(i-t)^2}{2\sigma^2}}\right)}
\end{equation}

The Gaussian filter ensures that the weight of each timestamp is smoothly distributed between different timestamps. $\sigma$ is a user-settable parameter representing the standard deviation of the Gaussian distribution. 

We invert the Gaussian distribution value to ensure that object pairs from closer timestamps are given lesser weights than those from farther. After applying the Gaussian filter, the distance between a reference object $o_i$ at time $t_r$ and a context object $o_j$ at time $t_c$ is 

\begin{equation}
\begin{split}
\delta(o_r, o_c, t_r) = \delta(o_r, o_c)\cdot(1-\gamma(t_r,t_c, \sigma))
\end{split}
\label{eq:temporal_weight_dist}
\end{equation}

\noindent where,

\begin{equation}\label{eq:gauss2}
    \gamma(t_r,t_c, \sigma)=\left(\frac{1}{\sqrt{2\pi\sigma^2}}e^{-\frac{(t_c-t_r)^2}{2\sigma^2}}\right) \nonumber
\end{equation}

\noindent $\gamma(t_r, t_c, \sigma)$ retrieves the weight factor from Eq.~\ref{eq:gauss} for reference timestamp $t_r$ and context timestamp $t_c$ for a given $\sigma$.

%where $W_l(O_i, O_j, t_r) = N_g(O_i,O_j)$. Here, the list $W_l$ is a list containing $\lvert{W_l}\rvert$ number rows representing the number of reference-context object pairs.

\subsection{Initial objective function}
\label{sec:simple_embedding_model}
We store all the reference-context object pairs and the corresponding discrepancy score in a list $D_s$ for static embeddings and $D_t$ for temporal embeddings. For static embeddings, each row of $D_{s}=\cup_{i=1}^{N} [o_r, o_c, \delta_{i}(o_r, o_c)]$ consists of a reference object, a context object, and the context discrepancy score. For temporal embedding, each row of $D_{t}=\cup_{i=1}^{N} [o_r, o_c, t_r, \delta_{i}(o_r, o_c, t_r)]$ has an additional column $t_r$ that represents the timestamp of the reference-context object pair. 

Our primary objective is to derive a low-dimensional object embedding model $E$ wherein the cosine distance between object vectors aligns closely with the discrepancy score. Equation~\ref{eq:theta_1_static} formulates the objective for static embedding as $\vartheta_{1 (static)}$. Here, we optimize the vectors in $E_s$ to reduce the difference between the cosine distance of each vector of reference-context object pair and the corresponding discrepancy score within $D_s$:
%or $\delta(O_r, O_c) = N_m(O_r,O_c)$. 
%Here, the list $W_l$ is a list containing $\lvert{W_l}\rvert$ number rows representing the number of center-context object pairs. 

\begin{equation}\label{eq:theta_1_static}
    \vartheta_{1 (static)}(E_s) = \sum_{i=1}^{N}\left(dist(e_{r}, e_{c}) - {{\delta}_i(o_r,o_c)}\right)^2
\end{equation}
Here, $dist(e_{r},e_{c})$ is the cosine distance between vector $e_{r}$ and
vector $e_{c}$.

For temporal embeddings, the objective function is delineated in
Equation~\ref{eq:theta_1_temporal}.  In this scenario, we optimize the vectors for reference and context objects at the reference object's timestamp within $E_t$. The goal is to minimize the difference between the cosine distance of each reference-context object pair at the reference timestamp and the corresponding discrepancy score present in list $D_t$ at timestamp $t_r$.

\begin{equation}\label{eq:theta_1_temporal}
    \vartheta_{1 (temporal)}(E_t) = \sum_{i=1}^{N}\left(dist(e_{r}^{t_r}, e_{c}^{t_r}, t_r) - {{\delta}_i(o_r,o_c,t_r)}\right)^2
\end{equation}

\noindent where $e_{r}^{t_r}$ is the embedding vector of reference object $o_r$ in timestamp $t_r$. 

Equation \ref{eq:theta_1_temporal} discretely selects the embedding vector $e_{r}^{t_r}$ of object $o_r$ in timestamp $t_r$ without considering the fact that the embeddings of $o_r$ in other timestamps might influence $e_{r}^{t_r}$. In the following subsection, we incorporate the impact of other embedding vectors of $o_r$ from timestamps other than $t_r$.

% The minimization of the above objective functions will ensure that our model captures the same distance as computed using Gaussian decay of Euclidean distances between object pairs.

\subsection{Incorporating temporal diffusion within the objective function}
\label{sec:temporal_weight}
One of the goals of obtaining temporal embeddings of objects in videos is to learn how the contexts of objects change over time. An analysis involving contexts can lead to the ability to explain a situation in an ongoing scenario in a video. To capture the contexts of an object from all timestamps, we need to generate embeddings that evolve smoothly over time. To introduce this concept in our objective function, we model the effect of every reference-context weight in all timestamps to some degree.

We use a Gaussian filter (Eq.~\ref{eq:gauss}) to \textit{diffuse} the contribution of each vector smoothly before and after the timestamp of the reference object. The filter considers its highest peak at the timestamp $t_r$ of the reference object $o_r$ with a decay before and after $t_r$. $\sigma$ is a user-settable parameter representing the standard deviation of the Gaussian distribution. 

Equation~\ref{eq:theta_2} presents the updated objective function, $\vartheta_2$, which includes the temporal weight of all the timestamps for $o_r$. It is important to emphasize that this adjusted objective function is exclusive to the temporal embedding model, given that static embeddings do not factor in time.

\begin{equation}
    \label{eq:theta_2}
    Loss1 =  \left(E1_{1}\cdot Vn_{2}  - Vn_{1}\cdot Vn_{2}\right)
\end{equation}

Equation~\ref{eq:theta_2} does not consider the frequency of an object's appearance in a timestamp (a timestamp may contain multiple frames), resulting in no variation between frequent and non-frequent objects. The following subsection incorporates the frequency.

\subsection{Influencing context via frequency of objects}
\label{sec:temporal_weight}
A crucial aspect in determining contextual temporal similarity among objects is the analysis of their frequency of occurrence. When a reference object and its surrounding entities consistently emerge within a given timestamp, their relational likelihood amplifies. However, an infrequent appearance of either or both objects hints at a lesser association between them during that period. To embed the frequency perspective into the training process, we infuse a frequency weight into the objective function.

We adopted multiple strategies into the objective function to smoothly integrate object frequency over the timestamps of the video data. Further insights and performance evaluations of these strategies are addressed in the experimental results section (Sec.~\ref{sec:experiments}). Initially, the frequency of each object at every timestamp is computed, which is then smoothed by normalizing with a Gaussian decay function, as illustrated earlier in equations~\ref{eq:gauss_decay} and ~\ref{eq:gauss_decay_norm}.

Given $f(o_k, t)$ is the frequency of an object $k$ at timestamp $t$, we convert the frequency to a nonlinear score via the Gaussian decay function (similar to Equation~\ref{eq:gauss_decay}). 
\begin{equation} \label{eq:gauss_decay_frq}
\begin{split}
\emph{G}(f(o_k,t)) &= \dfrac{1}{\sqrt{2\pi{\sigma}^2}}{e}^{\dfrac{-(f(o_k,t))^2}{2{\sigma}^2}}\\
\end{split}
\end{equation}
%, \text{where}, \emph{G}_\text{max} &= \dfrac{1}{\sqrt{2\pi{\sigma}^2}}
Analogous to Equation~\ref{eq:gauss_decay_norm}, the Normalized Gaussian decay function of frequency is given by:
\begin{equation}
N_g(f(o_k,t)) = \frac{\emph{G}(f(o_k,t))+\epsilon}{\max_{t=1}^{|T|}\emph{G}(f(o_k,t))+\epsilon}
\label{eq:gauss_decay_norm_frq}
\end{equation}
%The value of $N_g(f(o_k,t))$ ranges between $0$ and $1$, with $0$ indicating an object’s absence at time $t$ and $1$ representing the highest frequency.
The value of $N_g(f(o_k,t))$ ranges from a number close to $0$ to $1$. A value close to $0$ indicates an object's absence at time $t$, and $1$ represents the highest frequency. The inclusion of a small value $\epsilon$ (in this case, we used $\epsilon=0.01$) prevents division by zero errors when $N_g(f(o_k,t))$ is used as a denominator.

The normalized object frequencies (Eq. \ref{eq:gauss_decay_norm_frq}) are incorporated into the objective functions using several methods. Some of the variations of the objective function involve weighing via the average or minimum between the frequencies of reference and context objects. The average frequency is computed by:
\begin{equation}
\phi_{avg} = \frac{N_g(f(o_r,t_r))+N_g(f(o_c,t_r))}{2}
\end{equation}
The minimum frequency is represented as:
\begin{equation}
    \phi_{min} = \min(N_g(f(o_r,t_r)),N_g(f(o_c,t_r)))
\end{equation}

The objective functions described in the preceding sections aim to minimize the distances between object vectors based on their spatial and temporal proximity within frames and timestamps. The subsequent variations of the objective functions introduce the incorporation of frequency as a factor, facilitating the convergence of object vectors that exhibit high frequencies at a given timestamp.

\textbf{$\boldsymbol{\vartheta}_{\boldsymbol{3}}(\boldsymbol{E})$:} In this objective function, we introduce an approach where the frequency of the reference and context objects is summed and then multiplied by the cosine distance between the temporally diffused embeddings of the object pair. The underlying idea behind this objective function is that higher frequencies of both the reference and context objects will amplify the cosine distance, resulting in an elevated mean square error relative to the target value (discrepancy score). To counteract this effect, the optimizer aims to reduce the distance between the embeddings of frequently occurring objects at the corresponding timestamp.
    \begin{equation}\label{eq:theta_3}
    \begin{split}
        \vartheta_{3}(E) &=\sum_{i=1}^{N}(N_g(f(o_r,t_r))+N_g(f(o_c,t_r))\\
        &\quad \cdot (dist( \gamma(t_r, \sigma) \cdot e_{r}, \gamma(t_r, \sigma) \cdot e_{c}, t_r))-\delta_i(o_r,o_c,t_r))^2
    \end{split}
    \end{equation}

One notable characteristic of this objective function is its dependency on the discrepancy score for optimization. This means that even when two objects are frequent at a timestamp, if their discrepancy score is high, the optimizer will separate their embeddings to increase cosine distance in order to match it with the discrepancy score. Therefore, the frequency factor can only influences the optimization process when the discrepancy score is lower.
    
\textbf{$\boldsymbol{\vartheta}_{\boldsymbol{4}}(\boldsymbol{E})$:} In this particular objective function, we integrate frequency with the discrepancy score to give frequency a more substantial influence within the objective function. Here, we weigh the discrepancy score by the combined frequencies of the reference and context objects, subtracted from 2. The combined frequencies are subtracted from 2 because the total frequency of two objects can be a maximum of 2.0, each contributing to a maximum of 1.0. This design ensures that object pairs with higher frequencies receive a lower weight for the discrepancy score, whereas those with lower frequencies will be assigned a higher weight. Consequently, high-frequency object pairs will exhibit lower target values, leading the optimizer to reduce the cosine distance for these high-frequency objects and bring their embedding vectors closer.

    \begin{equation}\label{eq:theta_4}
    \begin{split}
        \vartheta_{4}(E) &=\sum_{i=1}^{N}(
    dist( \gamma(t_r, \sigma) \cdot e_{r}, \gamma(t_r, \sigma) \cdot e_{c}, t_r)\\
        &\quad -(2-N_g(f(o_r,t_r))+N_g(f(o_c,t_r)))\cdot \delta_i(o_r,o_c,t_r)
        )^2
    \end{split}
    \end{equation}
A limitation of this objective function is that the weight is not normalized to a range of 0 to 1 but spans from 0 to 2. Consequently, the multiplication with the discrepancy score can yield a value exceeding 1. However, it is desirable to keep the target value within the range of 0 to 1, as the cosine distance varies from 0 (identical vectors) to 1 (no correlation) in the positive mathematical space.
    
\textbf{$\boldsymbol{\vartheta}_{\boldsymbol{5}}(\boldsymbol{E})$:} In this objective function, the target value is weighted based on the average of the frequencies of reference and context objects, subtracted from 1. While the aim aligns with the prior objective function in assigning lower weights to higher-frequency object pairs and higher weights to lower-frequency pairs, the weight in this function is strictly confined between 0 and 1.
    
    \begin{equation}\label{eq:theta_5}
    \begin{split}
        \vartheta_{5}(E) &=\sum_{i=1}^{N}(dist( \gamma(t_r, \sigma) \cdot e_{r}, \gamma(t_r, \sigma) \cdot e_{c}, t_r)\\
        &\quad -(1-\phi_{avg})\cdot \delta_i(o_r,o_c,t_r))^2
    \end{split}
    \end{equation}

A drawback of using the average frequency is that it can not distinguish between objects with vastly different frequencies and those with similar frequencies. When one object has a very high frequency and the other has a very low frequency, their average falls in the middle. Similarly, when both objects have medium frequencies, the average is also medium, making it challenging to differentiate between dissimilar and similar-frequency objects.
    
\textbf{$\boldsymbol{\vartheta}_{\boldsymbol{6}}(\boldsymbol{E})$ and $\boldsymbol{\vartheta}_{\boldsymbol{7}}(\boldsymbol{E})$:} In the following objective function, rather than averaging the frequencies of the object pair to weigh the target value, we adopt the minimum frequency between the two. This approach ensures a higher weight when either object has a low frequency. The weight decreases only when both objects exhibit high frequencies, resulting in a reduction of the target value. The following two variations of the objective function emphasize the minimum frequency of the object pair.
    \begin{equation}\label{eq:theta_6}
    \begin{split}
         \vartheta_{6}(E) &=\sum_{i=1}^{N}(dist( \gamma(t_r, \sigma) \cdot e_{r}, \gamma(t_r, \sigma) \cdot e_{c}, t_r)\\
        &\quad -(1- \phi_{min})\cdot \delta_i(o_r,o_c,t_r))^2
    \end{split}
    \end{equation}
 
    \begin{equation}\label{eq:theta_7}
    \begin{split}
        \vartheta_{7}(E) &=\sum_{i=1}^{N}(dist( \gamma(t_r, \sigma) \cdot e_{r}, \gamma(t_r, \sigma) \cdot e_{c}, t_r)\\
        &\quad -(1-\frac{\phi_{min}}{2})\cdot \delta_i(o_r,o_c,t_r))^2
    \end{split}
    \end{equation}

\textbf{$\boldsymbol{\vartheta}_{\boldsymbol{8}}(\boldsymbol{E})$:} In this objective function, we employ a non-linear approach to the minimum frequency. Rather than directly using the minimum frequency, we calculate its natural logarithm (base e) and then multiply the result by 2. This non-linear transformation assigns a higher weight when the minimum frequency is low and a substantially lower weight when the minimum frequency is high.
    \begin{equation}\label{eq:theta_8}
    \begin{split}
        \vartheta_{8}(E) &=\sum_{i=1}^{N}(dist(\gamma(t_r, \sigma) \cdot e_{r}, \gamma(t_r, \sigma) \cdot e_{c}, t_r)\\
        &\quad -(2*\ln(\frac{1.5}{\phi_{min}})\cdot \delta_i(o_r,o_c,t_r)))^2
    \end{split}
    \end{equation}

\textbf{$\boldsymbol{\vartheta}_{\boldsymbol{9}}(\boldsymbol{E})$:} This variation of the objective function is a modified version of the previous one, where we utilize the average frequency in place of the minimum frequency. Additionally, the frequency weight is conditioned based on the average frequency value. If the average frequency exceeds $0.5$, the log-based weight is directly multiplied by the target value (discrepancy score). However, for an average frequency of $0.5$ or less, the log-based weight is doubled, thereby escalating the weight for less frequently paired objects.

    The log-based weight on frequency is described as,

    \begin{equation}\label{eq:omega}
     \begin{split}
        \omega_{\ln} &= \begin{cases}
       \ln(\frac{1.5}{\phi_{avg}}) &\quad\text{; when } \phi_{avg}> 0.5\\
       2*\ln(\frac{1.5}{\phi_{avg}}) &\quad\text{; when } \phi_{avg}<=0.5
     \end{cases} 
     \end{split}
    \end{equation}
    \begin{equation}\label{eq:theta_9}
    \begin{split}
        \vartheta_{9}(E) &=\sum_{i=1}^{N}(dist( \gamma(t_r, \sigma) \cdot e_{r}, \gamma(t_r, \sigma) \cdot e_{c}, t_r)\\
        &\quad -(\omega_{\ln}*\cdot \delta_i(o_r,o_c,t_r)))^2
    \end{split}
    \end{equation}

These objective functions cannot provide an ultimate solution; they highlight different analytical aspects depending on the application the generated embeddings will be used for. The experimental results section (Section \ref{sec:experiments}) provides further details on the resultant outcomes of these objective functions.

  % \item Frequency weight on cosine distance of embeddings.
  %   \begin{equation}\label{eq:theta_4}
  %       \begin{split}    
  %       \vartheta_{4}(\mathcal{U}) = \sum_{k=1}^{\lvert{W_l}\rvert}\sum_{p=-wt}^{wt}\left( \frac{f(t_r,o_i)+f(t_r,o_j)}{2} \cdot \\ dist(\gamma(t_{(r+p)}, \sigma) \cdot u_{i}, \gamma(t_{(r+p)}, \sigma) \cdot u_{j}, t_{(r+p)}) \\- {{W_l}_k(o_i,o_j,t_r)}\right)^2
  %       \end{split}
  %   \end{equation}
  % \item Modified frequency weight. 
  
  % Given, $E(f(t,o_i))=\frac{1+e^{f(t,o_i)}}{0.25+e^{f(t,o_i)}}$
  %   \begin{equation}\label{eq:theta_5}
  %       \begin{split}    
  %       \vartheta_{4}(\mathcal{U}) = \sum_{k=1}^{\lvert{W_l}\rvert}\sum_{p=-wt}^{wt}\\ \left(dist(\gamma(t_{(r+p)}, \sigma) \cdot u_{i}, \gamma(t_{(r+p)}, \sigma) \cdot u_{j}, t_{(r+p)}) \\ E(f(t_{(r+p)},o_i))\cdot E(f(t_{(r+p)},o_j))- {{W_l}_k(o_i,o_j,t_r)}\right)^2
  %       \end{split}
  %   \end{equation}

  %       \begin{equation}\label{eq:theta_5}
  %       \begin{split}    
  %       \vartheta_{4}(\mathcal{U}) = \sum_{k=1}^{\lvert{W_l}\rvert}\sum_{p=-wt}^{wt}\\ \left(dist(\gamma(t_{(r+p)}, \sigma) \cdot u_{i}, \gamma(t_{(r+p)}, \sigma) \cdot u_{j}, t_{(r+p)}) \\ -max(E(f(t_{(r+p)},o_i)), E(f(t_{(r+p)},o_j))) {{W_l}_k(o_i,o_j,t_r)}\right)^2
  %       \end{split}
  %   \end{equation}

\subsection{Incorporating negative context}
\label{sec:negative_sampling}

%where transitions from one frame to another already captures negative samples, Due to the lack of labeled video data, we have analyzed the output of our model on image dataset as well, where sequential images have no relation like video frames. As the sequence of images has no impact on context so we disregard context from sequential images and just take context from each individual images only. After analyzing the outcome, we found, only positive relations between the context objects are not working better to establish contextual relation through embedding vector. Thus we introduces some non-contextual relation by using the hypothesis that, objects in different images and never seen together in one image, have less contextual connection than the objects present in the same images. 

%We use the concept of negative  as utilized in word2vec~\cite{Mikolov13b}. In our negative  approach, for every unique target objects $o^r$, we randomly select
%$n_{\text{negative samples}} = 1/4 /times \text{Total Unique Objects}$
%\noindent and set the target and negative sample pair contextual strength as $\text{Negative}_d(o^r,o^{ns}) = 0$.
In order to increase the separation between embedding vectors of non-context objects in the embedding space, we introduced negative relationships between object pairs that lack proximity-based connections. These negative relationships involve objects that are not found in the reference frame (for static embedding on image data), surrounding frames (for static embedding on video data), or the reference timestamp (for temporal embedding on video data). Negative objects are randomly selected from the pool of objects appearing outside the context window. This approach draws inspiration from negative sampling techniques used in word2vec~\cite{Mikolov13b}. However, we employ the cumulative frequency of objects at each timestamp to make weighted random selections from the objects located outside the context window. This enables us to choose the negative context for a reference object from those that occur more frequently outside the reference object's context window. The probability of selecting the $i^{th}$ object from a list is given by:

\begin{equation}
\begin{split}
    f(o_i) &= \sum_{t=1}^{|T|}f(o_i,t)\\
     P(i) &= \frac{f(o_i)-f(o_i,t_r)}{\sum_{j=1}^{|O|}(f(o_j)-f(o_j,t_r))}
\end{split}
\end{equation}

\subsection{Embedding generation using a neural network model}
\begin{figure}[!ht]
\begin{center}
  \includegraphics[width=0.9\linewidth]{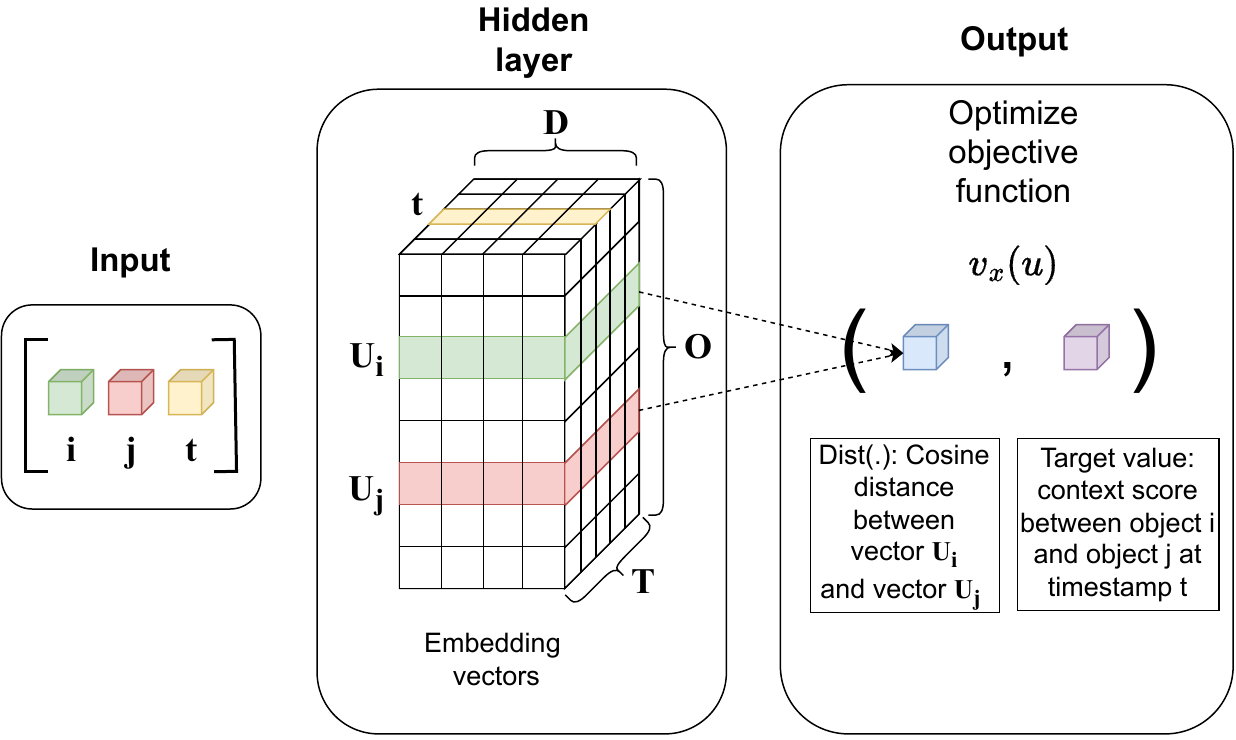}
\end{center}
\caption{Neural network model to generate the temporal visual object embeddings.}
\label{fig:temporal_embedding_nn}
\end{figure}

We implemented a neural network-based model using Tensorflow to generate our \textbf{temporal object embeddings}. An overall view of the architecture of our neural network is shown in figure~\ref{fig:temporal_embedding_nn}. The goal of the neural network is to optimize the objective functions detailed in the previous subsection. The embeddings for all objects in all timestamps are generated in the hidden layer. We initialize the weights in the hidden layer in the range [0, 1].

The neural network needs three inputs: the reference object ($o_i$), the context object($o_j$),  and the timestamp($t_r$), along with one target value representing the discrepancy score. The hidden layer has a size of $|O|*|T|*|e|$, where each embedding vector corresponds to an object at a specific timestamp.  The target value reflects the expected cosine distance between $o_i$ and $o_j$ at time $t_r$ calculated using the discrepancy score discussed in Sec.~\ref{sec:context_Strength}.

In the case of static embedding, the input layer does not have any timestamp $t_r$. Consequently, the hidden layer is of size $|O|*|e|$ with each embedding vector of size $|e|$ representing an object.
%The training occurs for every target object in every frame of video $F$.

\subsection{Integration of Visual Features and Temporal Contextual Embedding}
\begin{figure}[h]
\begin{center}
  \includegraphics[width=0.7\linewidth]{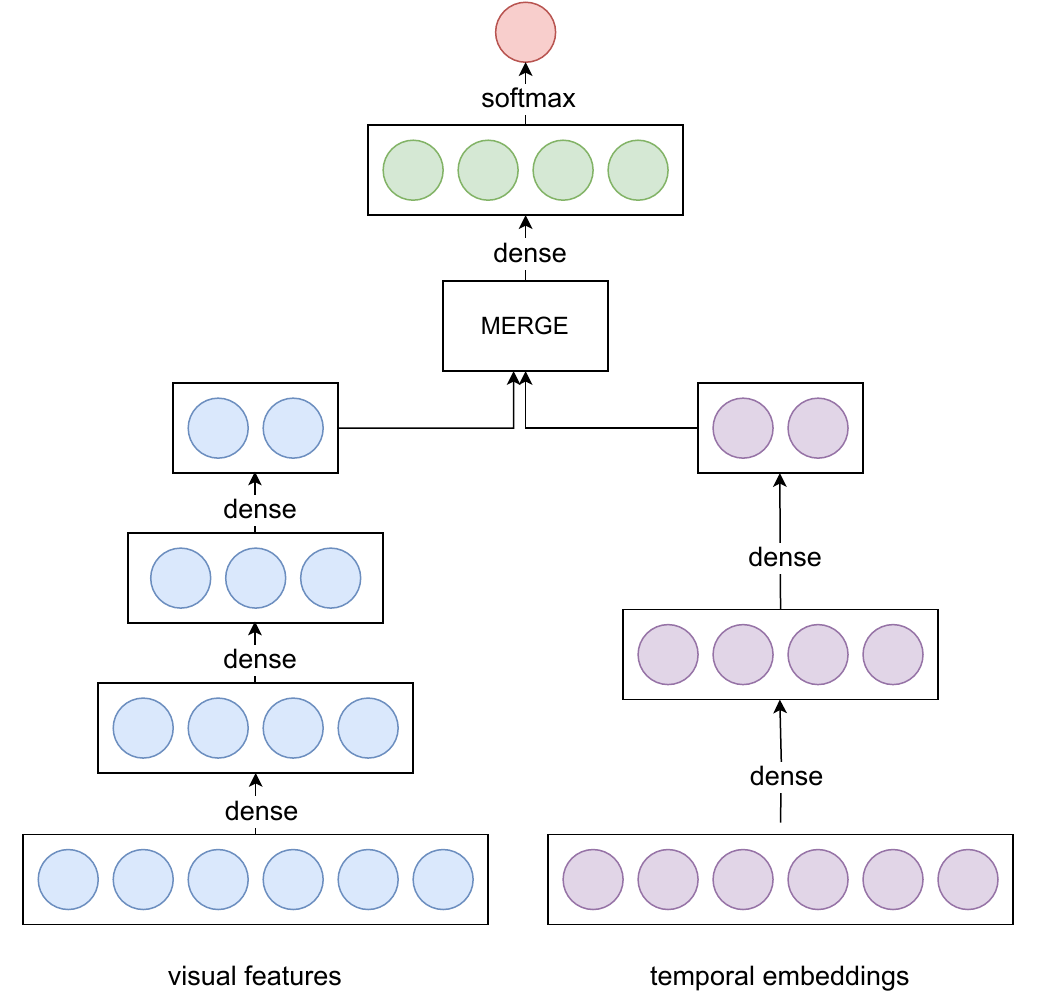}
\end{center}
\caption{Fusion of visual features (using either CNN or ResNet50) and temporal contextual embeddings.}
\label{fig:fusionEmbedding}
\end{figure}
So far, in our objective functions, we have not considered the visual features of the objects. Visual features, along with contextual temporal embeddings, have immense potential in downstream applications (later demonstrated in the experimental results section). The current section describes the fusion of visual features and the temporal contextual embeddings produced by our neural network of Figure \ref{fig:temporal_embedding_nn}. 

We used two methods to obtain the visual features of detected objects in video frames.

\begin{enumerate}
\item \textbf{CNN Features}: We designed a custom CNN model with three convolutional layers, each succeeded by a max pooling layer and ReLU activation. After the convolutional stages, the extracted features are processed through dense layers optimized for a contextual classification task as a downstream application.

\item \textbf{ResNet50 Features}: We leverage the widely recognized pre-trained model, ResNet50 to extract visual features from detected objects. Following extraction, these features are channeled through dense layers that are specifically configured for the purpose of the same contextual classification problem associated with the CNN features.
\end{enumerate}

Fig~\ref{fig:fusionEmbedding} illustrates a simplified diagram of the fusion process. The goal is to integrate visual features extracted from either CNN or ResNet50 models with temporal contextual object embeddings to increase the performance of the targeted downstream application. In this paper, we selected a downstream application of contextual classification. Therefore, the fusion is tuned for improved classification accuracy. The temporal contextual embeddings are three-dimensional, characterized by the dimensions $|O|\times|T|\times|e|$, where $|O|$ indicates the number of objects, $T$ represents the timestamps, and $e$ denotes the dimensions of the embeddings. To facilitate the fusion process, we convert these embeddings into a two-dimensional format, $|O|\times[|T|\times|e|]$, where each object is represented by a $|T|\times|e|$ sized vector. These temporal contextual embeddings are processed through several dense layers to derive a refined representation comparable to the visual representation length. In parallel, visual features from the CNN or ResNet50 model are channeled through multiple dense layers to achieve their respective representations. 

We have experimented with various methods for merging visual features with contextual embeddings (the MERGE box in Fig~\ref{fig:fusionEmbedding}), including feature concatenation, concatenation post-PCA, attention-based concatenation, bilinear pooling, graph convolution, and gated feature integration for the fusion of the visual and temporal contextual representations. 

%The performance of each method is evaluated in the experimental results section~\ref{sec:exp:contextual_classification}.

%In our experiments, when we used motion, the user-settable parameter ($n$) to indicate number of frames prior and after the current frame $f^r$ was set to 10, resulting in a window size of 21. Without the use of motion (when the camera moves steadily), we set the value to 5, resulting in a windows-size of 11. The equation with motion automatically diminishes the importance of similar frames in the window and hence a larger window-size is appropriate.

\section{Datasets}\label{sec:dataset}
Benchmark datasets for object embedding in video contexts are relatively undeveloped, primarily because this is an emerging area of research. Commonly available benchmark datasets with object annotations are primarily intended for object recognition or detection. In our research of generating static or temporal embeddings of video data, the neural network is trained to learn embedding vectors for detected objects in video data. To the best of our knowledge, no dataset has been specifically crafted to evaluate the learning mechanism of object embeddings. Hence, for this study, we created our own annotated video datasets. We utilized an existing object detection model to detect objects and label them. In addition, for one video, we manually annotated objects frame by frame. 

As summarized in Table~\ref{tab:datasetList}, our research utilized four distinct types of video data: (1) synthetic, (2) videos captured using our camera, (3)  videos downloaded from YouTube, and (4) renowned annotated datasets such as COCO~\cite{Lin14} and LabelMe~\cite{yuen09}. 

\begin{table}[!h]
\caption{Dataset List.}\label{tab:datasetList}%
\begin{tabular}{@{}llllll@{}}
\toprule
Name  & \# Frames  & \# Unique Objects  & \# Instances  & Label Annotator\\
\midrule
Synthetic 1 & 20,000 & 62 & 158,000 & Inherent with the process  \\
Synthetic 2 & 3,000 & 40 & 32,060 & Inherent with the process  \\
Shot video & 494 & 93 & 1,330 & Human  \\
Youtube\footnotemark[1] & 10,125 & 63 & 27,419 & Yolo9000  \\
LabelMe\footnotemark[2] & 333,243 & 58 & 23,730 & YoloV4 \\
COCO\footnotemark[3] & 117,266 & 80 & 800,308 & COCO\\
\botrule
\end{tabular}
\footnotetext{About some of the sources:}
\footnotetext[1]{A list of modular home videos: \url{https://bit.ly/2YhGuUL}}
\footnotetext[2]{The videos were sourced from the public collections of sequential frames available on the website http://labelme2.csail.mit.edu~\cite{yuen09}}
\footnotetext[3]{Common Objects in Context (COCO) labeled images~\cite{Lin14}}

\end{table}

The synthetic videos were created to test the capability of our models in accurately retrieving known contexts. The annotations of the objects were already known from the process of creating the data. The video recorded with our camera was manually annotated (named as Shot video in Table \ref{tab:datasetList}). Given the extensive length of the downloaded YouTube videos, we relied on YOLO~\cite{Redmon16, Bochkovskiy20} for object detection (detection threshold = 0.2) in each frame. 

Although our proposed model is primarily designed for video data, we have conducted additional experiments using images to evaluate the context-capturing capability of the generated embeddings. For these image-based experiments, we used annotations from the COCO dataset. Our results are compared with those of an existing model that focuses on the contextual embedding of visual objects in images \cite{LUDDECKE19}.

\section{Experiments}\label{sec:experiments}
In the problem description (Section~\ref{sec:problem_description}), we outlined a two-step process for generating temporal contextual object embeddings. Initially, we create contextual object embeddings, which capture the context of each detected object in a video. In this paper, we refer to this model as the static contextual object embedding model, as this model represents each object with a single vector. Following this, we incorporate a time dimension to develop temporal contextual object embeddings. This temporal model produces an embedding vector for each object at every timestamp. The temporal model is designed not only to capture the context of objects but also to capture changes in objects' context over time. In this experimental analysis section, we evaluate the effectiveness of both static and temporal models with steps involved in them. The experimental results are divided into two Subsections (Subsection \ref{sec:exp_static} and \ref{sec:exp_temporal}), with one focusing on evaluating the static contextual embedding model and the other on assessing the temporal contextual embedding model.

\subsection{Evaluation of the Static Embedding Model}\label{sec:exp_static}
This subsection seeks to answer the following questions relevant to static contextual embeddings.
\begin{enumerate}
\item How well do the static embeddings capture the context when video frames are considered independent images? (Subsection~\ref{sec:clusterObjOneFrame})
%\item How well can the generated object vectors predict unseen relationships through the contextual connection? Subsection~\ref{sec:clusterObjOneFrame}
\item How well do static embeddings capture the context of objects when a video is considered a sequence of frames? (Subsection~\ref{sec:surroundingContext})
\item How well do vectors from static embedding cluster visual objects in images compared to a state-of-the-art model? (Subsection~\ref{sec:staticImages})
%\item What is the impact of incorporating motion in static embedding for video data? (Subsection~\ref{sec:motionExperiment}) \afarhan{incorporating motion methodology is not added}
\item Case study: How well do the generated vectors of visual objects in video data agree with a language model? (Section~\ref{sec:case1})
\item Case study: How can static embeddings be used to analyze the context of visual objects in a video? (Subsection~\ref{sec:case2})
\item Case study: How can the mathematical space generated by static embeddings be used to analyze a video? (Subsection~\ref{sec:case3})
\end{enumerate}

\subsubsection{Embedding with reference-context pair in the reference frame}
\label{sec:clusterObjOneFrame}
%In this section, we present the experiments performed using three different datasets. For the first experiment, we generated a synthetic dataset containing 62 different objects. The images of the objects are taken from Emnist dataset containing 26 images of A-Z, 26 images of a-z and 10 images of 0-9. Video frames are generated using these three sets of object images where objects from same set are grouped together (Fig.~\ref{fig:syntheticData1}).

%For the rest of the experiments, we utilized some sample videos consisting of home tour videos and city tour videos. The home tour video contains 100 different objects where most objects are home accessories such as chair, table, sofa, etc. And the city tour video contains 60 different objects where most objects are seen around roads such as car, truck, signal, umbrella, person, etc. As the concept of vector representation of visual objects from videos is novel and no previous work was done in this area, we have created our baseline model to compare the performance of different objective functions. For the baseline model, the nearest neighbor of any target object is calculated by computing the distances of objects that appear in the same video frame or the surrounding frames with the target object. Therefore, the baseline can provide us with the neighbors that actually seen around the target object but latent information like, objects that never appear in the same frame but share the same neighbor is missing in the baseline.

\begin{figure}[!h]
\centering
\begin{subfigure}[b]{.32\textwidth}
  \centering
  \fbox{\includegraphics[width=.5\textwidth]{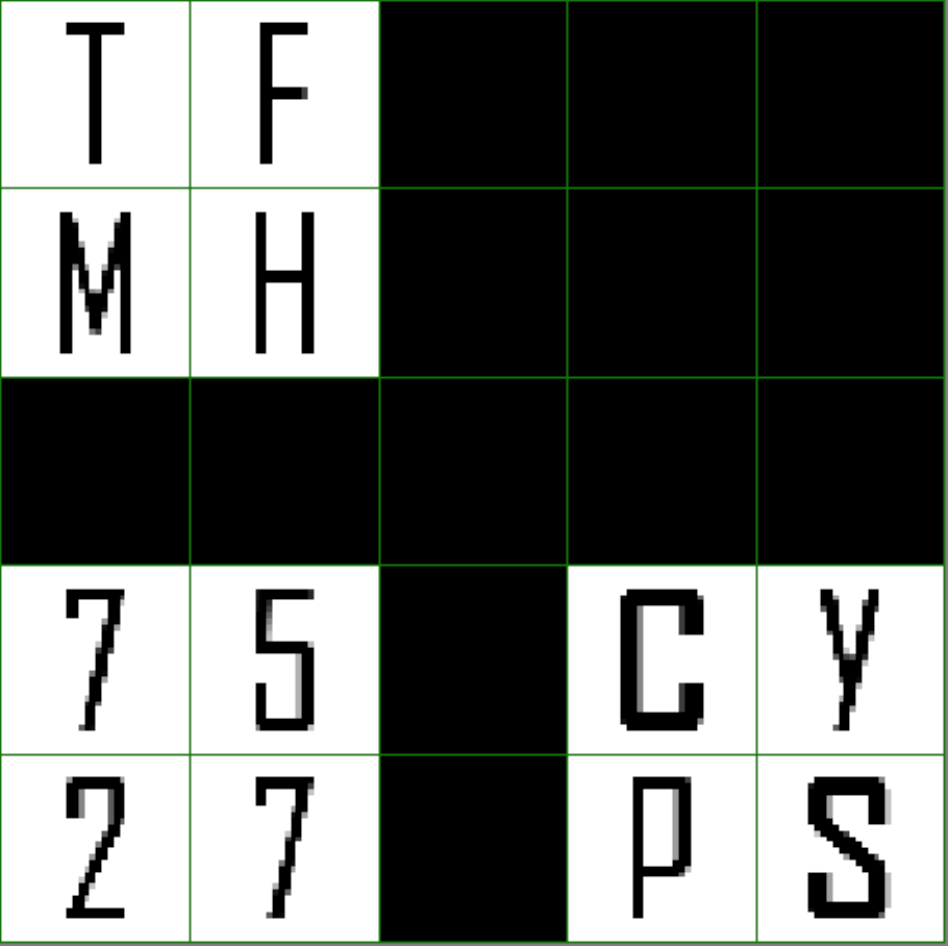}}
  %\caption{Frame 1}
\end{subfigure}%
\begin{subfigure}[b]{.32\textwidth}
  \centering
  \fbox{\includegraphics[width=.5\textwidth]{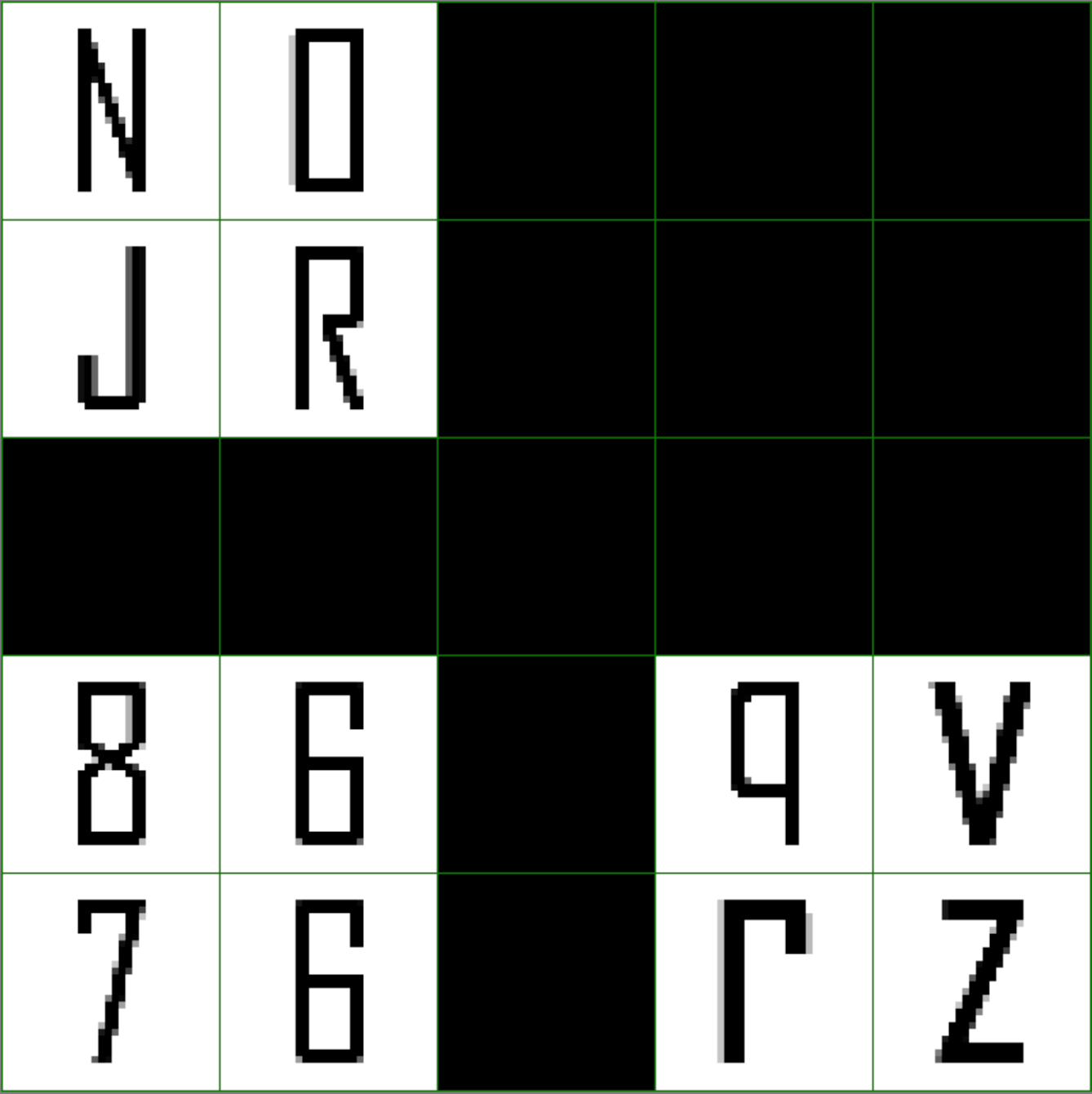}}
  %\caption{Frame 2}
\end{subfigure}
\begin{subfigure}[b]{.32\textwidth}
  \centering
  \fbox{\includegraphics[width=.5\textwidth]{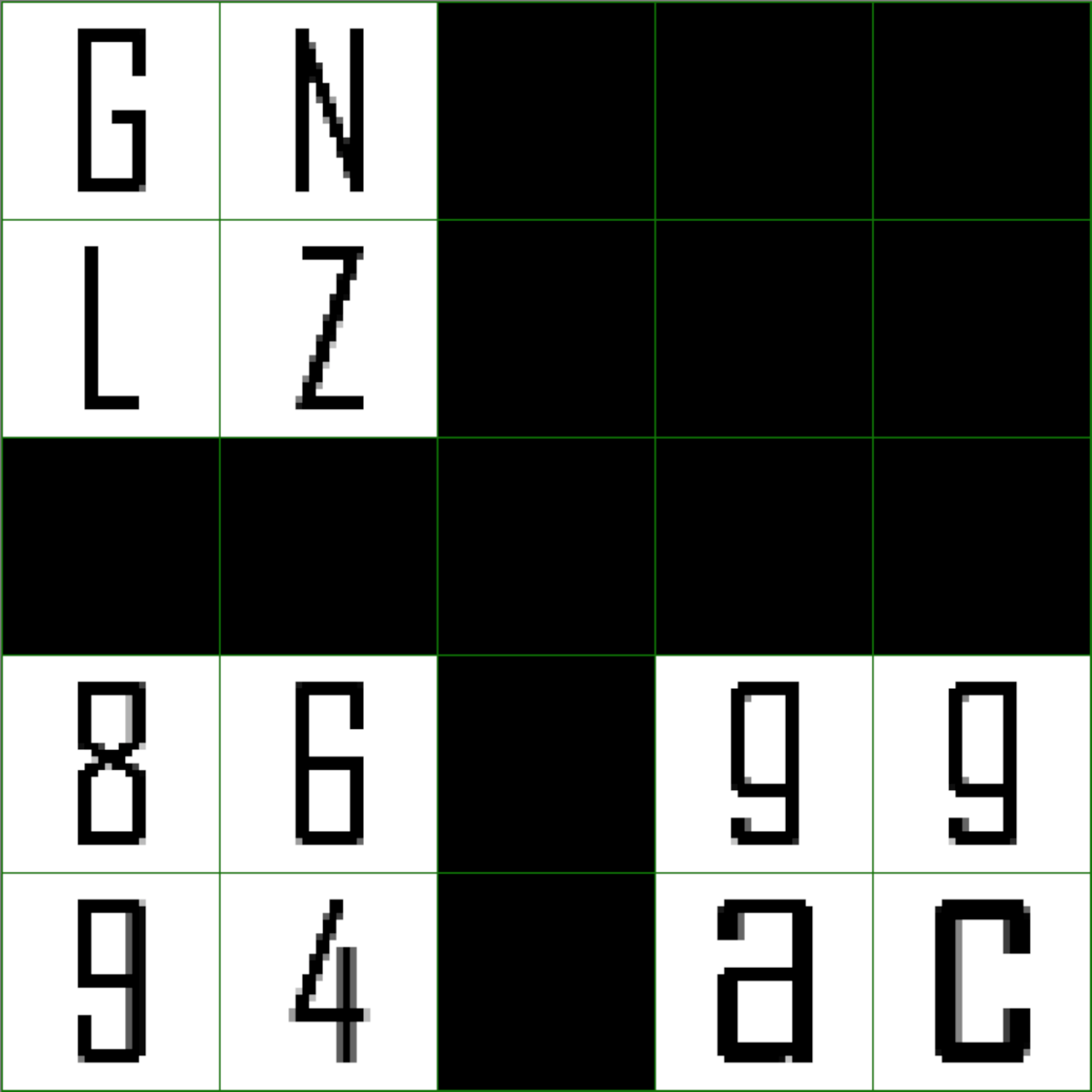}}
  %\caption{Frame 3}
\end{subfigure}

\caption{Synthetic Dataset with $5\times 5$ grids.}
\label{fig:syntheticData1}
\end{figure}

This experiment evaluates the effectiveness of our static embedding model in clustering objects that are concurrently present in the same video frame in close proximity. In this approach, each video frame is considered an independent image, with object contexts being derived exclusively from that specific frame, excluding context objects from surrounding frames (Context 1 in Section~\ref{sec:context_selection}). To conduct this study, we generated a synthetic video containing 10,000 frames, each containing a grid of $5 \times 5 = 25$ squares (Fig.~\ref{fig:syntheticData1}). Among these square grids, 12 grids contain images of letters or digits from three distinct sets  -- \{A to Z\}, \{a to z\}, and \{0 to 9\}. Notably, objects from the same set consistently appear close to one another in groups of four side-by-side grids in all the frames.
%Each grid contains a letter or digit-image from the three following sets 
%We create three vizObj2Vec models
We employ three of our reference-context scoring approaches (Equations~\ref{eq:threshold},~\ref{eq:min_max_norm}, and~\ref{eq:gauss_decay_norm}) to train the static embedding model~\ref{eq:spatial_distance}. Each training iteration produces a set of embedding vectors for all objects. Given that each frame contains characters from three distinct sets, we utilize \textit{k}-means clustering with \textit{k}=3 to assess the efficacy of our embeddings in separating these three clusters.

%In this experiment with synthetic data, we study the quality of object vectors to cluster the objects present in the same frame. Each frame in synthetic data (Fig.~\ref{fig:syntheticData1}) contains $5 \times 5 = 25$ square boxes where each box contains either a image from 3 sets of objects (A-Z, a-z, 0-9) or remain blank (black pixels). Also, the objects from same set always appears side by side. For the experiment, we have utilized different context strength selection methods to generate the object vectors and compare the cluster quality of generated vectors. Applying the K-means cluster (where k=3) over the object vectors, we get a 100\% cluster match for embedding vectors from all the methods. In Fig.~\ref{fig:silhoetteComp}, we compared the silhouette score of K-means cluster for different methods using different hyper-parameters.

\begin{figure}[!h]
  \centering
  \includegraphics[width=0.7\textwidth]{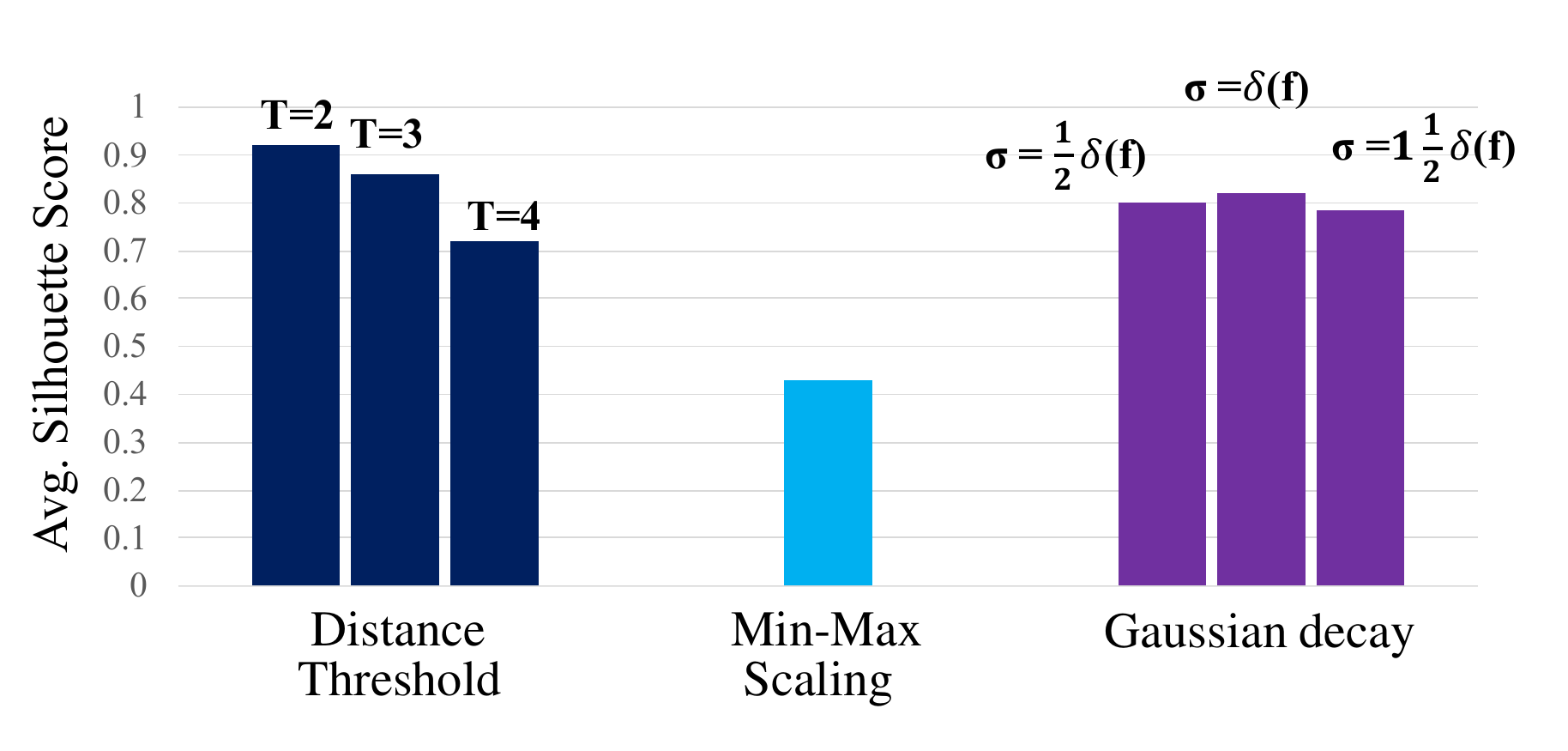}
  \caption{Comparison of average silhouette score for three clusters using three variations of context scores in the static embedding model.}
  \label{fig:silhoetteComp}
\end{figure}

Across all reference-context scoring methods, the clustering outcomes, utilizing various hyperparameters of the algorithms, demonstrate positive average silhouette coefficients (ASC). Positive ASC indicates good structure in the clusters in the embedding space. Fig.~\ref{fig:silhoetteComp} demonstrates the average silhouette coefficients. Notice that the method with min-max scaling (Eq.~\ref{eq:min_max_norm}) does not have any hyperparameter. As a result, only one bar is provided with the min-max scaling method in Fig.~\ref{fig:silhoetteComp}. 

Even though the average silhouette coefficient values vary between approaches, we observed that all the clustering results had a Rand index of 1.0, indicating 100\% accuracy in clustering the three classes. Hence, incorporating the spatial distance between objects into our objective functions successfully captures the contextual relationships of objects in discrete images.

The experimental analysis with synthetic data shows promising results in clustering objects in proximity in the synthetic frame. However, the dynamics of object interactions in actual video footage are less structured than synthetic data. To evaluate the performance of our model in clustering neighboring objects within real-world dataset, we evaluated the neighborhood intersections between the true neighboring objects in video frames and the neighbors in our generated embedding space. For this experiment, we trained our static object embedding using a publicly accessible YouTube video\footnote{A list of home videos: \url{https://bit.ly/2YhGuUL}}.
%\\url{https://www.youtube.com//playlist?list=PLN7zBw9BH5TrhZssCM8XH9EBNcrnT-d-E}}. 
\begin{figure}[!h]
  \centering
  \includegraphics[width=0.6\linewidth]{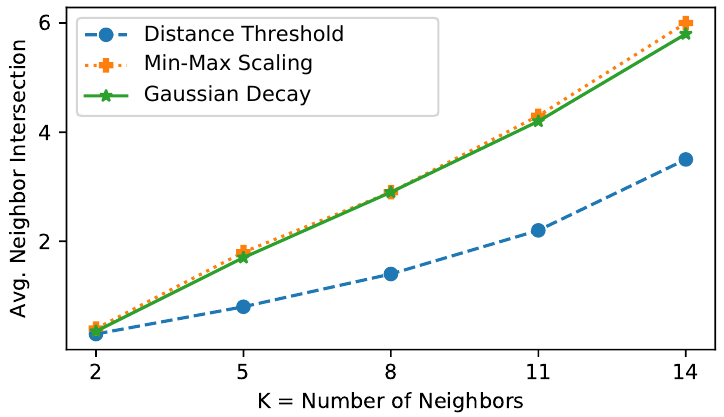}
  \caption{Number of same-frame objects detected  
  in the $k$ neighborhood of the embedding space.}
\label{fig:avgNeighInterHomeTour}
\end{figure}

We employed three reference-context scoring techniques (Equations~\ref{eq:threshold},~\ref{eq:min_max_norm}, and~\ref{eq:gauss_decay_norm}) to generate the static embeddings. Given the nearest neighbor of each objects in each frames are known, we calculated the average intersection of top \textit{k} nearest neighbors of each objects in the embedding space. Fig.~\ref{fig:avgNeighInterHomeTour} shows that the Gaussian decay (Eq.~\ref{eq:gauss_decay_norm}) and min-max scaling (Eq.~\ref{eq:min_max_norm}) methods perform competitively. Both methods are significantly better than the threshold-based approach in detecting neighbor objects in discrete frames. In Figure~\ref{fig:silhoetteComp}, we observed that the Gaussian decay function-based embedding model provides stable results in average silhouette score without much requirement of threshold-based hyperparameters. Additionally, as shown in Figure~\ref{fig:silhoetteComp}, the silhouette score of clusters originating from embeddings trained with Gaussian decay normalization is notably higher compared to those derived from embeddings trained using min-max scaling normalization. This suggests that embeddings derived from Gaussian decay normalized distances yield higher-quality clustering. This is why in the subsequent experimental analysis, we use Gaussian decay-based reference-context discrepancy score (Eq.~\ref{eq:gauss_decay_norm}) for the static embedding model.

%The previous experiment shows promising results in clustering objects that appear nearby in the synthetic frames. However, the scenario is not the same in real-world datasets. To evaluate the performance of our visualObj2vec model in the real-world dataset to track nearby seen objects, we compare the neighborhoods obtained by our method with the baseline neighborhoods. Fig.~\ref{fig:avgNeighInterHomeTour} presents the average number of objects intersections between the neighborhoods of size k obtained using the baseline model and those obtained using the proposed model.
\begin{figure}[!hb]
  \centering
  \includegraphics[width=0.7\linewidth]{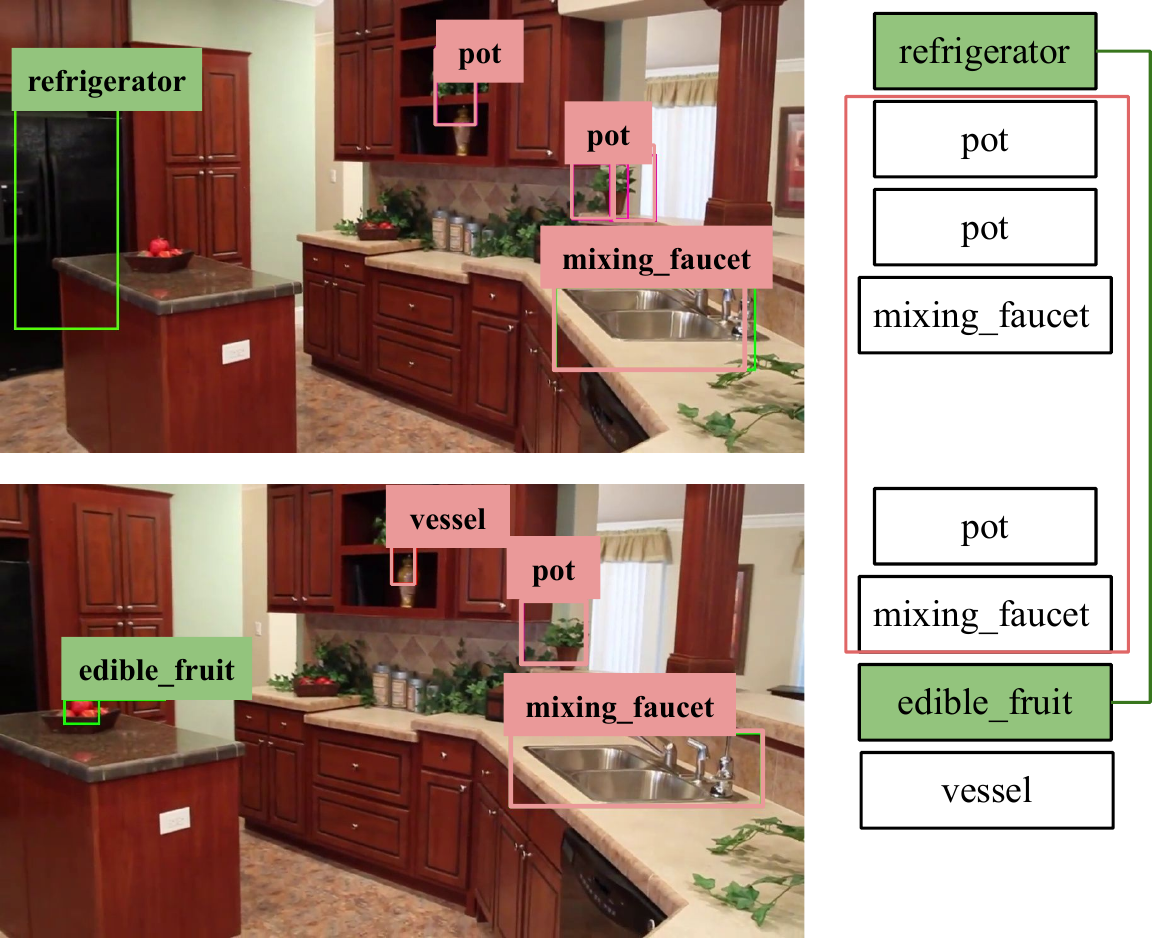}
  \caption{``refrigerator'' and ``mixing\_faucet'' detected in one frame by YOLO. ``mixing\_faucet'' and ``edible\_fruit'' detected in another frame. Our static embedding model brings all three objects in a neighborhood of the embedding space.}
  \label{fig:contextualSim}
\end{figure}

Our static embedding model can capture contextual connections between objects that do not co-occur within the same frame but are linked by common contextual objects. Take, for instance, the scenario depicted in Fig.~\ref{fig:contextualSim} where one frame captures a refrigerator, a pot, and a mixing faucet, while the subsequent frame shows an edible fruit, a pot, and a mixing faucet. Notably, in our dataset, the refrigerator and the edible fruit never detected together in any frame. However, upon examining the nearest neighbors through our static embedding, ``edible\_fruit'' emerges as one of the top eight nearest neighbors of ``refrigerator,'' bearing a cosine similarity of 0.65. This implies that our model successfully associates ``refrigerator'' and ``edible\_fruit'' within the embedding space's vicinity due to their shared context involving ``mixing\_faucet'' and ``pot,'' although they are not concurrently detected in any frame. Such findings underscore our model's capability to encapsulate the contextual relationships among objects that appear across separate frames despite we trained the static embedding model by reference-context object pairs from distinct frames.

% We found ``edible\_fruit'' in the eight-nearest neighbor list of ``refrigerator'', with a cosine similarity of 0.65, indicating that vizObj2Vec was able to include objects from other frames as context.

\subsubsection{Impact of surrounding frames for context selection}
\label{sec:surroundingContext}

\begin{figure}[!h]
\centering
\begin{subfigure}[b]{.24\textwidth}
  \centering
  \fbox{\includegraphics[width=.5\textwidth]{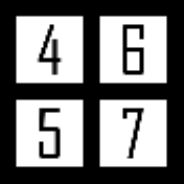}}
  \caption{Frame 2}
\end{subfigure}%
\begin{subfigure}[b]{.24\textwidth}
  \centering
  \fbox{\includegraphics[width=.5\textwidth]{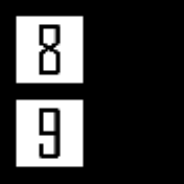}}
  \caption{Frame 3}
\end{subfigure}
\begin{subfigure}[b]{.24\textwidth}
  \centering
  \fbox{\includegraphics[width=.5\textwidth]{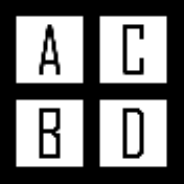}}
  \caption{Frame 4}
\end{subfigure}
\begin{subfigure}[b]{.24\textwidth}
  \centering
  \fbox{\includegraphics[width=.5\textwidth]{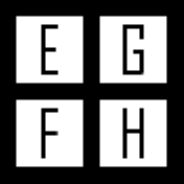}}
  \caption{Frame 5}
\end{subfigure}
\caption{Synthetic dataset with four grids. }
\label{fig:syntheticData2}
\end{figure}

This experiment assesses the capacity of our static embedding model to cluster objects that co-occur across consecutive video frames. In this approach, the model considers object contexts from both the reference and surrounding frames (Context 2 in Section~\ref{sec:context_selection}). To study the impact of incorporating surrounding frames in context construction, we created a synthetic video that incorporated letter and digit images from the Chars74K dataset~\cite{Campos09}. Each frame in the video had space to contain four letters or digits. A frame may contain no more than four consecutive letters from exactly one of the following sets: \{0 to 9\}, \{A to Z\}, and \{a to z\}. We created the consecutive frames in such a way that, four consecutive objects are picked sequentially from the three sets. That is, the content of the frames are: \{0, 1, 2, 3\}, \{4, 5, 6, 7\}, \{8, 9\}, \{A, B, C, D\}, \{E, F, G, H\}, so and so forth. Several frames of the video are shown in Fig.~\ref{fig:syntheticData2}.

% \begin{figure}[!b]
%   \centering
%   \includegraphics[width=0.7\linewidth]{Figures/paper1/8_CFrame.pdf}
%   %\includegraphics[width=1\linewidth]{latex/Figures/CurrentFrame.pdf}
%   \caption{Context selection from ``Current" frame only.}
%   \label{subfig:tsnePlotsynthetic_curr}
%   \vspace{1em}
% \end{figure}
% \begin{figure}[!b]
%   \centering
%   \includegraphics[width=0.7\linewidth]{Figures/paper1/8_NFrame.pdf}
%   %\includegraphics[width=1\linewidth]{latex/Figures/SurroundingFrames.pdf}
%   \caption{Context selection from current and surrounding frames.}
%   \label{subfig:tsnePlotsynthetic_surr}
% \caption{Comparison of T-SNE plots.}
% \label{fig:tsnePlotsynthetic}
% \end{figure}

%In this experiment, we study the quality of object vectors to cluster the objects present in surrounding frames. First, starting with synthetic data. Each of the synthetic frames Fig.~\ref{fig:syntheticData2} consist of 4 distinct objects in each frame, where objects in current and surrounding frames are in sequence (0-9, A-Z, and a-z ). The objective of this experiment is to analyze which of the two context selection approaches is better to capture context. Two different training set is generated using the two context object selection approaches. Our embedding model is trained using these two training sets to obtain two sets of object embedding vectors.
If we include surrounding frames of each reference frame in the context of a visual object, the embedding space should be able to capture the numbers and the letters in a sequence. That is, objects in frame \{8, 9\}, should have A, B, C, and D as context because \{A, B, C, D\} appears as the next frame of \{8, 9\} (Fig.~\ref{fig:syntheticData2}).

We repeated the sequence of frames for a rich training video, which resulted in a total of 10,000 frames. That is, after the frame \{y, z\} of the sequence, \{0, 1, 2, 3,\} reappeared in the video. Therefore, the static embedding model should keep the vectors of y and z close to the vectors of 0, 1, 2, and 3 in the embedding space. 

\renewcommand{\tabcolsep}{1pt}
\begin{table}[!h]
\centering
\caption{Ten nearest neighbors of some objects. Green color highlights the correctly detected context elements.}
\begin{tabular}{|c|c|l|l|l|l|l|l|l|l|l|l|c|l|l|l|l|l|l|l|l|l|}
\hline
\begin{tabular}[c]{@{}c@{}}Object\end{tabular} & \multicolumn{10}{c|}{\begin{tabular}[c]{@{}c@{}}Embedding space 1:\\Context selection using \\ ``Reference'' frame\end{tabular}} & & \multicolumn{10}{c|}{\begin{tabular}[c]{@{}c@{}}Embedding space 2:\\Context selection using \\``Reference'' \& ``Surrounding'' \\frames\end{tabular}} \\ \hline
0 & \cellcolor{green} 2 & \cellcolor{green} 1 & \cellcolor{green} 3 & s & d & 6 & W & k & y & \cellcolor{green} 4 & & \cellcolor{green} 2 & \cellcolor{green} 1 & \cellcolor{green} 3 & \cellcolor{green} y & \cellcolor{green} 4 & \cellcolor{green} 6 & \cellcolor{green} 5 & \cellcolor{green} z & \cellcolor{green} 7 & \cellcolor{green} 8 \\ \hline
9 & \cellcolor{green} 8 & t & s & R & E & O & J & q & Z & n &  & \cellcolor{green} 8 & \cellcolor{green} 5 & \cellcolor{green} B & \cellcolor{green} 4 & \cellcolor{green} 7 & \cellcolor{green} A & \cellcolor{green} D & \cellcolor{green} C & \cellcolor{green} 6 & \cellcolor{green} F \\ \hline
A & \cellcolor{green} B & \cellcolor{green} C & \cellcolor{green} D & W & 0 & \cellcolor{green} 4 & V & r & s & 1 &  & \cellcolor{green} B & \cellcolor{green} C & \cellcolor{green} D & \cellcolor{green} 8 & \cellcolor{green} E & \cellcolor{green} G & \cellcolor{green} F & \cellcolor{green} 9 & \cellcolor{green} H & \cellcolor{green} 4 \\ \hline
Z & \cellcolor{green} Y & M & b & N & K & c & o & p & i & I &  & \cellcolor{green} Y & \cellcolor{green} V & \cellcolor{green} b & \cellcolor{green} X & \cellcolor{green} d & \cellcolor{green} a & \cellcolor{green} U & \cellcolor{green} c & \cellcolor{green} W & \cellcolor{green} F \\ \hline
a & \cellcolor{green} d & \cellcolor{green} b & \cellcolor{green} c & G & k & \cellcolor{green} 2 & i & 1 & E & z &  & \cellcolor{green} b & \cellcolor{green} c & \cellcolor{green} d & \cellcolor{green} Y & \cellcolor{green} e & \cellcolor{green} g & \cellcolor{green} f & \cellcolor{green} Z & \cellcolor{green} h & \cellcolor{green} i \\ \hline
z & \cellcolor{green} y & k & 4 & 6 & Q & a & R & i & \cellcolor{green} 0 & \cellcolor{green} 1 &  & \cellcolor{green} y & \cellcolor{green} v & \cellcolor{green} 1 & \cellcolor{green} x & \cellcolor{green} 3 & \cellcolor{green} u & \cellcolor{green} 0 & \cellcolor{green} 2 & \cellcolor{green} w & \cellcolor{green} r \\ \hline
\end{tabular}
\label{tab:NearestNeighborsynthetic}
\end{table}

Table~\ref{tab:NearestNeighborsynthetic} presents the top 10 nearest neighbors for selected objects in two distinct embedding spaces -- one limited to the context of the reference frame and the other incorporating context from both the reference and surrounding frames. The nearest neighbors were computed using cosine similarity between embedding vectors. Correctly captured contexts of objects are highlighted in green. 
Here we can observe that embeddings based solely on single-frame context identify neighbors within their respective frames. In contrast, embeddings that integrate context from both reference and surrounding frames demonstrate the capability to identify neighboring objects spanning the surrounding frames. For example, in the first embedding space, among the top 10 nearest neighboring objects of ``9'', only ``8'' is relevant due to their co-occurrence in the same frame (row 2 in Table~\ref{tab:NearestNeighborsynthetic}). Conversely, all the top neighbors in the second embedding space are relevant as they are present in the surrounding frames. 

%\afarhan{previously written} Among the ten nearest neighbors of each object, the static embedding model that includes surrounding frames provides more correct contextual letters and digits than the model that only takes the context objects in the reference frame into account.

%In table~\ref{tab:NearestNeighborsynthetic}, the green marked objects are presented in the current frame and the yellow marked objects are from the surrounding frames of the reference objects. Where both approaches can capture neighboring objects from the current frame, only the second approach can capture neighboring objects from surrounding frames.
\begin{figure}[!h]
\centering
\begin{subfigure}{.48\textwidth}
  \centering
  \includegraphics[width=0.98\textwidth]{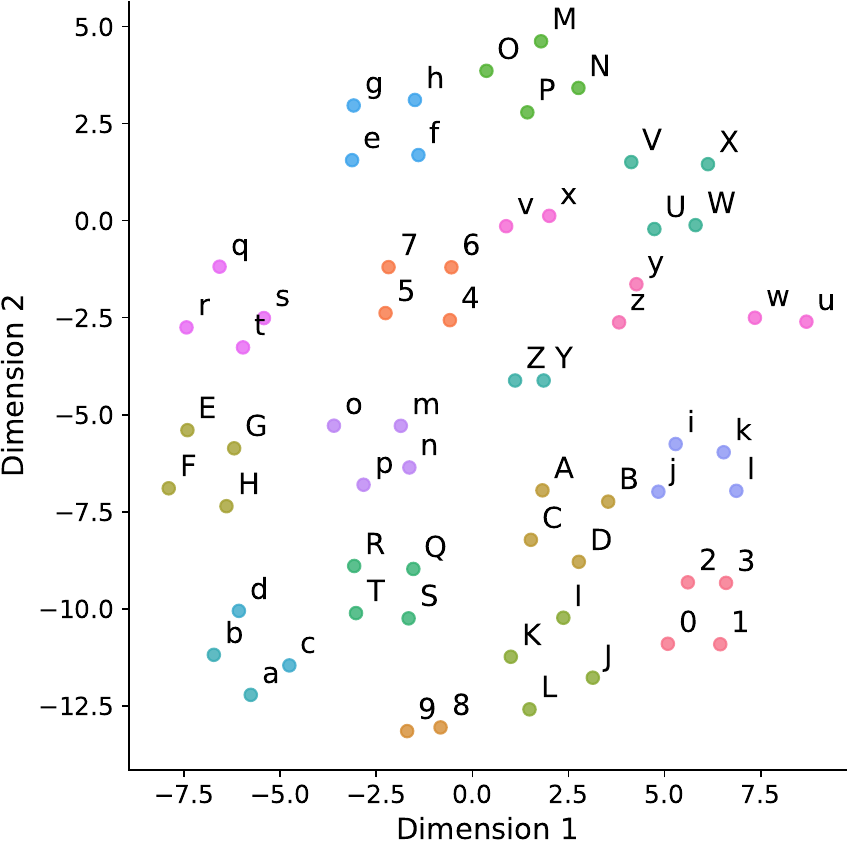}
  \caption{Context selection from ``Reference" frame only.}
  \label{subfig:tsnePlotsynthetic_curr}
\end{subfigure}
\begin{subfigure}{.48\textwidth}
  \centering
  \includegraphics[width=0.98\textwidth]{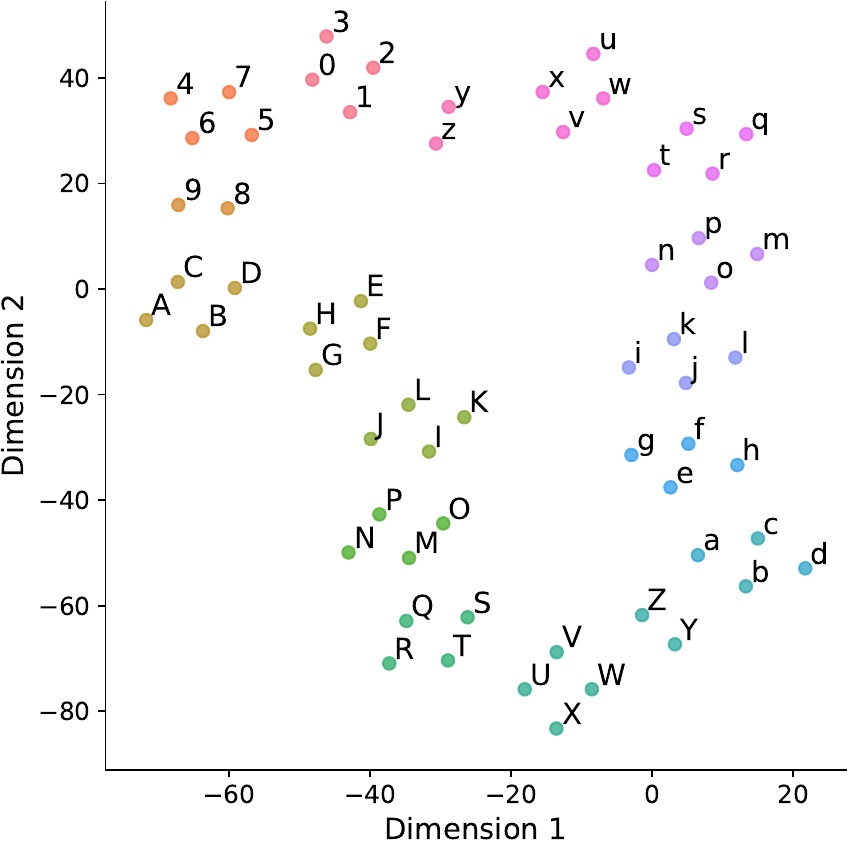}
  \caption{Context selection from ``Reference" and ``Surrounding'' frames.}
  \label{subfig:tsnePlotsynthetic_surr}
\label{fig:tsnePlotsynthetic}
\end{subfigure}
\caption{Comparison of T-SNE plots.}
\end{figure}

Fig.~\ref{fig:tsnePlotsynthetic} presents a comparison of two corresponding T-SNE plots (t-Distributed Stochastic Neighbor Embedding plot)~\cite{Maaten08} of 62 object vectors of the synthetic dataset. A T-SNE plot demonstrates higher-dimensional data on a projected low-dimensional space, in our case, a two-dimensional plane. Fig.~\ref{subfig:tsnePlotsynthetic_curr} demonstrates that objects appearing concurrently within a frame are positioned as neighbors in the embedding space., such as objects in \{8, 9\} create a group. Similarly, objects in \{A, B, C, D\} form another group, so and so forth. However, the objects in \{8, 9\} are not close to the objects of \{A, B, C, D\} even though the two frames appear one after another in the video (Fig.~\ref{fig:syntheticData2}). Fig.~\ref{subfig:tsnePlotsynthetic_surr} considers objects in the current and surrounding frames as context. That is why the object groups in Fig.~\ref{subfig:tsnePlotsynthetic_surr} follow the consecutive patterns of objects in the video. For example, \{A, B, C, D\} is close to \{E, F, G, H\}; then \{E, F, G, H\} is close to \{I, J, K, L\}, so and so forth. Fig.~\ref{fig:tsnePlotsynthetic} illustrates that including context from surrounding frames enhances the static embedding's ability to reflect the arrangement and temporal sequence of objects in a video.

This experimental analysis confirms that both context selection methods are effective and function as designed. Depending on the application, one can opt for single-frame context selection for image embeddings or choose to incorporate surrounding frames for video embeddings.

\subsubsection{Embedding with reference-context pair in static image dataset}
\label{sec:staticImages}
Due to the scarcity of labeled video data to test our model, we include COCO image dataset~\cite{Lin14} and compared our results with an image embedding method proposed by Lüddecke et al~\cite{LUDDECKE19}. We used single-frame context window (Context 1 in Section~\ref{sec:context_selection}), and negative sampling (as described in Section~\ref{sec:negative_sampling}), to make our model suitable for image data.

For the evaluation, we used two mechanisms as proposed by Lüddecke et al~\cite{LUDDECKE19}. The evaluation is driven by two metrics: (1) clustering consistency, and (2) system-to-human correlation. Both the techniques are defined below.
%As there are no significant work has done previously on object embedding in video data, dataset of labeled objects in video data is scarce. For the limitation of labeled video data, we have done some experiments on coco image dataset~\ref{tab:datasetList} considering each image is equivalent to frames from video. As consecutive images have no relation like videos, only context object from current frame is considered. However, we introduced negative sampling to set low context strength of objects from different images. For result analysis, we have followed the evaluation technique from~\cite{LUDDECKE19} as their works focus on embedding of objects from image dataset. They discussed two quality measures to evaluate obtained semantic vectors: (1) clustering consistency and (2) system-to-human. We have compared our result with the the results from this paper.

\textbf{Clustering consistency}: Clustering consistency compares clusters of objects obtained from an embedding with clusters defined by super-categories of objects in the labeled data. In the COCO dataset, 80 object classes are grouped into eleven super-categories, such as animal, vehicle, and kitchen~\cite{Lin14}. In cluster consistency, we measure the proportion of \textit{K}-nearest neighbors of the embedding vector in the same super-category as the object's super-category.
%In an ideal case, the closest neighbors of a given semantic vector would all belong to the same super-category. We measure to which proportion this is satisfied. Through cluster consistency experiment, we want to measure, which portion of the closest neighbors of a given semantic vector belong to the same super-category of the object of that semantic vector. Specifically, for every object $o$ in a super-category, we consider its vector from the VizObj2Vec model and then find the k nearest neighbor vectors. Of those k vectors we count the number of objects that fall into the same super-category as $o$ and normalize the count by k. Then we average results for all $o$ in that super-category, for all super categories and, finally, we average scores for k = 1 to k = 5 (5 is the size of the smallest super-category in COCO) to obtain a single scalar score which is the cluster consistency of the VizObj2Vec model.

\textbf{System-to-human correlation}: This evaluation computes as score using two rankings -- (a) based on vectors and (b) based on human-annotated ranks -- by estimating the Spearman rank correlation. We used Scene250~\cite{LUDDECKE19} annotation as the benchmark human-annotation of ranking. 
%For this we compared similarity and relatedness scores of word-pairs determined by human raters, using the benchmarks MEN, Simlex, and Scene250, with ratings for the same pairs calculated by the different automatic procedures. Spearman rank correlation calculation is performed on the vectors obtained from VizObj2Vec against MEN, Simlex, and Scene250.

\begin{figure}[!b]
\centering
\begin{subfigure}[b]{\textwidth}
  \centering
  \includegraphics[width=0.5\textwidth]{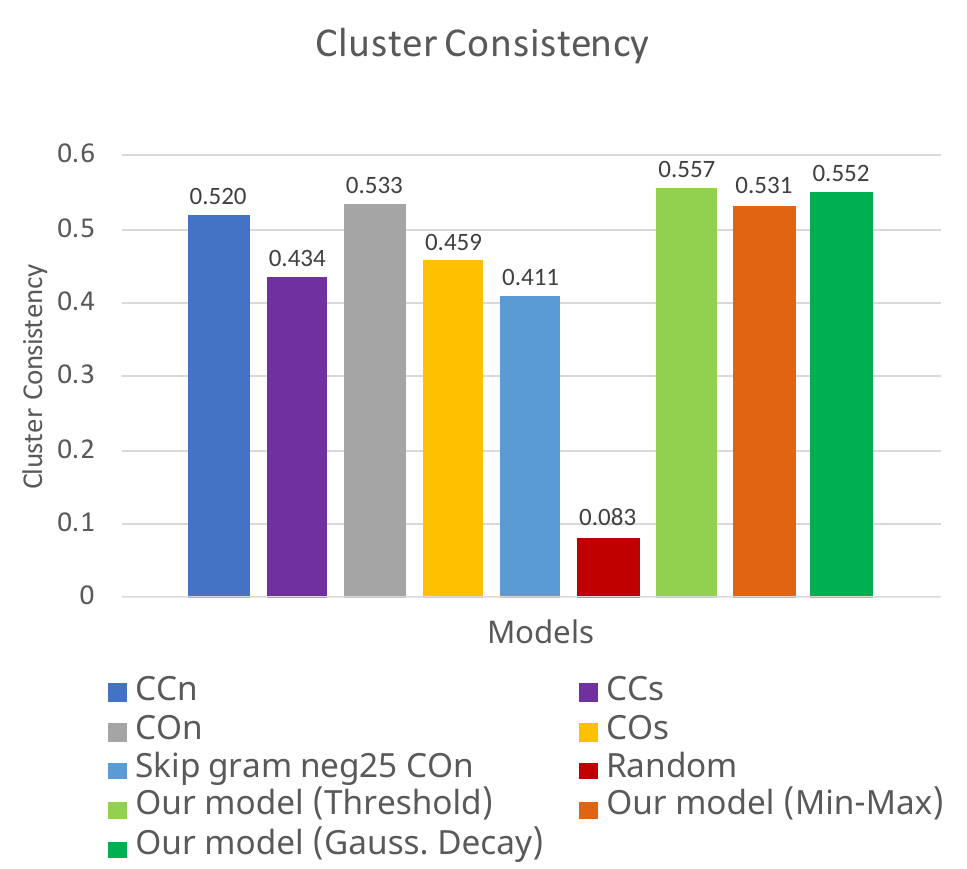}
        \caption{Cluster Consistency}
        %\label{fig:top}
\end{subfigure}%

\begin{subfigure}[b]{.49\textwidth}
  \centering
  \includegraphics[width=0.95\textwidth]{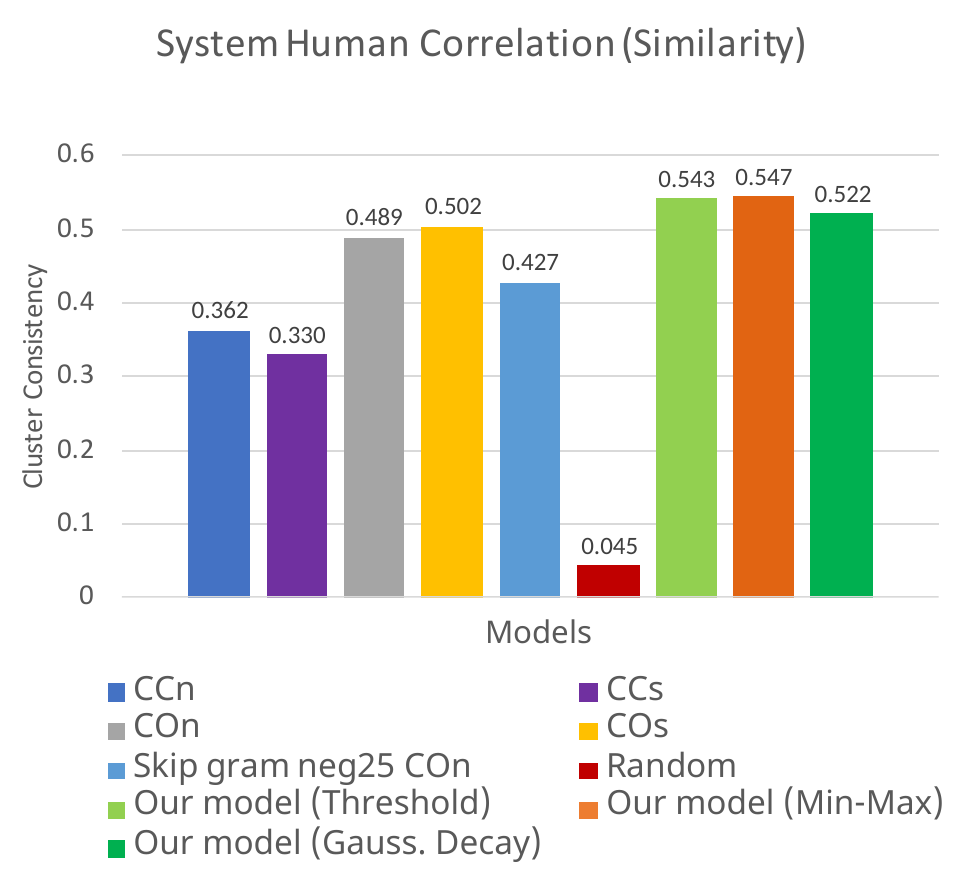}
        \caption{System Human Correlation (Similarity)}
        %\label{fig:bottom_left}
\end{subfigure}
\hfill
\begin{subfigure}[b]{.49\textwidth}
  \centering
  \includegraphics[width=0.95\textwidth]{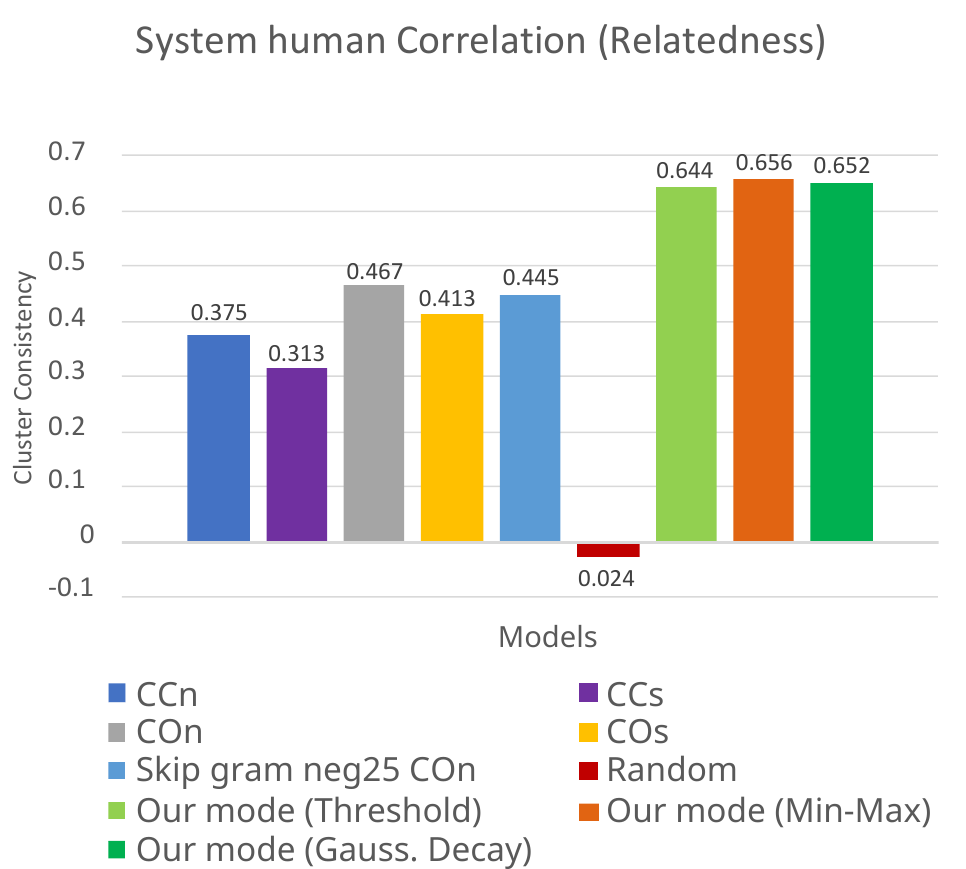}
        \caption{System human Correlation (Relatedness)}
        %\label{fig:bottom_right}
\end{subfigure}
\caption{Experimental evaluation and comparison of different context models (given by different colors, see legend at the bottom of the figure) on COCO dataset. The top chart shows clustering consistency. The two bottom charts show system-to-human correlation for Scene250 data measuring similarity and relatedness.}
\label{fig:coco_dataset_exp}
\end{figure}

We compared results from our static embedding model with the \textbf{context-based models} of Luddecke et al~\cite{LUDDECKE19}: co-count-noself context (CCn), co-count-self context (CCs), co-occurrence-noself context (COn), co-occurrence-self context (COs), and word2vec(Skip-gram neg25 COn). For a fair comparison, we only included contextual aspects of the methods presented in Luddecke et al~\cite{LUDDECKE19} in this experiment. We included three of our contextual discrepancy scores ( Distance threshold, Min-Max scaling, and Gaussian decay). 
%The visual feature based model in~\cite{LUDDECKE19} are not discussed in this experiment as our model is designed to consider the contextual relation between objects rather than considering visual similarity between objects.

% \begin{figure*}[!h]
% \centering
% \begin{subfigure}[b]{.33\textwidth}
%   \centering
%   \includegraphics[width=1\textwidth]{Figures/paper1/9_ClusterConsistency1.pdf}
%   %\caption{Frame 1}
% \end{subfigure}%
% \begin{subfigure}[b]{.33\textwidth}
%   \centering
%   \includegraphics[width=1\textwidth]{Figures/paper1/9_ClusterConsistency2.pdf}
%   %\caption{Frame 2}
% \end{subfigure}
% \begin{subfigure}[b]{.33\textwidth}
%   \centering
%   \includegraphics[width=1\textwidth]{Figures/paper1/9_ClusterConsistency3.pdf}
%   %\caption{Frame 3}
% \end{subfigure}
% \caption{Experimental evaluation and comparison of different context models (given by different colors, see legend at the bottom of the figure) on COCO dataset. The left-most chart shows clustering consistency. On the right, system-to-human correlation for Scene250 data measuring similarity and relatedness are shown.}
% \label{fig:coco_dataset_exp}
% \end{figure*}

Fig.~\ref{fig:coco_dataset_exp} shows that our models driven by static embedding outperform almost all of the contextual models proposed by Luddecke et al~\cite{LUDDECKE19}. In each plot of Fig.~\ref{fig:coco_dataset_exp}, the three right-most bars represent static embedding models. The cluster consistency and system human correlation (similarity) scores of static embedding models are competitive to other models. In the case of relatedness, static embedding models results in much higher scores. Our static embedding framework uses visual resemblance only for detection of objects, not for context generation. It is natural that contextually connected visual objects -- such as a monitor and a keyboard -- might not have any visual resemblance at all. The static embedding exhibits higher relatedness scores because it prioritizes contextual similarity of objects over the visual resemblance.
%Note that, our model is not designed to categorize visually similar objects--which is why we do not consider visual features-- rather our model establish contextual relation between nearly seen objects. That is why we obtain higher relatedness score as related objects are not required to be visually similar rather required to be contextually connected.

Through this comparison of cluster consistency, similarity, and relatedness, we observe that, even though our static embedding model is designed for video data, it performs competitively or outperforms a state of the art contextual embedding method for image data. 
\subsubsection{Sense-making using with a language model: Case study 1}
\label{sec:case1}
In subsection~\ref{sec:staticImages}, we compared our static embedding model in terms of objects similarity and relatedness to an image embedding model where the ground truth of similarity was taken from the COCO dataset and ground truth of object relatedness was taken from human-annotated ranks.
In this subsection, we attempt to make sense of our static embedding model by examining the similarity of pairs our visual objects found in a natural language. We pick a pair of visual objects, either from the same frame or from two different frames of the labelme dataset~\cite{yuen09} (Table~\ref{tab:datasetList}), and compute the cosine similarity between the vectors of the pair of objects. 
We picked objects for which there are known vectors in the language model word2vec with English Google News corpus\cite{Mikolov13a, Mikolov13b}. %For the comparison purpose, we compute the cosine similarity between the word2vec-generated vectors of the same pair of objects. That is, 
We examine the similarity computed from our video-based vectors with the similarity computed by word2vec vectors.
%For example, 
%we compute the cosine similarity between two objects --- keyboard and mouse --- using the vectors generated by our vizOvj2Vec model. 

The cosine similarity between keyboard and mouse is 0.96 using our model. The cosine similarity between the vectors generated by word2vec for the same two objects is 0.47. Table~\ref{tab:cosinSimCompare} shows such similarity comparison with eight pairs of objects.

With object pairs from the same frame, in each case, the cosine similarity computed from vectors of vizOvj2Vec is higher than that of word2vec. Clearly, when it comes to an object pair in the same frame, our model creates vectors that are easily distinguishable.

\begin{table}[!h]

\caption{Sample cosine similarities between pairs of vectors generated by our static object embedding and word2vec. Our embedding is trained by video frames from LabelMe dataset and objects in video frames are detected by yoloV4.}
\begin{tabular}{|c|c|c|c|c|}
\hline
\textbf{\begin{tabular}[c]{@{}c@{}}Objects \\ detected\\ in *\end{tabular}} & \textbf{object 1} & \textbf{object 2} & \textbf{\begin{tabular}[c]{@{}c@{}}Static Embedding\end{tabular}} & \textbf{W2Vec} \\ \hline
\multirow{4}{*}{\begin{tabular}[c]{@{}c@{}}Same \\ frame\end{tabular}} & keyboard & mouse & 0.96 & 0.47 \\ 
 & refrigerator & oven & 0.93 & 0.61 \\ 
 & bottle & laptop & 0.68 & 0.20 \\ 
 & refrigerator & keyboard & 0.70 & 0.26 \\ \hline
\multirow{4}{*}{\begin{tabular}[c]{@{}c@{}}Different\\ frames\end{tabular}} & bottle & diningtable & 0.81 & 0.10 \\ 
 & oven & cup & 0.82 & 0.26 \\ 
 & bottle & bird & -0.17 & 0.17 \\  
 & diningtable & kite & -0.18 & -0.002 \\ \hline
\end{tabular}
\label{tab:cosinSimCompare}

\end{table}

With objects from different frames, the similarity using our model is high only when the objects are from nearby frames, indicating a contextual similarity. Such high values are observed among pairs bottle-diningtable and oven-cup. Although bottle and dining table are not in the same frame, they are linguistically connected based on our sense about kitchen-related items. With object-pairs that are from frames that are far away, our model exhibits negative cosine similarity indicating well separation of the data points in the space. word2vec also provides low cosine similarity for such pairs. For example, with the bottle-bird pair, the cosine similarity using our model is -0.17 and word2vec is 0.17. Dining table and kite have a cosine similarity of -0.18 using our video-based model and -0.002 with word2vec.  

In summary, Table~\ref{tab:cosinSimCompare} demonstrates that the video-based similarity of pairs of objects using our model supports the word2vec linguistic model.

\subsubsection{Sense-making using contextual vectors: Case study 2}
\label{sec:case2}

\begin{figure}[!h]
    \centering
    \begin{subfigure}[b]{0.33\textwidth}
        \includegraphics[width=\textwidth]{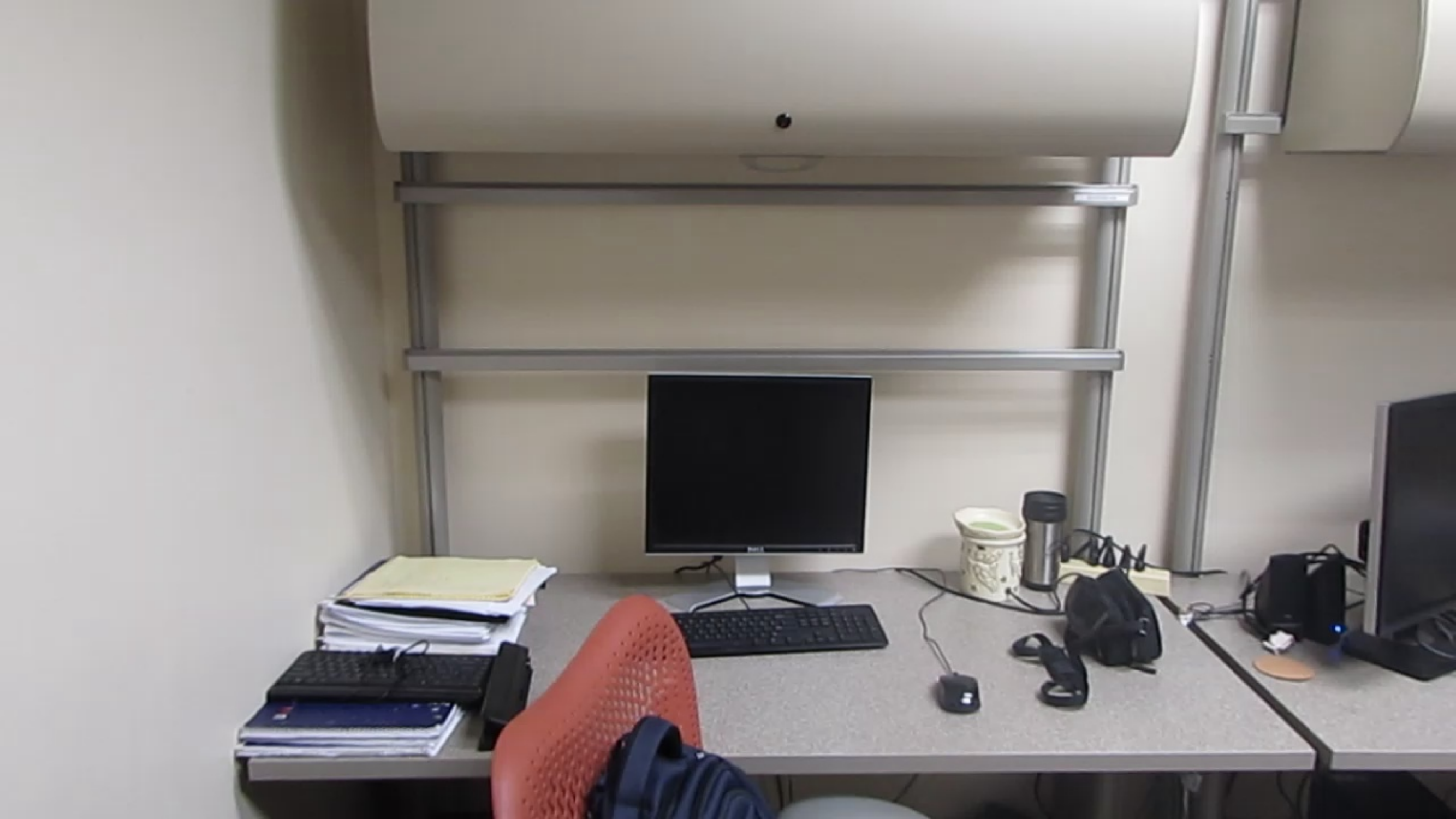}
        \caption{Lab}
        \label{fig:lab}
    \end{subfigure}
    ~ %add desired spacing between images, e. g. ~, \quad, \qquad, \hfill etc. 
    \begin{subfigure}[b]{0.30\textwidth}
        \includegraphics[width=\textwidth]{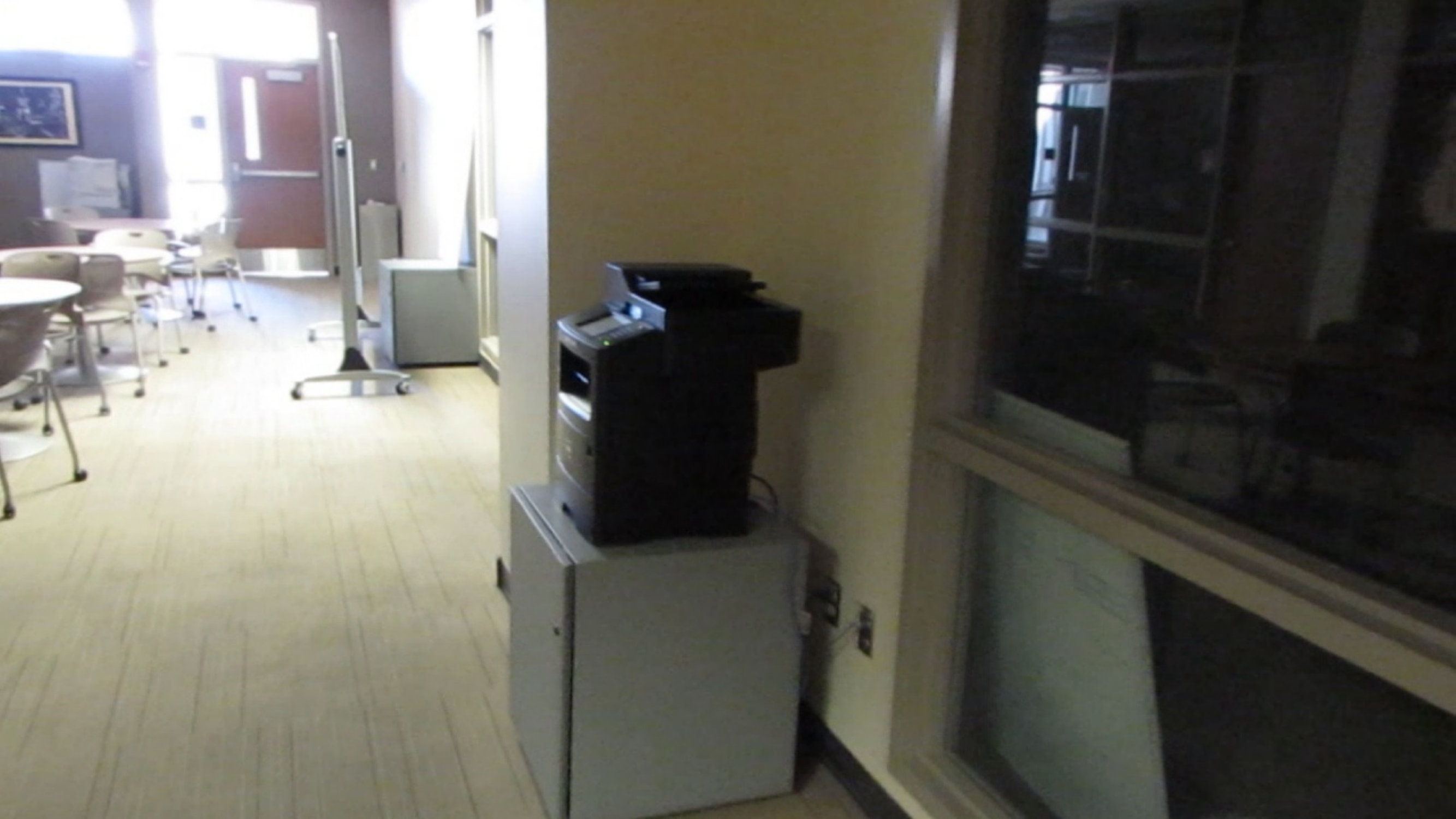}
        \caption{Corridor}
        \label{fig:corridor}
    \end{subfigure}
    ~ %add desired spacing between images, e. g. ~, \quad, \qquad, \hfill etc. 
    \begin{subfigure}[b]{0.30\textwidth}
        \includegraphics[width=\textwidth]{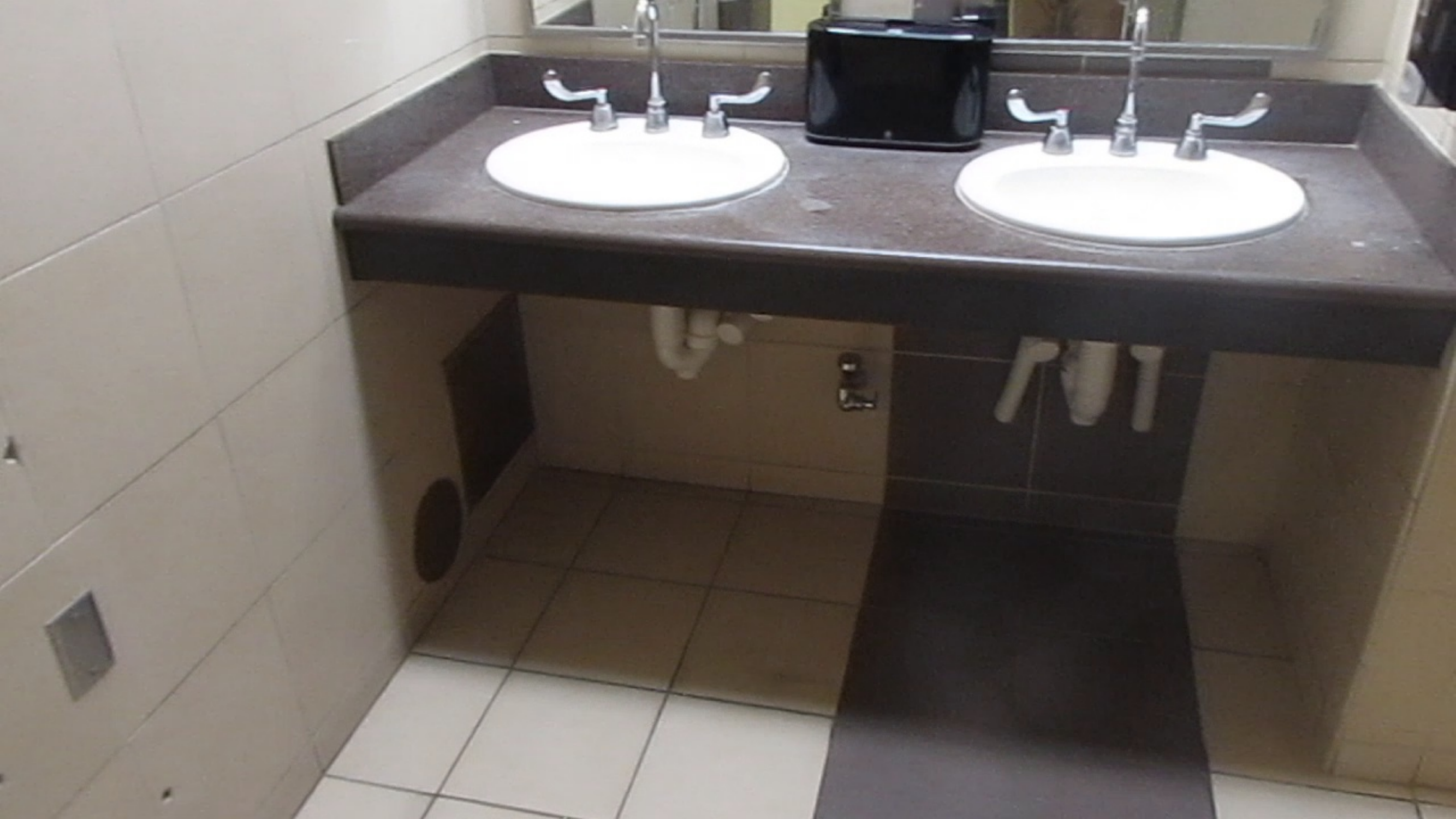}
        \caption{Restroom}
        \label{fig:lift}
    \end{subfigure}
    \begin{subfigure}[b]{0.30\textwidth}
        \includegraphics[width=\textwidth]{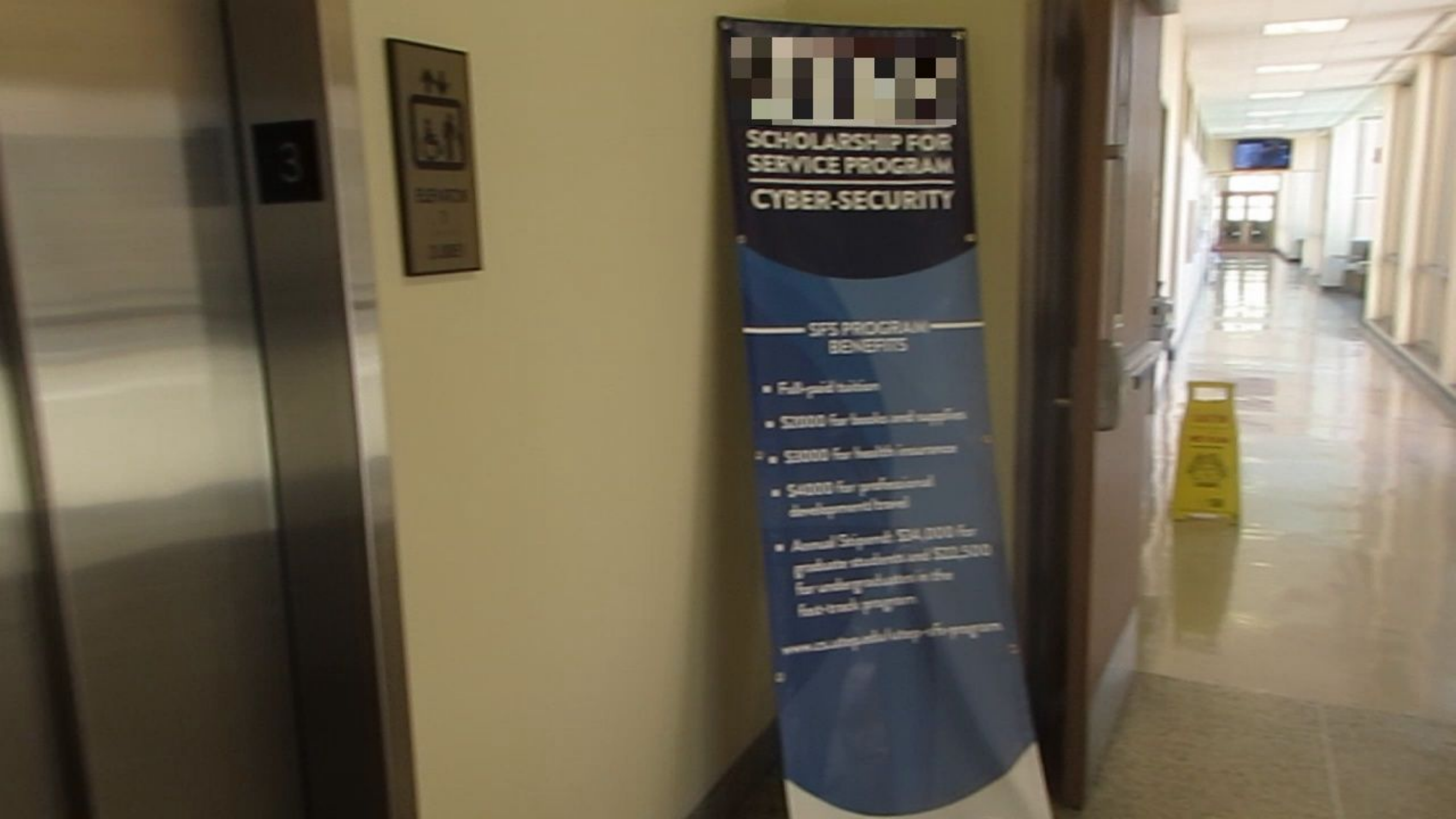}
        \caption{Elevator}
        \label{fig:lift}
    \end{subfigure}
    \begin{subfigure}[b]{0.30\textwidth}
        \includegraphics[width=\textwidth]{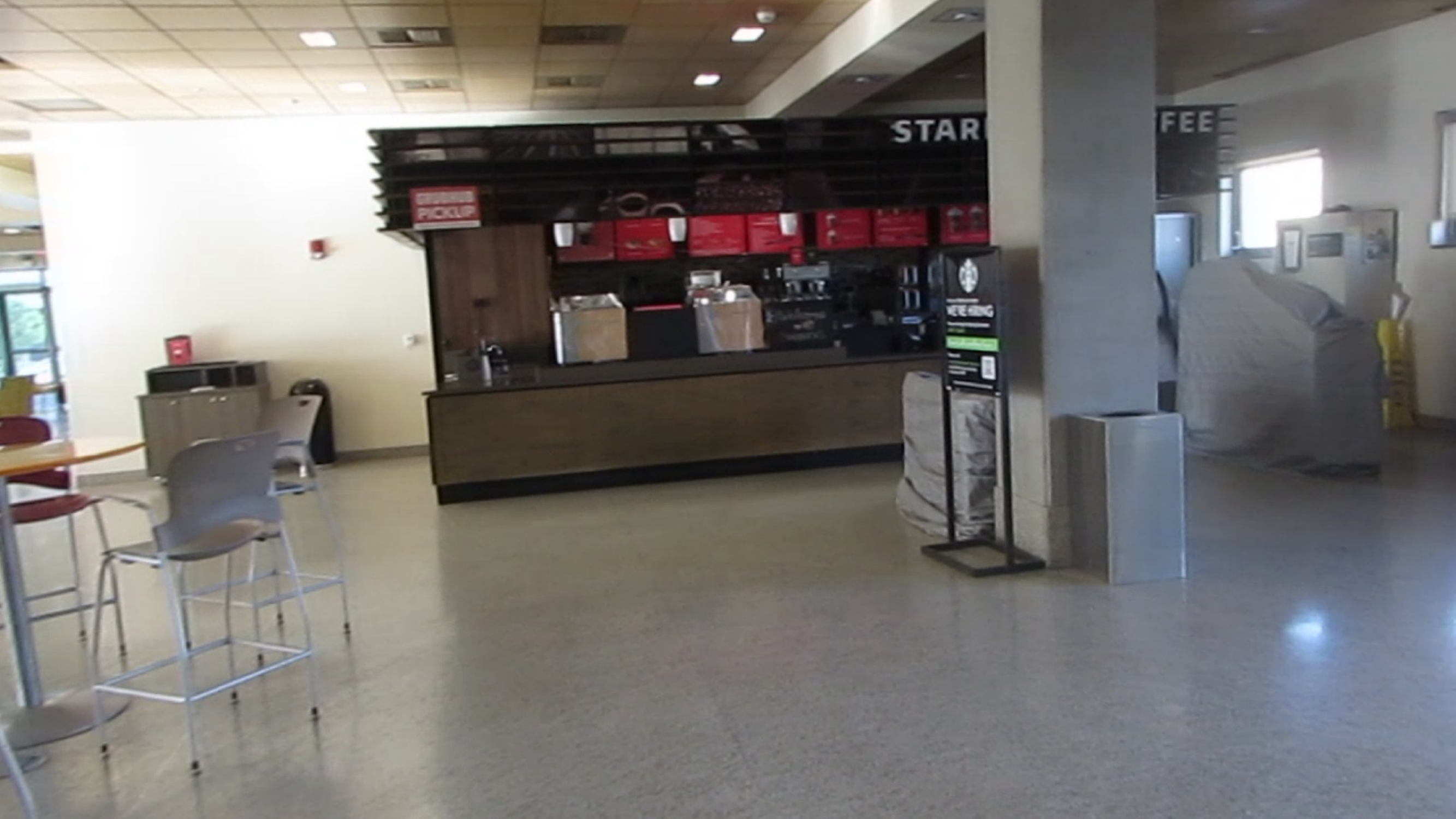}
        \caption{Cafe}
        \label{fig:cafe}
    \end{subfigure}
    \begin{subfigure}[b]{0.30\textwidth}
        \includegraphics[width=\textwidth]{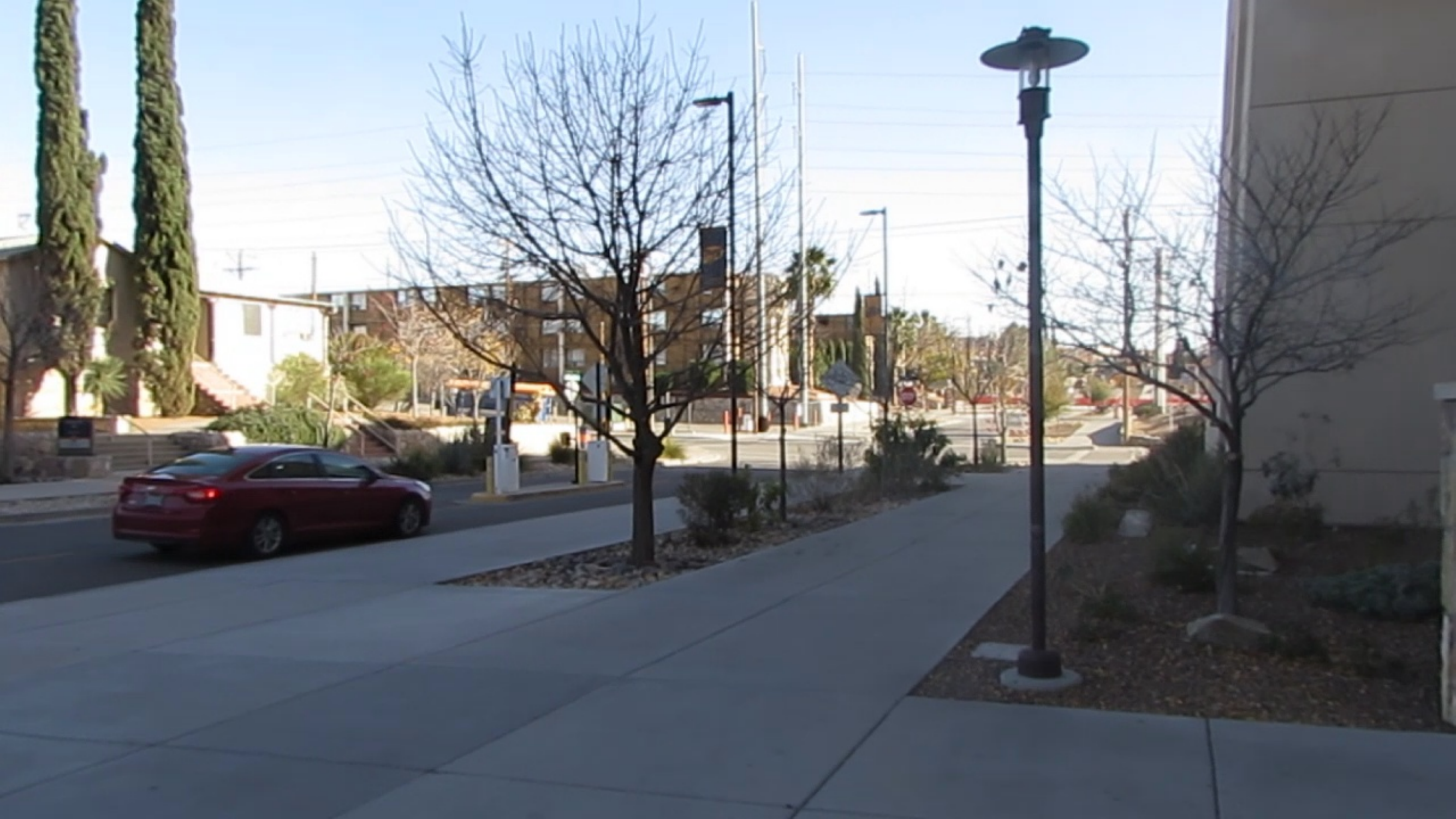}
        \caption{Outdoor}
        \label{fig:outdoor}
    \end{subfigure}
    \caption{Frames from recorded video.}\label{fig:frames}
    \label{fig:framesRecorded}
\end{figure}

In this study, we explore the utility of static embeddings in analyzing the context of visual objects within video content. For this study, we recorded a video beginning in a research lab, progressing through a corridor, visiting a restroom, using an elevator, passing by a Starbucks, and finally, moving the camera to the outside of the building. Fig.~\ref{fig:framesRecorded} shows some of the frames of the video.
%In this case study, we analyzed a video that we shot ourselves. We started recording from a research lab, then move to the corridor, then to the toilet, then we use a lift, go by Starbucks, and finally, we move the camera to the outside of the building. Fig.~\ref{fig:framesRecorded} shows some of the frames of the video.

\begin{figure}[t]
  \centering
  \includegraphics[width=0.5\textwidth]{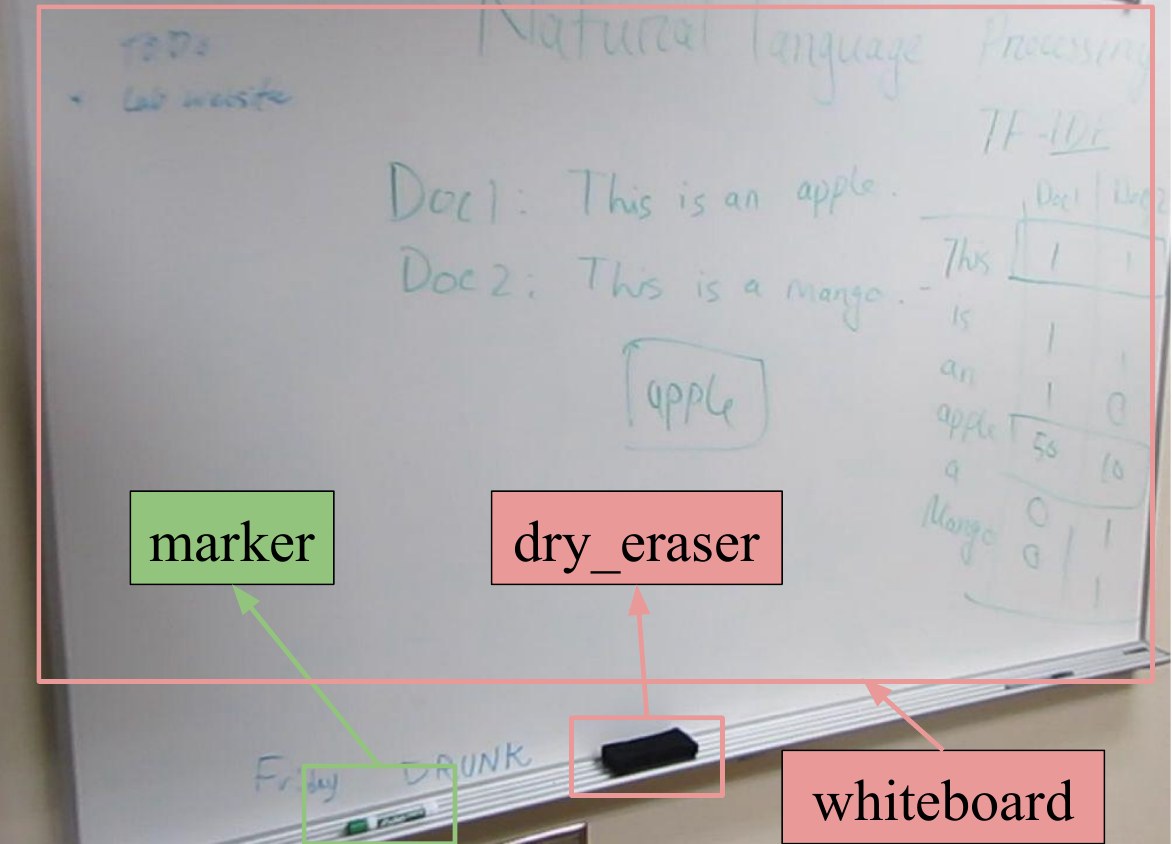}
  \caption{The frame containing a ``marker''.}
  \label{fig:marker}
\end{figure}

In this experiment, we select vectors of context objects in the embedding space given a reference object. We select the context vectors based on several nearest neighbors (using cosine similarity between reference-context pairs) of the reference object. We used two approaches for the context selection, one selects contexts from the same frame and the other includes surrounding frames. The reference-context strength scoring in both approaches used the Gaussian decay-based formula of Eq.~\ref{eq:gauss_decay_norm} for the static embedding model. As an example of reference and context, in this subsection, we used a ``marker'', which has a ``dry\_eraser'' and a ``whiteboard'' as the context present in the same frame, as shown in Fig.~\ref{fig:marker}.

%The synthetic data is much structured for finding the contextual relation. On the other hand, the automated annotated video data contains some detection error which creates some error in contextual vector relation Fig.~\ref{fig:faucetToilet}. Therefore to evaluate the actual performance of visualObj2vec, we manually annotated objects in a recorded video that contains scenarios from different environments. Started from a computer lab, the video recording cover lab corridor, building corridor, a toilet, lift, cafe and outside the lab building sequentially (Fig.~\ref{fig:framesRecorded}). 
%To analyze context integration from surrounding frames in real-video data we have created a base model where the nearest neighbors of each reference object are sorted by their distances (Eq.~\ref{eq:distance}) with the reference objects. As we calculate two different vector representations using two context selection methods, we get two different k-nearest neighbors set for each objects. Then the overlap of nearest neighbor with baseline is calculated and plotted in Fig.~\ref{fig:neighInterContexSelection}. This figure exhibits, surrounding frame context selection process has more overlap of nearest neighbor with the baseline model compared to the current frame context selection process.

Table~\ref{tab:markerNeighbor} demonstrates that both the ``dry\_eraser'' and the ``whiteboard'' were detected as two neighbor objects of ``marker'' using both context selection approaches. The single-frame-only approach failed to detect any other correct object that was detected in the surrounding frames. It captured ``fire\_alarm'', ``push\_button'', and ``mug'' as neighbor objects that were, in reality, random because the embedding vectors from the third nearest neighbors were drastically different from the first two neighbors in the embedding space. The approach that included both the reference frame and the consecutive frames was able to capture several other objects that were in the lab around the marker such as ``table'', ``shelf'', and ``telephone''. The case study reveals that the embedding of the objects using contexts from surrounding frames encodes a story of how the events relevant to the objects proceeded. 
\begin{table}[!b]
\caption{Nearest neighbor of ``marker'' vector by context selection methods using (1) the reference frame only, and (2) reference and surrounding frames. Green background indicates high quality context object. Vector similarity values are provided in the parenthesis.}
\begin{tabular}{|c|c|}
\hline

\begin{tabular}[c]{@{}c@{}}\textbf{Context selection using} \\ \textbf{``Reference'' frame}\end{tabular} & \begin{tabular}[c]{@{}c@{}}\textbf{Context selection using ``Reference''}  \\\textbf{\& ``Surrounding'' frames}\end{tabular} \\ \hline
\cellcolor{green} dry\_eraser (0.97) & \cellcolor{green} whiteboard (0.99) \\ \hline
\cellcolor{green} whiteboard (0.92) & \cellcolor{green} dry\_eraser (0.98) \\ \hline
 fire\_alarm (0.40) & \cellcolor{green} table (0.70) \\ \hline
push\_button (0.27) & \cellcolor{green} shelf (0.69) \\ \hline
mug (0.27) & \cellcolor{green} telephone ( 0.69) \\ \hline
\end{tabular}
\label{tab:markerNeighbor}
\end{table}
We analyzed the nearest neighbor contexts of all 93 reference objects labeled in the video. The nearest neighbor contexts of each reference object were meaningful and contained higher quality neighbors when we included surrounding frames for context selection, indicating that the inclusion of surrounding frames is crucial in embedding video-objects.

\begin{figure*}[!b]
    \centering
    \begin{subfigure}[b]{0.8\linewidth}
        \includegraphics[width=\textwidth]{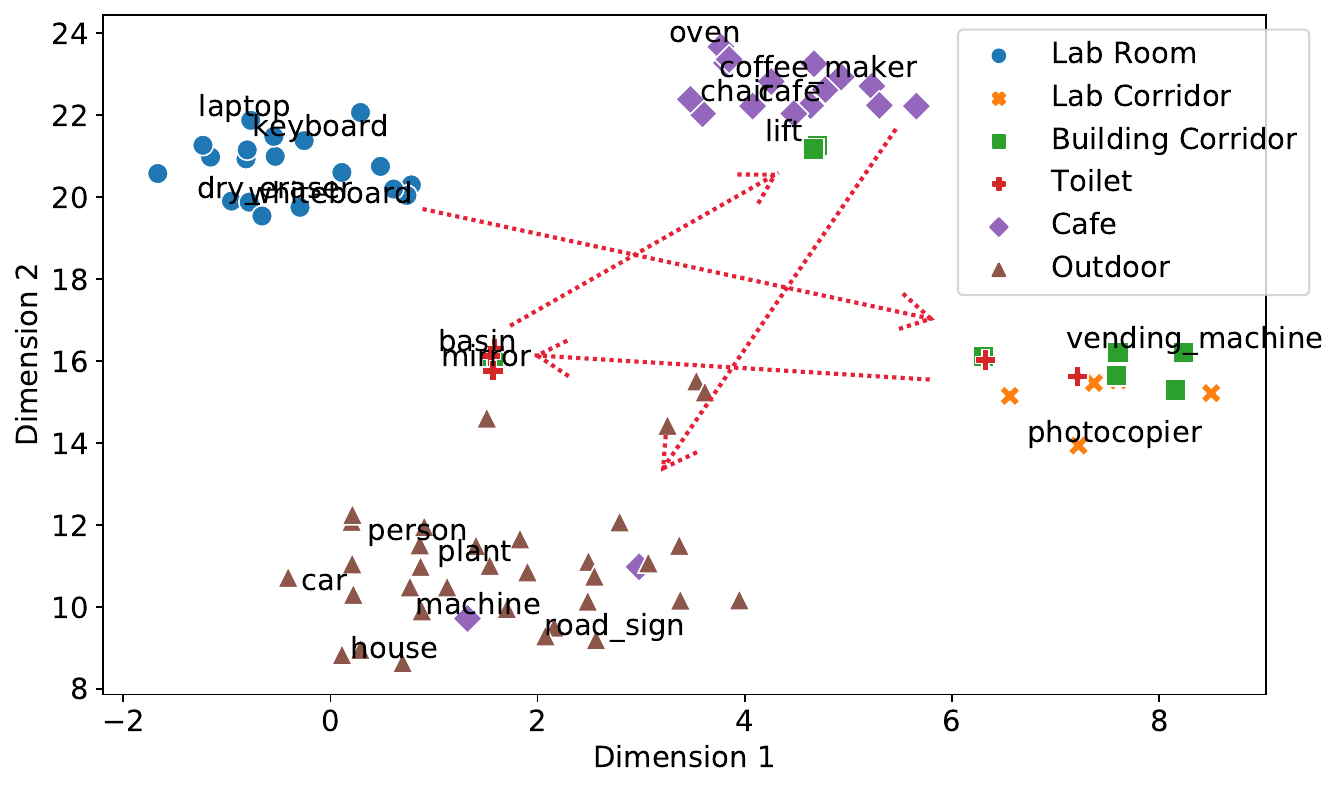}
        \caption{Context selection using ``Reference" frame only.}
        \label{fig:tsneContextCurrentFrame}
    \end{subfigure}
    ~ %add desired spacing between images, e. g. ~, \quad, \qquad, \hfill etc. 
    \begin{subfigure}[b]{0.8\linewidth}
        \includegraphics[width=1\textwidth]{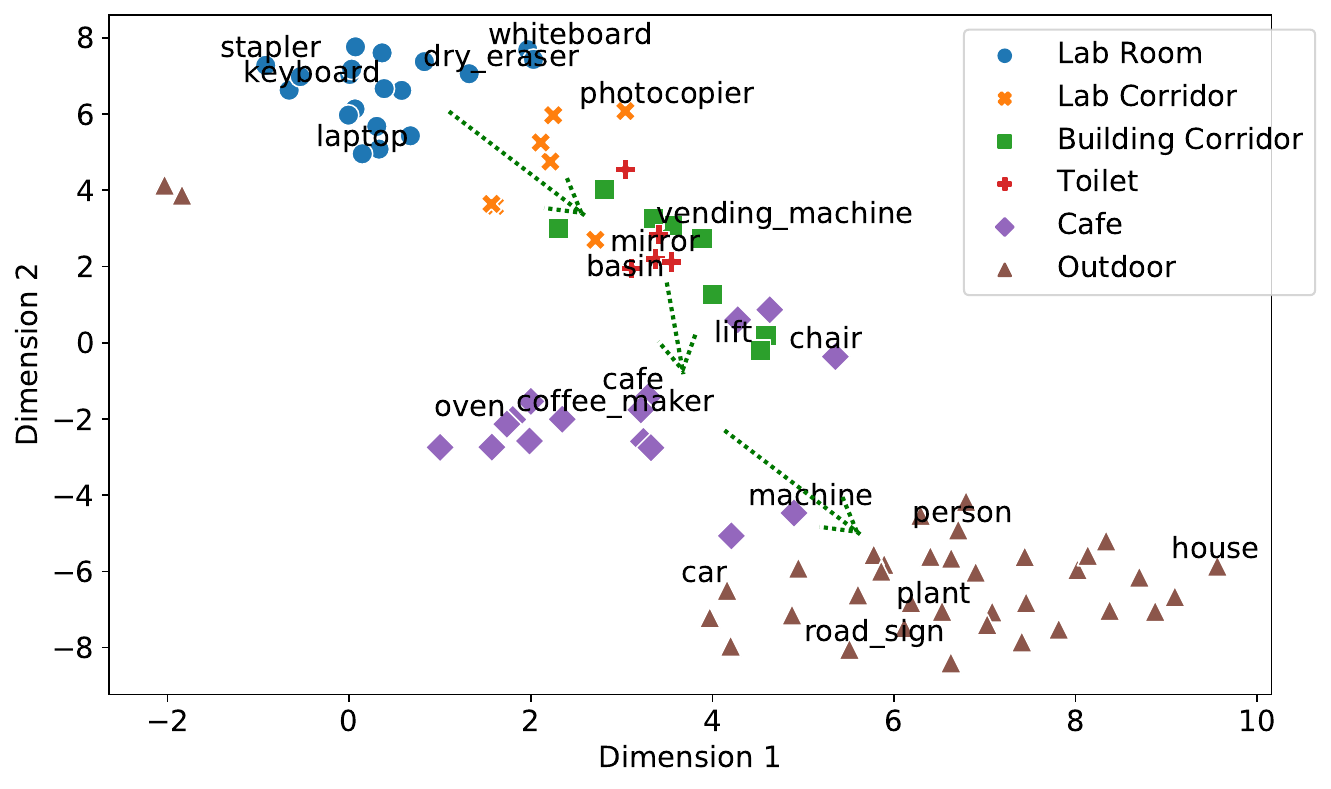}
        \caption{Context selection using ``Reference" and ``Surrounding" frames.}
        \label{fig:tsneContextNeighborFrame}
    \end{subfigure}
    \caption{T-SNE plots of the embedding spaces for two context selection methods.}\label{fig:16}
\end{figure*}

\subsubsection{Sense-making using an embedding space: Case study 3}
\label{sec:case3}
In this section, we focus on the entire holistic view of the embedding space created by the static embedding model. Here, we compute the reference-context discrepancy scores by employing the Gaussian decay normalization function, denoted by Equation~\ref{eq:gauss_decay_norm}. We compared the context selection using the reference-frame-only approach with the approach that includes objects from the surrounding frames for context selection.

%We compared the single-frame-only approach with the approach that includes objects from the surrounding frames for context selection.

%In this section, we analyze the ability of the object vectors to differentiate indoor and outdoor objects. For this experiment, we merged two different videos. The first part of the merged video contains frames from inside a house and the next part contains frames from a city tour video \footnote{City tour video: \url{https://www.youtube.com/watch?v=dw3fkbdEv4c}}. After running our visualObj2Vec model, we get 30 length vectors for all the objects from the merged video. The T-SNE tool is used to visualize 30-dimensional vectors in 2d space. Fig.~\ref{fig:tsneIndoorOutdoor} presents the T-SNE plot of the object vectors. This plot clearly shows us, how good the object vectors are to separate the indoor and outdoor objects. 

We used the same video of the previous subsection (Figure~\ref{fig:framesRecorded}) and constructed T-SNE plots of all the vectors using two approaches -- context selection using the reference frame only and context selection using the reference and surrounding frames. Fig.~\ref{fig:tsneContextCurrentFrame} shows that the objects from similar scenes are clustered in groups. We drew arrows manually to indicate the movements of the camera from one set of similar scenes to another set. Since the context selection using the reference frame-only approach does not encode the sequence of the frames, objects that are connected contextually via single frames are grouped. The arrows are meaningless in this scenario. Since the ``\textit{reference frame only}'' approach does not encode the sequence of frames, it has a strong contextual clustering capability.
%Fig.~\ref{fig:tsneContextCurrentFrame} shows that the objects are clustered in groups. The arrows indicate movements of the camera from one cluster to another.

The T-SNE plot (in Fig.~\ref{fig:tsneContextNeighborFrame}) for the embedding that includes objects of the surrounding frames as context balances object-clusters and encoding of the sequence of objects in the video. The arrows indicate the flow of the sequence of the video. The story of the video can be explained from the T-SNE with the arrows: the camera moved from a cluster of objects containing laptop, keyboard, dry eraser, and whiteboard to the vending machine area. Then the camera visited an area that has a cluster of objects such as coffee maker, oven, and cafe. The camera finally went outside and objects such as road sign, cars, and houses were detected.

This case study provides evidence that both context selection approaches provide meaningful outcomes that summarize contextual relationships between visual objects in a video. That is, video summarization can be an application of the proposed static embedding model.

%This case study provides an insight that both context selection using the \textit{current frame only} approach and the approach that includes the surrounding frames provide meaningful outcomes. 

%One may decide to choose one approach over the other based on analytic needs.
%We have also analyzed the TSNE output of the object's embeding vectors for the human annotated dataset. In Fig.~\ref{fig:tsneContextCurrentFrame} and Fig.~\ref{fig:tsneContextNeighborFrame}, we can see object vectors from the "context from current frame" model are grouped closely while object vectors from "context from current \& surrounding frames" model are scattered. The reason behind this is the context of objects has increased due to introducing objects from the surrounding frame as a context object. That is why the connection of the object with other objects increases and the vectors of the objects become scattered to establish the connection. Another observation from Fig.~\ref{fig:tsneContextNeighborFrame} is the object vectors are in sequence they appear in video frames. The video starts from the lab room then travels to the lab corridor, building corridor, lift, cafe, and outdoor. This sequence is also reflected by the object vectors TSNE projection. 

\begin{figure}[!b
]
\begin{center}
  \includegraphics[width=0.8\textwidth]{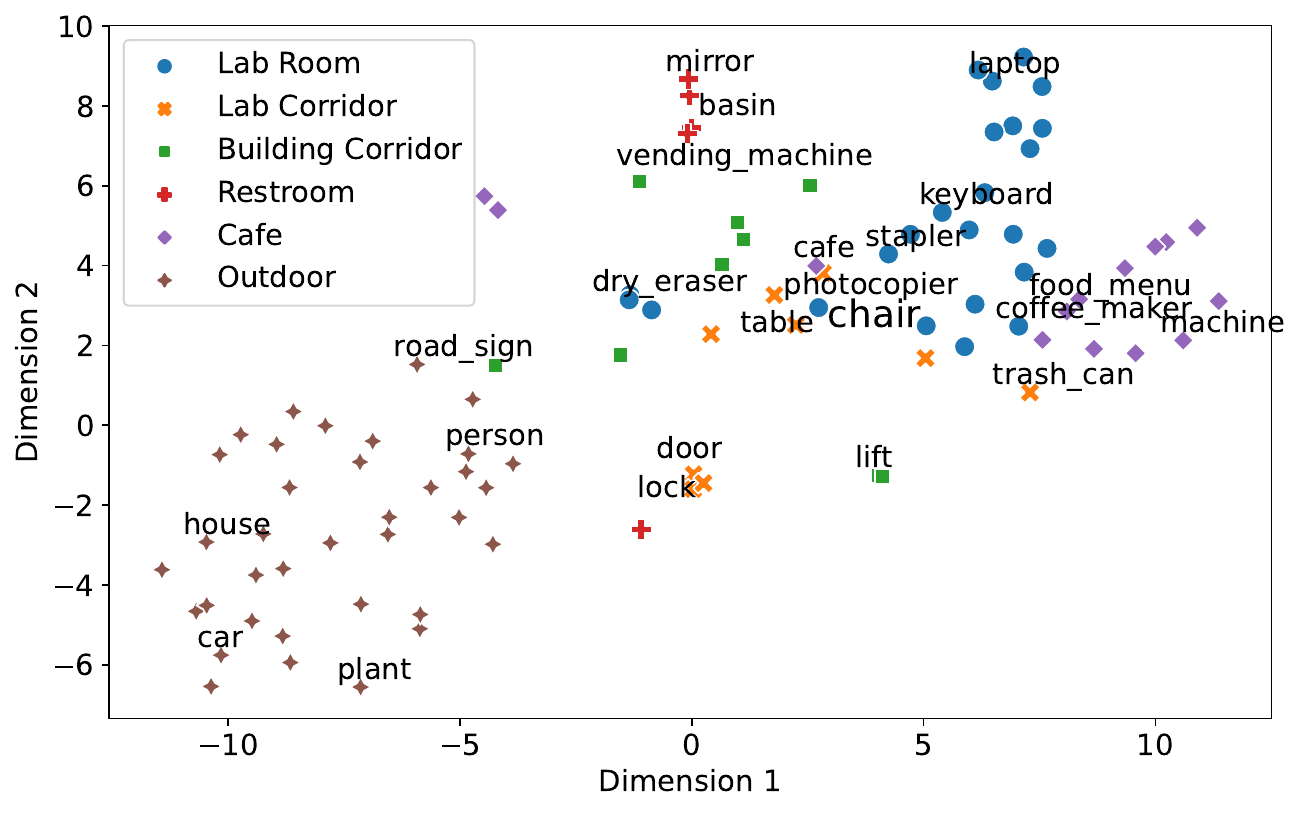}
\end{center} 
\caption{T-SNE plots of the embedding space considering similar objects appearing in multiple scenarios.}
\label{fig:tsneContextNeighborError}
\end{figure}

In our static contextual embedding model, each object is represented by a singular embedding vector that captures the objects' overall context for the entire video. An object can have contextual relationships with many other objects in different scenarios. An object's static contextual embedding vector captures a summarized view of the contexts of all scenarios. For the case study presented in Figure~\ref{fig:tsneContextCurrentFrame} and~\ref{fig:tsneContextNeighborFrame}, we considered objects that are present in unique scenarios and removed multiple appearances of the same object in different scenes. For example, if ``Chair'' appeared in two scenarios -- lab and cafe -- for experiments with static contextual embeddings, we removed the chair from one scenario, in this case, the lab. The projection of embedding vectors in Figure~\ref{fig:tsneContextNeighborFrame} shows that static embeddings effectively distinguish objects from different scenarios. However, when an object appears in multiple scenarios throughout the video, the static embedding tends to amalgamate contexts from diverse scenes, pulling other relevant objects in the neighborhood of the reference object. To examine this aspect, we trained our static embedding model with objects appearing in several scenarios. For instance, a chair object in training data is detected in a lab room, lab corridor, and near a cafe. Figure~\ref{fig:tsneContextNeighborError} shows that the chair's embedding vector is proximate to objects cafe, photocopier, and stapler from different scenes. The neighborhood indicates that the chair appears in multiple scenes, and the static embedding aggregates and represents the context from all these varied scenes. 

While the static representation has its strength in providing a summary of all contexts of the object, it falls short in indicating specific time frames of specific contexts like lab or cafe context. Moreover, depending on analytical needs, a downstream application might consider this aspect of mixing contexts from different scenarios as a strength or weakness. The static contextual embeddings are meaningful for an application that needs to identify all contexts associated with an object. An application that needs to identify the most prominent context of an object will require postprocessing to remove duplicate labels in different scenarios based on analytical needs. 

A downstream application that needs to realize the evolution of the context of an object will require the temporal embeddings addressed in the following subsection.

\subsection{Temporal Embedding Model}\label{sec:exp_temporal}
The following sub-sections discuss the experimental evaluation of temporal embeddings, where our primary aim is to address the subsequent questions.

\begin{enumerate}
\item  How do different objective functions perform in predicting the neighboring objects of a target object within frames? ~\ref{sec:exp:objective_function_selection}

\item Can our model effectively detect contextual relationships that extend across long distances?~\ref{sec:exp:objective_function_selection}

\item Does our model successfully track the evolution of objects' context in video?~\ref{sec:exp:change_context_overtime}

\item How can we utilize temporal contextual object embedding to construct a narrative for a video?~\ref{sec:video_summarization}

\item Does our model enhance performance in applications such as classification when integrated with visual features?~\ref{sec:exp:contextual_classification}

\end{enumerate}

\subsubsection{Selection of optimal objective functions:}\label{sec:exp:objective_function_selection}
We incorporated various features to formulate the objective functions of our temporal embedding model, as detailed in the methodology section (Section ~\ref{sec:methodology}). The functions include diverse adaptations of spatial distance between objects and the diffusion of object frequency. In this subsection, we evaluate the effectiveness of the objective functions through both quantitative and qualitative analysis.

\textbf{Quantitative Evaluation: }
In this quantitative evaluation, we assess whether our temporal object embeddings can accurately identify the actual nearest neighbors of objects. We have employed the hit@k metric to evaluate the effectiveness of each objective function in identifying potential neighbors of objects within a timestamp. This metric calculates the average intersection of k nearest neighbors between the actual neighbors and neighbors identified by the temporal embeddings of randomly chosen objects in random timestamps. We determined these actual neighbors of an object in a timestamp based on the average Euclidean distance of the surrounding objects within the frames in that timestamp. This group of neighbors is denoted as the base neighbors.

We randomly choose video-frames for training and testing sets. We generate temporal embedding vectors using different objective functions trained on the object pairs in the training set frames. Subsequently, these generated temporal embedding vectors are utilized to identify the nearest neighbors of each object in the timestamps of the test set frames. The hit@k is computed by comparing the base neighbors and the neighbors identified by our objective functions for each object in the test frames. The average hit@k is calculated across randomly chosen objects in the test set frames to obtain a comprehensive performance metric.
\begin{figure}[!h]
\begin{center}
  \includegraphics[width=0.8\textwidth]{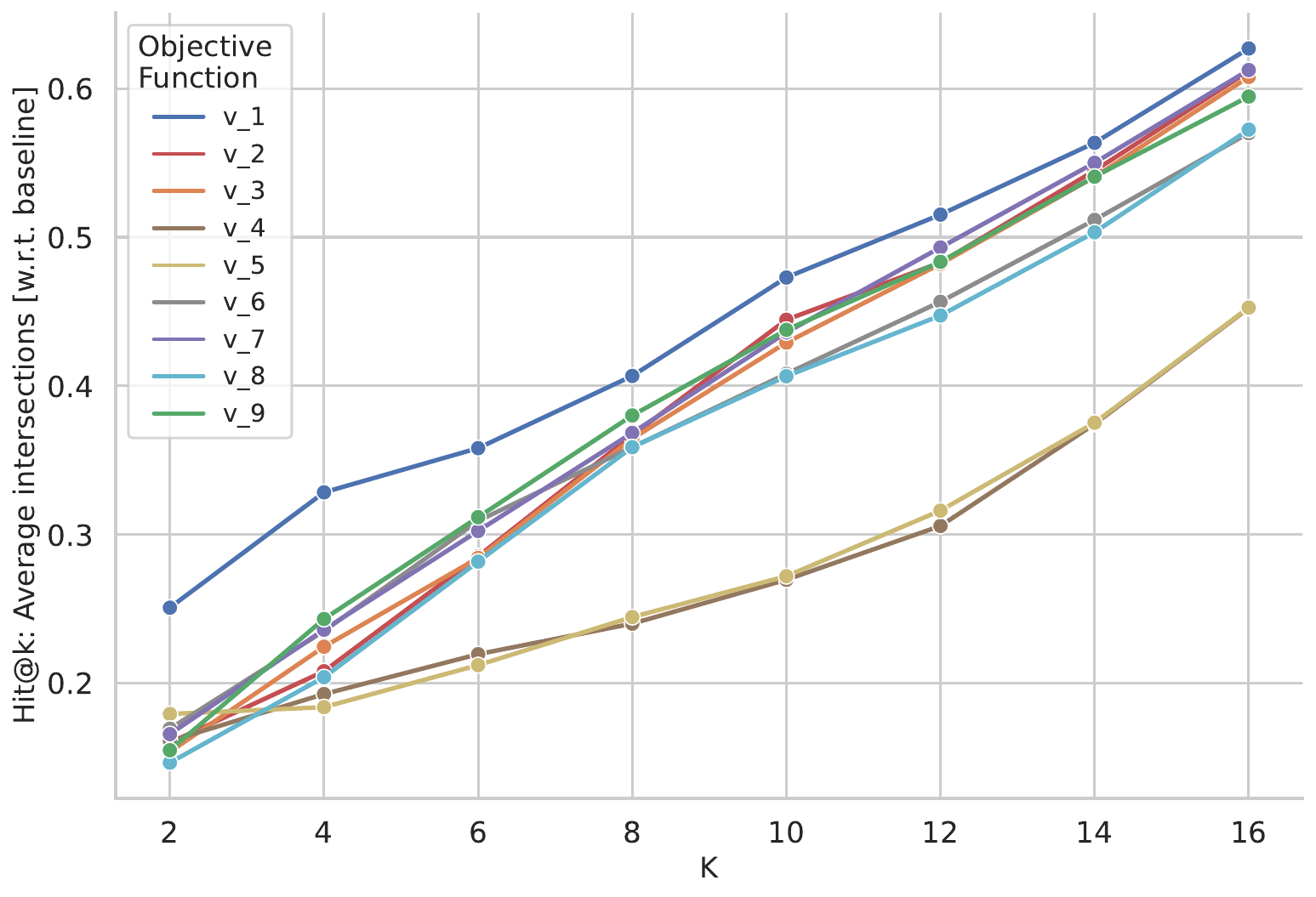}
\end{center}
\caption{Performance of each objective function in terms of the hit@k metric.}
\label{fig:exp_1_hit@k}
\end{figure}

Figure~\ref{fig:exp_1_hit@k} displays the hit@k scores for various values of k across different objective functions. The figure shows that our primary objective function ($v_1$) consistently achieves the highest hit@k values for each k. This superior performance can be attributed to the fact that the objective function, $v_1$ in (Equation~\ref{eq:theta_1_temporal}), employs normalized Euclidean distance between objects, which aligns closely with the method used to calculate the base neighbors. Nonetheless, this objective function does not take into account temporal diffusion and therefore is unable to capture more distant relationships between objects. This limitation is further explained in the qualitative evaluation part of this section.

The remaining objective functions in our study integrate additional aspects like temporal diffusion and the frequency distribution of objects besides Euclidean distances. These additional features are introduced for including context from surrounding timestamps and considered noise compared to baseline. Despite incorporating temporal diffusion and object frequency distribution, the objective functions labeled $v_2$, $v_3$, $v_7$, and $v_9$ display performance metrics closely mirroring the primary objective function ($v_1$), showcasing their effectiveness in identifying object neighbors.  Conversely, objective functions $v_4$ and $v_5$ show noticeably lower hit@k values. Furthermore, functions $v_6$ and $v_8$ perform less effectively compared to $v_2$, $v_3$, $v_7$, and $v_9$. The reduced efficacy of these functions can be attributed to their emphasis on frequency distribution, which introduces a substantial amount of noise into the model.

Based on these findings, objective functions $v_1$, $v_2$, $v_3$, $v_7$, and $v_9$ have been selected for further experimental analysis due to their demonstrated effectiveness in capturing the neighbors of objects.

\textbf{Qualitative evaluation:}
The prior hit@k experiment revealed that certain objective functions effectively identify objects' neighbors. In the previous setup, our base model considers neighbors of objects from surrounding frames. However, if an event occurs several frames before, and a similar environment is seen again near the event area after several timestamps, a base model can not capture the potential neighbors.
% \begin{figure}[!h]
%     \centering
%     \begin{subfigure}[b]{0.33\textwidth}
%         \includegraphics[width=\textwidth]{Figures/paper1/11_lab.jpg}
%         \caption{Lab}
%         \label{fig:lab}
%     \end{subfigure}
%     ~ %add desired spacing between images, e. g. ~, \quad, \qquad, \hfill etc. 
%       %(or a blank line to force the subfigure onto a new line)
%     \begin{subfigure}[b]{0.30\textwidth}
%         \includegraphics[width=\textwidth]{Figures/paper1/11_coridor1.jpg}
%         \caption{Corridor}
%         \label{fig:corridor}
%     \end{subfigure}
%     ~ %add desired spacing between images, e. g. ~, \quad, \qquad, \hfill etc. 
%     %(or a blank line to force the subfigure onto a new line)
%     \begin{subfigure}[b]{0.30\textwidth}
%         \includegraphics[width=\textwidth]{Figures/paper1/11_toilet.jpg}
%         \caption{Restroom}
%         \label{fig:lift}
%     \end{subfigure}
%     \begin{subfigure}[b]{0.30\textwidth}
%         \includegraphics[width=\textwidth]{Figures/paper1/11_lift_censored.jpg}
%         \caption{Elevator}
%         \label{fig:lift}
%     \end{subfigure}
%     \begin{subfigure}[b]{0.30\textwidth}
%         \includegraphics[width=\textwidth]{Figures/paper1/11_cafe.jpg}
%         \caption{Cafe}
%         \label{fig:cafe}
%     \end{subfigure}
%     \begin{subfigure}[b]{0.30\textwidth}
%         \includegraphics[width=\textwidth]{Figures/paper1/11_outdoor.jpg}
%         \caption{Outdoor}
%         \label{fig:outdoor}
%     \end{subfigure}
%     \caption{Frames from recorded video.}\label{fig:frames}
%     \label{fig:framesRecorded}
% \end{figure}
In this qualitative study, our goal is to evaluate the capability of our model to identify object contexts over longer distances. we created a synthetic dataset depicting an event near a school at a specific time. This event scenario involved students and certain individuals labeled as `persons of interest' close to a scene of malicious activity. After several timestamps, these persons of interest reappeared near the students at the same event location. Through analyzing temporal embeddings generated by various objective functions, this study aims to assess whether our model can successfully indicate a heightened correlation between the malicious event and the entities present in the event area despite the missing malicious event label in the later timestamp.

\begin{figure}[!h]
\begin{center}
  \includegraphics[width=0.8\textwidth]{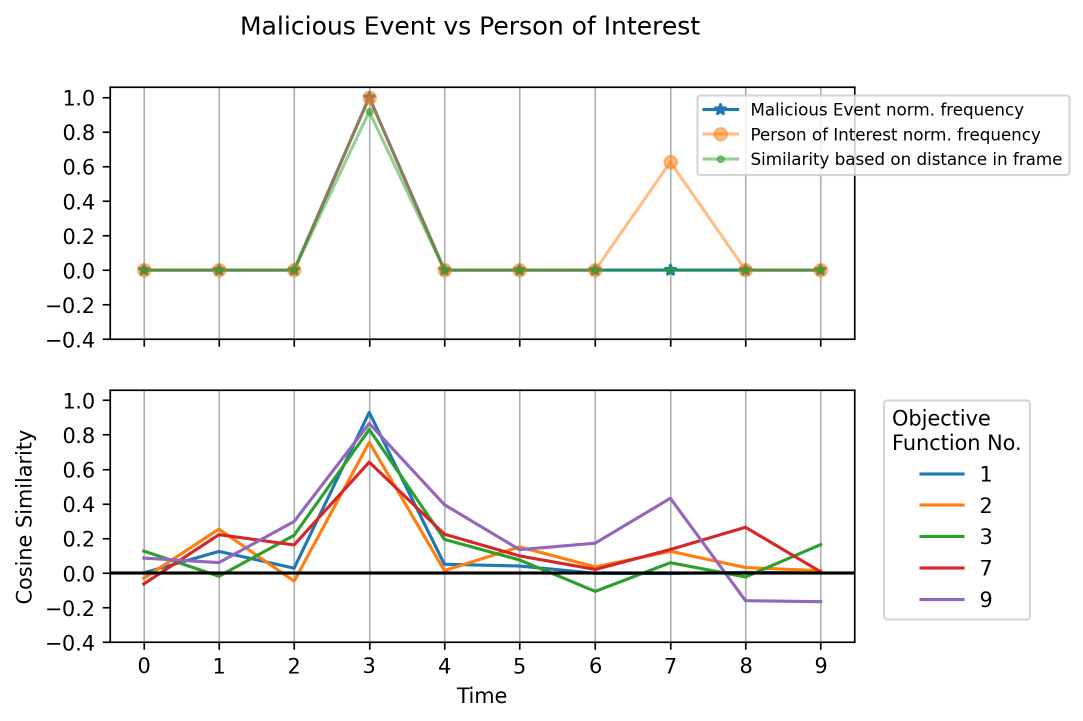}
\end{center}
\caption{The upper graph displays the distribution of frequencies and base similarity based on distance in frames between ``Malicious Event'' and ``Person of Interest''. The lower plot shows the cosine similarity between the temporal embeddings of these two entities, comparing the results of different objective functions.}
\label{fig:exp_1_cos_sim1}
\end{figure}

Figure~\ref{fig:exp_1_cos_sim1} features two distinct plots. In the upper plot, the blue and orange lines indicate the normalized frequency of a malicious event and a person of interest, respectively. A notable observation is the high frequency of both objects at timestamp 3. The green line in the upper plot measures the similarity based on the spatial distance between the event and the person of interest across different timestamps. A peak in the green line at timestamp 3 indicates that the event and person of interest are nearby in some frames at this timestamp. Post time 3, the green line drops to zero, indicating no co-occurrence of the event and person of interest in subsequent frames. Additionally, the orange line (frequency of person of interest) peaks again at timestamp 7, signaling the reappearance of the person of interest near the school. This study focuses on determining whether our embedding vectors can predict the potential connection between a malicious event and a person of interest at timestamp 7, even though the malicious event label is not present in timestamp 7 or in any nearby timestamps.

The bottom plot in Figure~\ref{fig:exp_1_cos_sim1}, illustrates the cosine similarity between the embeddings of the event and the person of interest across various timestamps. In this plot, it is evident that embedding vectors from all the objective functions successfully capture the similarity between the event and person of interest at timestamp 3. Among all the evaluated objective functions, only objective function 9 exhibits increased similarity between the malicious event and the person of interest at timestamp 7. This is portrayed by the rise of the line associated with objective function 9 (purple color) in timestamp 7. The rise at timestamp 7 highlights the unique ability of objective function 9 to recognize and emphasize a previously learned connection between the event and the person, even though the event label is not present in timestamp 7.

\begin{figure}[!h]
\begin{center}
  \includegraphics[width=0.8\textwidth]{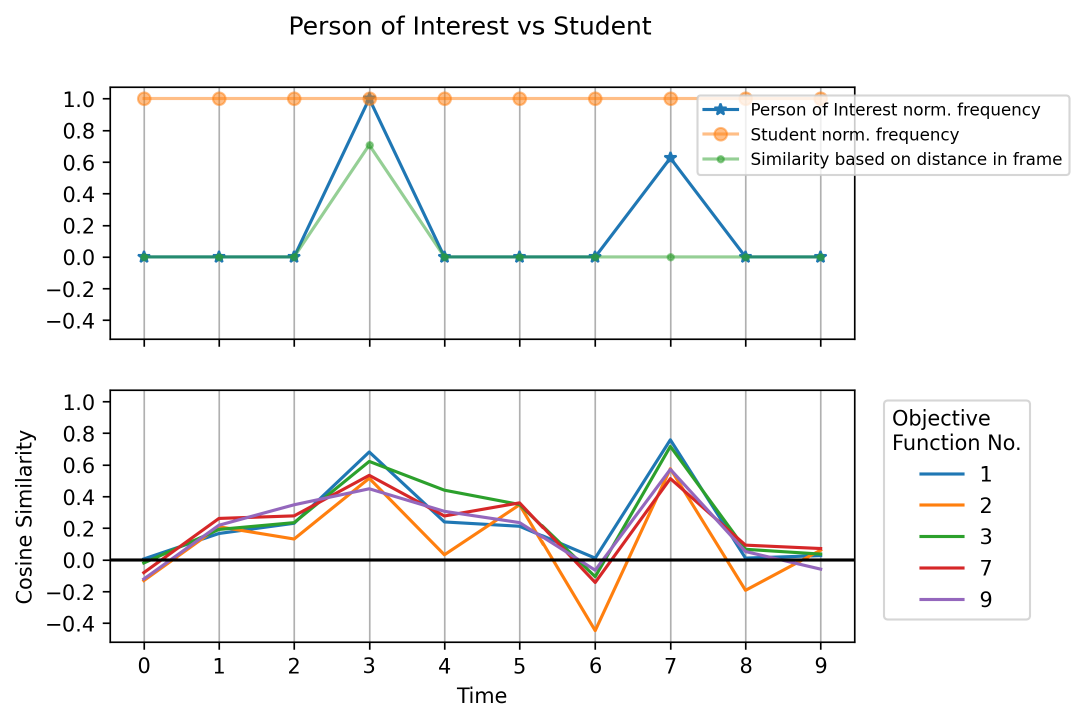}
\end{center}
\caption{The upper graph displays the distribution of frequencies and base similarity based on distance in frames between ``Person of Interest'' and ``Student''. The lower plot shows the cosine similarity between the temporal embeddings of these two entities, comparing the results of different objective functions.}
\label{fig:exp_1_cos_sim2}
\end{figure}

In the first analysis(~\ref{fig:exp_1_cos_sim1}), we plotted the cosine similarity between the embeddings of objects and events at different timestamps. Figure~\ref{fig:exp_1_cos_sim2} shows the cosine similarity between a person of interest and a student. Here, embeddings from various models consistently indicate increased similarity between the student and the person of interest at timestamps 3 and 7, aligning with their frequent appearances. Interestingly, even though the base similarity of these two objects is zero at timestamp 7 (green line on top), implying they do not co-occur in the same frame, our temporal embedding still manages to capture their similarity. This is due to the ``context selection from surrounding frames'' approach and the diffusion technique employed in our model. Notably, of all the objective functions tested, objective function 9 demonstrates the highest increase in similarity between these two entities at timestamp 7.

\begin{figure}[!ht]
\begin{center}
  \includegraphics[width=0.8\textwidth]{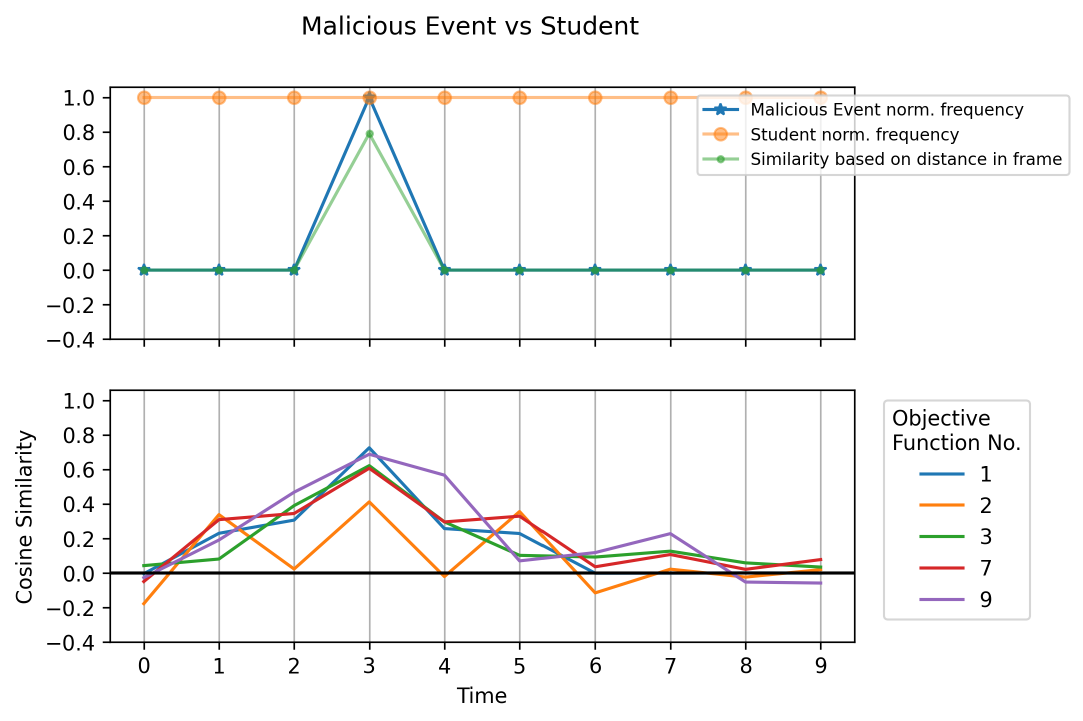}
\end{center}
\caption{(top) The distribution of frequencies and base similarity based on distance in frames between ``Malicious Event'' and ``Student''. (bottom) The cosine similarity between the temporal embeddings of these two entities, comparing the results of different objective functions.}
\label{fig:exp_1_cos_sim3}
\end{figure}

Figure~\ref{fig:exp_1_cos_sim3} shows the cosine similarity between a malicious event and a student. Here, the frequency of malicious activity suggests that the activity happened at the same time when the person of interest was present in the scene (timestamp 3). The cosine similarity between embeddings from different models suggests that the similarity between student and malicious activity increased at timestamp 3. However, we can also see a small peak at timestamp 7, though the malicious activity has zero frequency during that time. The increase in cosine similarity at timestamp 7 suggests that the embeddings of event and student come closer at timestamp 7. That is because, though the event does not happen at timestamp 7, the person of interest who is related to the event is present at timestamp 7 (Figure~\ref{fig:exp_1_cos_sim3}). This shows the ability of our temporal embedding model to detect contextual connections that extend across long distances.

\subsubsection{Change in object context over time}\label{sec:exp:change_context_overtime}
In this experiment, we analyzed how our temporal embeddings capture the change in object context over time. An object can be seen or detected in different contexts at different times. 
Previously, in Figure~\ref{fig:tsneContextNeighborError}, we observed that if objects appear at different times in different contexts, static contextual embeddings result in a mixed or summarized context. Static contextual embedding cannot capture the change in context as it generates a single embedding vector of an object for the entire video. Analysis of the contextual change of objects over time can help us understand the scene transition in a video. Our temporal embedding model generates embedding vectors for each object at each timestamp. 

\begin{figure}[!h]
\begin{center}
  \includegraphics[width=1\textwidth]{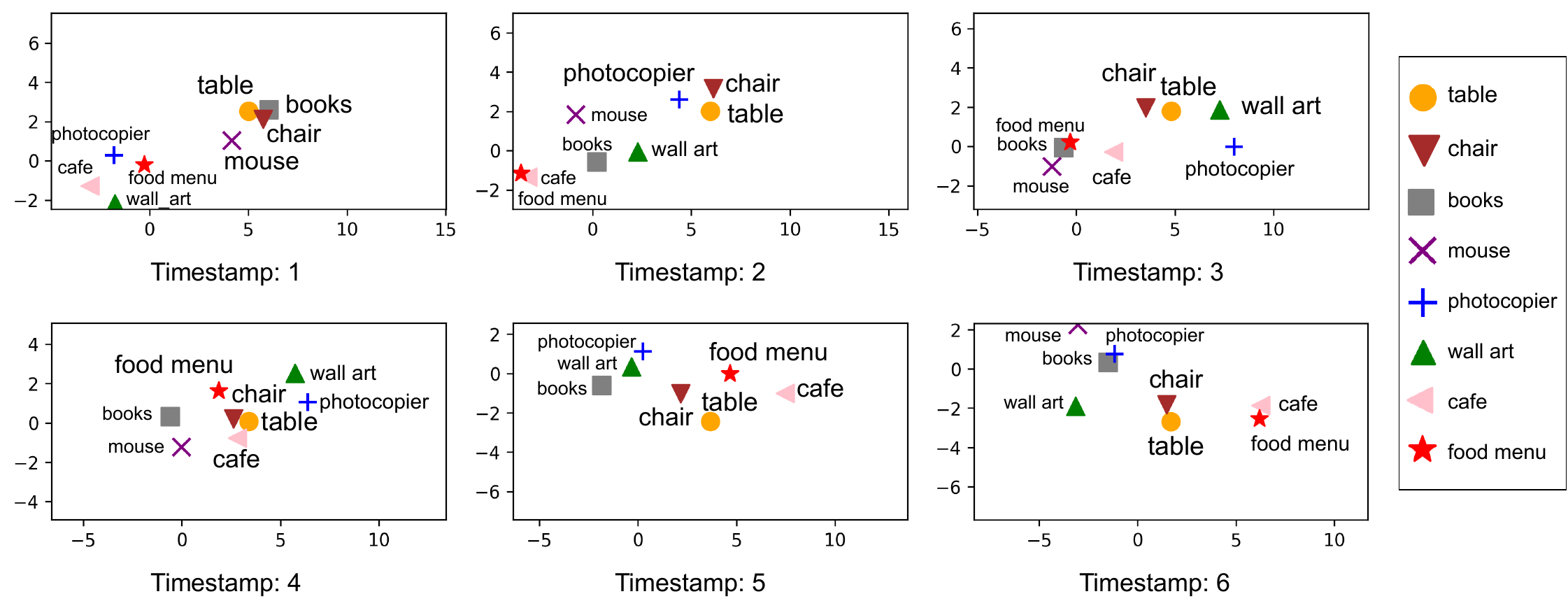}
\end{center}
\caption{PCA vectors of ``chair'' and some neighbor objects at different timestamps.}
\label{fig:exp_2_pca_campus}
\end{figure}

In Figure \ref{fig:exp_2_pca_campus}, we projected the embedding vectors in 2D space utilizing the Principal Component Analysis (PCA) to analyze context evolution. In this subsection, we focus on a subset of objects and their surrounding context in the embedding space to better visualize contextual changes over time, avoiding cluttering. This experiment is conducted on the campus video and synthetic video data.

The PCA spaces of Figure~\ref{fig:exp_2_pca_campus} display the nearest neighbors of the embedding vector of ``chair'' at various timestamps. We see that the ``chair'' is initially associated with objects related to lab and lab corridor scenes, such as books, a photocopier, and wall art. As time progresses, the ``chair'' becomes more closely related to items found in the cafeteria setting, such as a cafe and food menu. This shift in the context of the ``chair'' corresponds with the video's story, where the scene transitions from a lab room scene to a cafeteria scene.

To evaluate if our temporal embedding can capture context from a longer temporal distance, we included some malicious activity around a school building in the synthetic data. In some of the frames, some people appear with a van and are seen walking near the school area while students enter the school campus. After some frames, these people (whom we refer to as persons of interest) are seen to be involved in malicious activity (at Timestamp 3). After the departure of these persons of interest, police officers are seen near the school. After several frames, the persons of interest are again seen near the school complex (at Timestamp 7). We have trained our temporal object embedding model with the video data and analyzed if our temporal object embedding model can capture the temporal relation between objects and events.

\begin{figure}[!ht]
\begin{center}
  \includegraphics[width=1\textwidth]{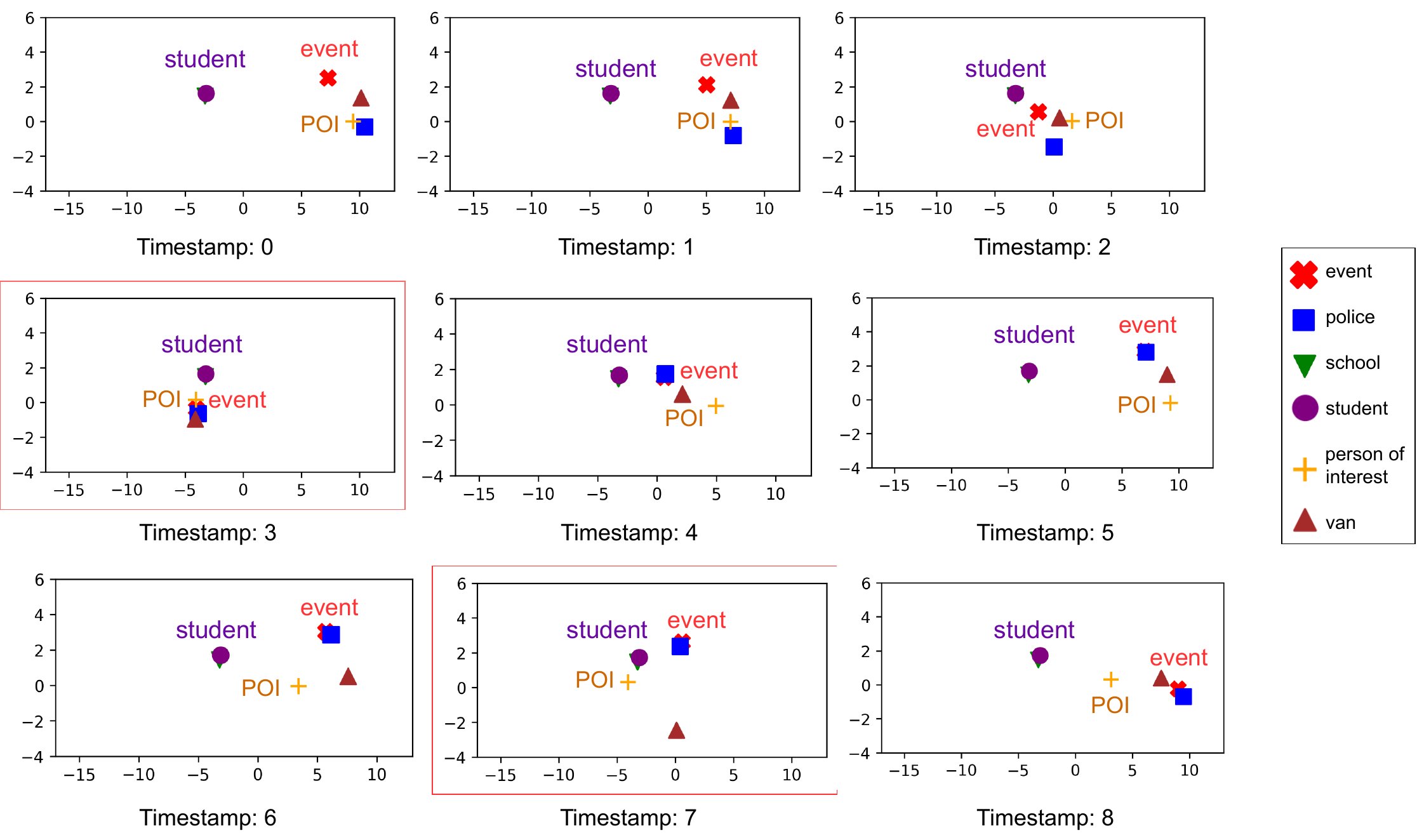}
\end{center}
\caption{PCA vectors of ``student' and some neighbor objects at different timestamps.}
\label{fig:exp_2_pca_student_1}
\end{figure}

The plotting of temporal embeddings at different timestamps in Figure~\ref{fig:exp_2_pca_student_1} shows that the event and person of interest gradually come closer to a student from timestamps 1 to 3 and then gradually separate in timestamps 4 and 5. Despite the person of interest only being present near the school area at timestamps 3 and 7, our embedding vectors exhibit a gradual transition. This smooth change in the embeddings is a result of the temporal diffusion technique employed in our model.

At timestamp 7 in our synthetic data, a person of interest is again seen near the school area, but there are no malicious events or police involvement at this time. The embedding space depicted in Figure~\ref{fig:exp_2_pca_student_1} for timestamp 7 shows the embedding vector of the person of interest in proximity to the student. Interestingly, the embeddings of both the event and the police also appear closer to the student now. This proximity of the embedding vector of the event and police to the embedding vector of the student is due to the similarity in the embeddings of the person of interest, the event, and the police across various timestamps. The increased similarity in timestamp 7 stems from their co-occurrence in the same or adjacent frames at timestamp 3. As a result, whenever any object from this interconnected group appears in a different scenario, the embeddings tend to bring the other related objects closer within the scene, as reflected in the embedding space at timestamp 7.

\subsubsection{Narrating the Formed Context using ChatGPT}\label{sec:video_summarization}

In this section, we explored the potential of temporal embeddings to encapsulate the narrative within a video. For this study, we calculated the cosine similarity between the embedding vectors of all object pairs at each timestamp. Subsequently, these similarity scores were sorted, and the pairs with the highest similarity for each timestamp were identified. The aim here is to inspect whether the most similar object pairs at each timestamp are contextually relevant to the scenes occurring at those specific times.

\begin{figure}[!h]
\begin{center}
  \includegraphics[width=1\textwidth]{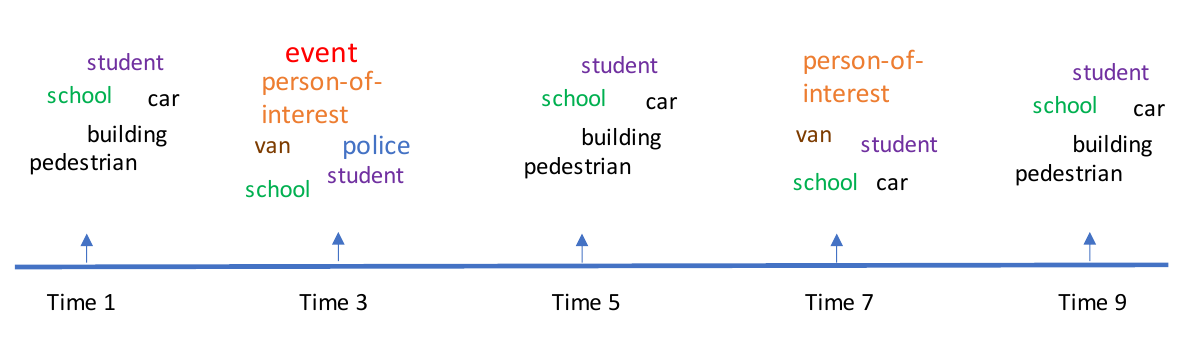}
\end{center}
\caption{Top similar objects at different timestamps in the synthetic data.}
\label{fig:top_sim_synthetic_data}
\end{figure}

Figure~\ref{fig:top_sim_synthetic_data} showcases the most similar objects at different timestamps within our synthetic dataset, as determined by the temporal embeddings of these objects. As depicted in the figure, the top similar objects across various timestamps reveal consistent patterns in our synthetic dataset, which was designed around a school setting. Commonly recurring objects in the dataset, like the school, students, buildings, and pedestrians, consistently rank high in similarity across all timestamps. In particular, at timestamps 3 and 7, the appearance of unusual entities such as the event and person of interest among the top similar objects indicates that our embedding model successfully identifies entities that are contextually significant at specific timestamps.

To generate a narrative using these contextually relevant objects at each timestamp, we fed some top similar object pairs along with their respective timestamps into ChatGPT. The actual prompt given to ChatGPT is as follows:

\fbox{
\begin{minipage}{12 cm}
A security camera captured the following object pairs at different time, can you summarize the video:

Time 1: (car1, student2), (student1, student4), (building1, tree2), (building1, random-walker1), (school, student1), (car1, school)\\
Time 2: (car1, student2), (student1, student4), (building1, tree2), (building1, random-walker1), (school, student1), (car1, school), (random-walker1, random-walker3)\\
Time 3: (office-worker2, office-worker3), (random-walker1, random-walker3),  (student2, student3),  (office-worker1, truck2), (car1, car2), (school, student2)\\
Time 4: (Malicious Event, police1), (man, lady), (suv, lady), (building1, police-car2), (police2, person-of-interest1),  (student1, student3), (Malicious Event, tree1)\\
Time 5: (random-walker2, random-walker4), (random-walker4, office-worker3), (random-walker2, office-worker3), (school, student3), \\
Time 6: (random-walker2, random-walker4), (man, lady), (suv, man), (suv, lady),(random-walker4, office-worker3), (random-walker2, office-worker3), (school, student3),  (car2, student4), \\
Time 7: (office-worker2, office-worker3), (random-walker1, random-walker3),  (student2, student3),  (office-worker1, truck2), (car1, car2), (school, student2)\\
Time 8: (car1, car5), (suv, man),  (suv, lady),  (man, lady), (car1, car2), (school, student2), building1, (van-of-interest), (school, person-of-interest1)\\
Time 9: (car1, student2), (student1, student4), (building1, tree2), (building1, random-walker1), (school, student1), (car1, school), (random-walker1, random-walker3)\\
Time 10: (car1, student2), (student1, student4), (building1, tree2), (building1, random-walker1), (school, student1), (car1, school), (random-walker1, random-walker3)
\end{minipage}
}

The generated narrative output is presented below.

\noindent\textit{Time 0-1: Normal activities with students, cars, and random walkers near a school and a building.}

\noindent\textit{Time 2: Presence of office workers, cars, and students. Some are seen with vehicles.}
\noindent\textit{Time 3: A malicious event occurs. Police and a person of interest are seen. Students and other individuals are also present.}

\noindent\textit{Time 4-5: Random walkers and office workers are visible. The school is seen with students.}

\noindent\textit{Time 6: Similar to Time 2, with office workers, students, and vehicles.}

\noindent\textit{Time 7: Presence of multiple cars, a van of interest, and a person of interest near the school and building.}

\noindent\textit{Time 8-9: Repeats the events of Time 0 and Time 1.}

\noindent\textit{The events seem to cycle back to normal activities after a period of disruption (Time 3 and Time 7), suggesting possible resolution or stabilization post the events observed.}

Given the contexts discovered by our model, ChatGPT generated a narrative that describes the video reasonably well. This study demonstrates that temporal contexts in conjunction with an LLM can provide a meaningful narrative-based summary of a video.

\begin{figure}[!h]
\begin{center}
  \includegraphics[width=1\linewidth]{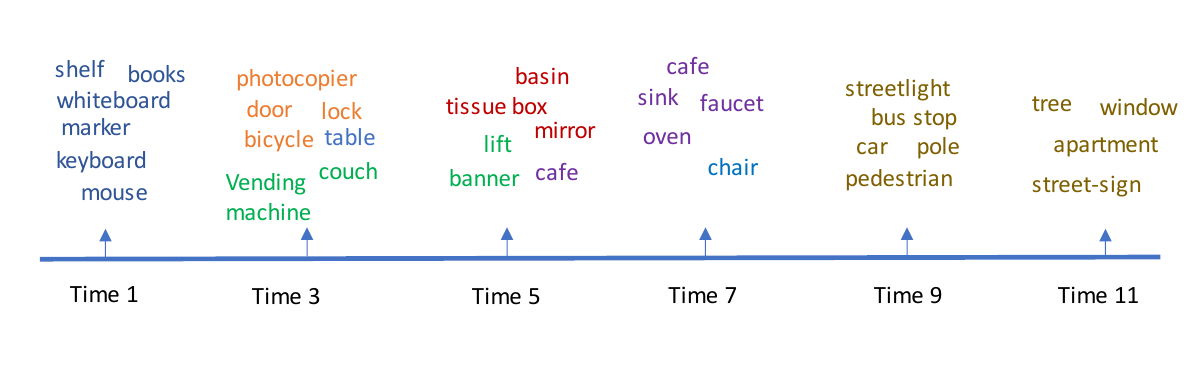}
\end{center}
\caption{Top similar objects at different timestamps in campus video data.}
\label{fig:top_sim_campus_data}
\end{figure}

Additionally, we extended this study to the campus video dataset. Our temporal embedding model identified the most similar objects at various timestamps, as shown in Figure~\ref{fig:top_sim_campus_data}. In this figure, the text colors represent the six consecutive scenarios captured in the campus dataset (referenced in Figure~\ref{fig:framesRecorded}).

The narrative generated by the Large Language Model using these contextually relevant objects at each timestamp is presented below.

\noindent\textit{Time 0: Office Environment: Items like a keyboard, mouse, coffee maker, computer monitor, speaker, shelf with books, papers with a laptop, marker, dry eraser, and whiteboard are observed, suggesting a typical office or study area.}

\noindent\textit{Time 2: Outside Transition Area: Objects like a lock, outlet, bicycle, door, chair, photocopier, and wall art are seen, indicating a transition from indoor to outdoor spaces or a utility area.}

\noindent\textit{Time 4: Common Area: The presence of a basin, mirror, tissue box, vending machine, wall art, couch near an elevator, photocopier, and water dispenser suggests a common area or a more public space within the building.}

\noindent\textit{Time 6: Cafeteria or Kitchen Area: Items like a sink, faucet, cafe with a food menu, oven, vending machine, jug, coffee maker, and credit card terminal are indicative of a cafeteria or kitchen area.}

\noindent\textit{Time 8: Outdoor Campus Area: The observation of a railing with stairs, a building with a street light, bush plant with a pole, building with a car, and a tree with a car suggests outdoor campus areas, possibly pathways or parking areas.}

\noindent\textit{Time 10: Street-side or Bus Stop Area: The capture of a bus stand with a road sign, pedestrian with a road sign, light pole with a road sign, a road sign with a street sign, building with a fire hydrant, light pole with a bus stop, and street light with a bus stop indicates a street-side area or a bus stop near the university.}

\noindent\textit{Summary: The video captures various facets of university life, transitioning from indoor office or study environments to outdoor and public spaces like common areas, cafeterias, and outdoor campus areas, finally moving towards street-side scenes near the university. The change in object pairs over time depicts a comprehensive view of university life, encompassing both academic and recreational aspects.}

The narrative of the campus video content produced by ChatGPT is reasonably accurate and thorough. This further demonstrates the capability of our temporal embeddings in creating prompts to effectively summarize video content using a large language model.

\subsubsection{Contextual classification}\label{sec:exp:contextual_classification}

The objective of this experiment is to explore how well temporal contextual embedding vectors can perform in a contextual classification problem for objects. We seek to answer the question, ``Do contextual embedding models improve the contextual classification of objects in video data.'' Here, contextual classification refers to grouping objects that contain similar contexts.
\begin{figure}[!h]
\begin{subfigure}{.48\textwidth}
  \centering
  \includegraphics[width=0.80\textwidth]{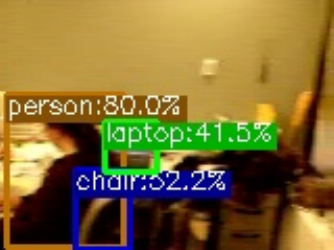}
  \caption{An indoor scene in the LabelMe dataset}
  \label{subfig:labelme_example_frame}
\end{subfigure}
\begin{subfigure}{.48\textwidth}
  \centering
  \includegraphics[width=0.98\textwidth]{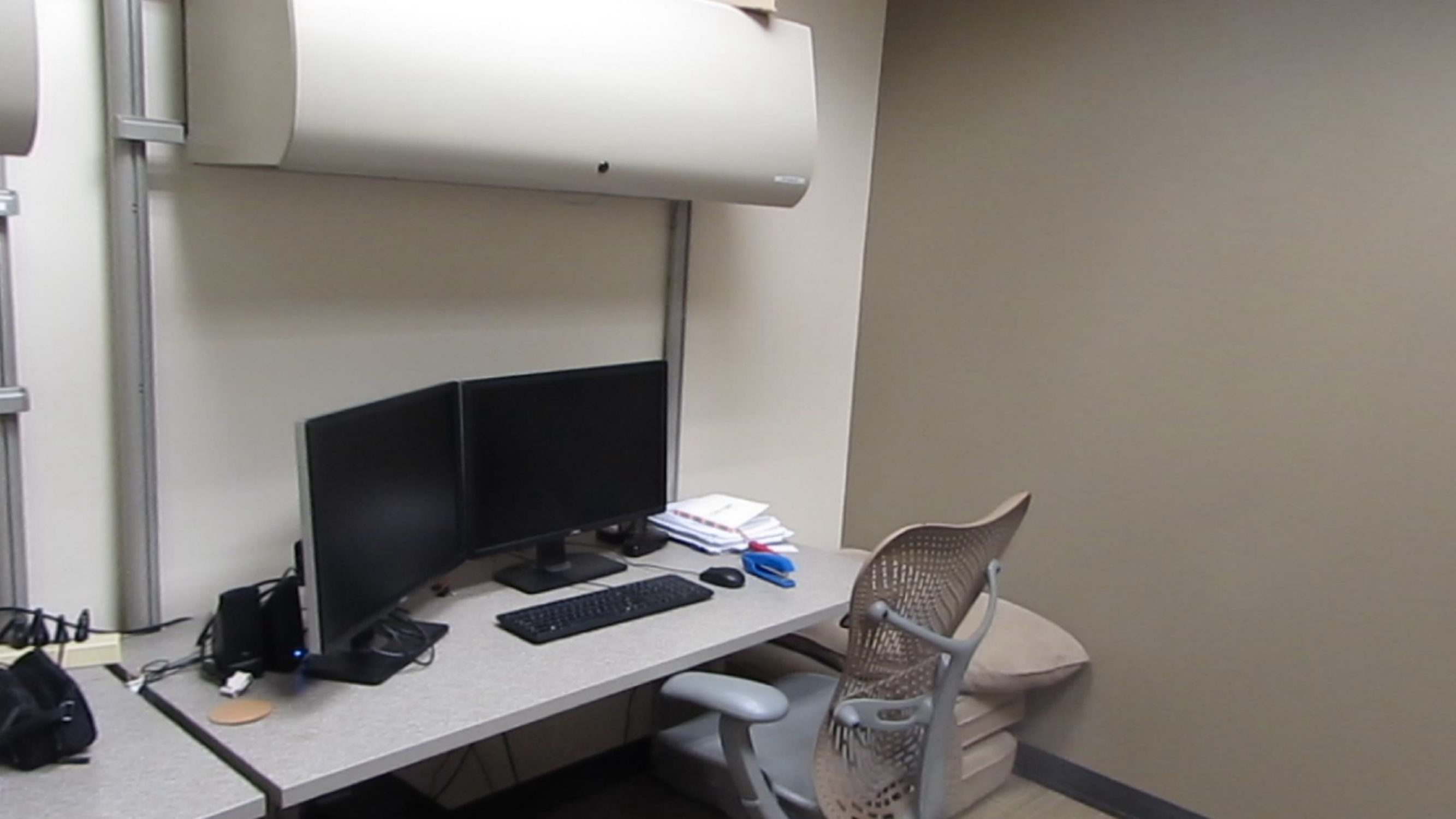}
  \caption{A lab room scene in the Campus dataset}
  \label{subfig:campus_example_frame}
\label{fig:example_frames_labelme_campus}
\end{subfigure}
\caption{Example frame from LabelMe and Campus dataset.}
\end{figure}

We conducted experiments on the classification task using two different sets of video data,

\begin{enumerate}
    \item LabelMe sequential frames:  The LabelMe database consists of a vast array of images and sequential frames created for object detection and recognition tasks~\cite{yuen09}. For this experiment, we specifically gathered the ``sequential frames'' dataset from LabelMe and employed YOLO for object detection in those frames. Sequential frames in the LabelMe dataset are labeled as either indoor or outdoor, therefore, we classified the detected objects into two categories -- indoor and outdoor -- based on the scene where each object was detected. An example frame from the LabelMe dataset is shown in Figure~\ref{subfig:labelme_example_frame}.
    \item Campus video: This dataset is the campus video we discussed in section~\ref{sec:case2}. As previously mentioned, all objects in this video were manually classified into six distinct categories: Lab room, Lab corridor, Building corridor, Restroom, Cafeteria, and Outdoor. An example frame from the campus video dataset is shown in Figure~\ref{subfig:campus_example_frame}.
\end{enumerate}

In this experiment, we evaluated:

\begin{itemize}
    \item The effectiveness of visual features alone, extracted using CNN or ResNet, in classifying objects into their respective categories.
    \item The capability of our temporal embedding vectors, which incorporate contextual features, in classifying objects without incorporating any visual features.
    \item The efficiency of combining temporal object embedding with visual features for object classification.
\end{itemize}

We split all detected object instances into training and test sets. Before classification, we extracted feature vectors from these instances employing five distinct approaches, which are elaborated below. The feature vectors of training set data are then utilized to train a classifier. Following this, we used the trained classifier to predict the classes of the objects in the testing data set.

Some explanations of model settings to generate feature vectors are provided below.
\begin{enumerate}
\item Basic CNN Model: The CNN model comprises three convolutional layers, with each layer followed by a sequence of a ReLU layer, a MaxPooling layer, and a Dropout layer. The output from the final convolutional layer is then flattened and passed through three dense layers for the classification task.
\item ResNet50 Model: Visual features of objects are extracted using a pre-trained ResNet50 model. Subsequently, these feature vectors are processed through three dense layers to perform the classification task.
\item Our temporal contextual embedding model: Our embedding model generates vectors for each object at every timestamp. To get the feature vector, we concatenate the embeddings of each object across all timestamps. These combined embedding vectors are subsequently processed through a dense layer for the classification task.
\item Fusion of the visual features with our temporal contextual embedding vectors: The feature vectors obtained from the CNN model or the ResNet50 model are concatenated with embedding vectors of our temporal embedding model. Then the fusion vectors are fed through a dense layer to perform the classification task.
\end{enumerate}

% \begin{table}[h]
% \caption{Context classification Accuracy}\label{tab:classification}%
% \begin{tabular}{|c|c|c|c|c|}
% \toprule
% Dataset &\# Object Instances &\# Classes  & Models & Accuracy\\
% \hline
% \multirow{5}{*}{LabelMe } & 17465    & 2   & CNN  & 61.76\%  \\
%  & 17465   & 2   & Our Embedding  & 100.00\%  \\
%  & 17465   & 2   & CNN + Our Embedding  & 100.00\%  \\
%  & 17465   & 2   & ResNet\footnotemark[1]  & 89.52\%\\
%  & 17465   & 2   & ResNet+Our Embedding  & 100.00\%\\
% \hline
% Campus & 1747    & 6   & CNN  & 59.44\%  \\
% Campus & 1747   & 6   & Our Embedding  & 88.33\%  \\
% Campus & 1747   & 6   & CNN + Our Embedding  & 88.89\%  \\
% Campus & 1747   & 6   & ResNet\footnotemark[1]  & 94.89\%\\
% Campus & 1747   & 6   & ResNet+Our Embedding  & 98.89\%\\
% \botrule
% \end{tabular}
% \footnotetext{Source: This is an example of table footnote. This is an example of table footnote.}
% \footnotetext[1]{Example for a first table footnote. This is an example of table footnote.}

% \end{table}

\begin{table}[h]
\centering
\caption{Accuracy comparison of different models on contextual object classification.}\label{tab:classification}%
\begin{tabular}{|c|c|c|c|c|}
\hline
Dataset & \begin{tabular}[c]{@{}c@{}}\# Object\\ Instances\end{tabular} &\# Classes  & Models & Accuracy\\
\hline
\multirow{5}{*}{LabelMe } & \multirow{5}{*}{17465 }    & \multirow{5}{*}{2 }   & CNN  & 61.76\%  \\
 &   &    & Our Embedding  & 100.00\%  \\
 &  &    & CNN + Our Embedding  & 100.00\%  \\
 &   &    & ResNet\footnotemark[1]  & 89.52\%\\
 &   &    & ResNet+Our Embedding  & 100.00\%\\
\hline
\multirow{5}{*}{Campus } & \multirow{5}{*}{1747 }     & \multirow{5}{*}{6 }   & CNN  & 59.44\%  \\
 &    &    & Our Embedding  & 88.33\%  \\
 &    &    & CNN + Our Embedding  & 88.89\%  \\
 &    &    & ResNet\footnotemark[1]  & 94.89\%\\
 &    &    & ResNet+Our Embedding  & \textbf{98.89}\%\\
\hline
\end{tabular}
\end{table}

Table~\ref{tab:classification} presents the accuracies obtained using the feature vectors explained earlier. Some key observations that emerge from these results are described below.
\begin{enumerate}
   \item Effectiveness of Temporal Contextual Embedding: As shown in the table, our temporal contextual embedding model outperforms the standalone CNN model in experiments on both datasets. In the LabelMe dataset, our embedding vectors achieve a 100\% accuracy rate. This high level of accuracy is attributed to the fact that indoor and outdoor objects can be easily distinguished based on their contextual features. Despite not being trained directly on visual features, our embedding model effectively classifies objects. Furthermore, when our embeddings are combined with CNN features, there is a notable 29\% increase in accuracy for the campus video dataset compared to using the CNN model alone. This improvement demonstrates the potential of our contextual embeddings to enhance classification tasks when used alongside visual features.
    \item ResNet50 and Our Temporal Contextual Embeddings: With the LabelMe dataset, the classification accuracy of the standalone ResNet50 model is not as high as that achieved by our temporal contextual embedding model. A possible explanation for this could be the low resolution of the LabelMe dataset videos, which may lead to lower-quality visual feature extraction. However, ResNet50 demonstrates notable accuracy in contextual classification with the campus video dataset. Moreover, the performance of ResNet50 is further enhanced when its features are combined with our contextual embedding vectors. In particular, combining ResNet50 features with our embeddings results in a 4\% increase in accuracy. This improvement underscores the significant contribution of contextual information in video object classification tasks.
\end{enumerate}

\section{Conclusion}

In this paper, we have introduced a temporal contextual object embedding model, which is designed to learn the contextual features of objects in videos, focusing on context rather than visual aspects. We introduce a novel diffusion mechanism that leverages the diffusion of object frequency and object spatial distance to generate temporal embeddings that can capture the context of visual objects and the evolution of context. The experimental analysis demonstrates that our temporal embeddings provide a representation capable of (1) capturing the context of objects, (2) modeling changes in objects' contexts as the video's scenario evolves, (3) summarizing videos and aiding in narrative generation through integration with Large Language Models (LLMs), and (4) enhancing downstream tasks such as classification when combined with visual features. Our model stands apart from traditional vision embedding models that primarily rely on visual features for embedding learning. Instead, we utilize contextual features to create contextual embeddings. As a result, our temporal embedding model excels in modeling context changes, offering deeper insights into scene transitions and the potential to identify threats by recalling past event associations. 

Looking ahead, we want to expand our research in context-aware video analysis, particularly in enhancing predictive analytics within video surveillance and threat detection. Additionally, we aspire to integrate Large Language Models (LLMs) for more sophisticated narrative generation and summarization capabilities where understanding the context and evolution of events in videos is crucial.

%%%%%%% Start Statements and Declarations %%%%%%%%%%%%%%%%%

\section*{Statements and Declarations}
\begin{itemize}
\item{Financial \& Non-financial interests:} The authors have no relevant financial or non-financial interests to disclose. All authors certify that they have no affiliations with or involvement in any organization or entity with any financial interest or non-financial interest in the subject matter or materials discussed in this manuscript. The authors have no financial or proprietary interests in any material discussed in this article.

\item{Conflicts of interest:} The authors have no conflicts of interest to declare that are relevant to the content of this article.
\end{itemize}
%%%%%%%%%%% End Statements and Declarations %%%%%%%%%%%%%%%%

\bibliography{sn-article.bib}% common bib file

\end{document}